\documentclass[preprint,12pt]{elsarticle}
\journal{Neurocomputing}

\usepackage{color}
\usepackage[utf8]{inputenc}
\usepackage[T1]{fontenc}
\usepackage{hyperref}
\usepackage{multirow} 
\usepackage{url}
\usepackage{graphicx} 
\usepackage[normalem]{ulem}
\useunder{\uline}{\ul}{}
\usepackage{amssymb,amsthm,amsmath,amsfonts}
\usepackage{algorithm}
\usepackage{algorithmicx} 
\usepackage{algpseudocode}

\begin{document}
\biboptions{sort&compress}
\begin{frontmatter}
\title{Physics-constrained Attack against Convolution-based Human Motion Prediction}
\author[1]{Chengxu Duan}\ead{dami0513@bupt.edu.cn}
\author[1]{Zhicheng Zhang}\ead{zczhang@bupt.edu.cn}
\author[1]{Xiaoli Liu}\ead{Liuxiaoli134@bupt.edu.cn}
\author[1]{Yonghao Dang}\ead{dyh2018@bupt.edu.cn}
\author[1]{Jianqin Yin \corref{cor1}}\ead{jqyin@bupt.edu.cn}
\affiliation[1]{organization={School of Artificial Intelligence, Beijing University of Posts and Telecommunications},
            city={Haidian},
            postcode={100876},
            state={Beijing},
            country={China}}
\cortext[cor1]{Corresponding author}

\begin{abstract}
Human motion prediction has achieved a brilliant performance with the help of convolution-based neural networks. However, currently, there is no work evaluating the potential risk in human motion prediction when facing adversarial attacks. The adversarial attack will encounter problems against human motion prediction in naturalness and data scale. To solve the problems above, we propose a new adversarial attack method that generates the worst-case perturbation by maximizing the human motion predictor’s prediction error with physical constraints. Specifically, we introduce a novel adaptable scheme that facilitates the attack to suit the scale of the target pose and two physical constraints to enhance the naturalness of the adversarial example. The evaluating experiments on three datasets show that the prediction errors of all target models are enlarged significantly, which means current convolution-based human motion prediction models are vulnerable to the proposed attack. Based on the experimental results, we provide insights on how to enhance the adversarial robustness of the human motion predictor and how to improve the adversarial attack against human motion prediction. The code is available at \href{https://github.com/ChengxuDuan/advHMP}{https://github.com/ChengxuDuan/advHMP}.
\end{abstract}


\end{frontmatter}

\section{Introduction} \label{sec:1}
Driven by deep neural networks(DNNs), recent works in human motion prediction(HMP) have achieved excellent performance \cite{mao2019learning,mao2020history,liu2020trajectorycnn,ma2022progressively}. Specifically, motion prediction forecasts the future motion of the target person according to the observed poses, which facilitates many applications such as intelligent security, autonomous driving, human-machine interaction, and so on. Human motion prediction is a fundamental ability that helps humans prepare in advance during interactions like catching. Similarly, precise motion prediction helps robot agents learn the future human posture and ensures the safety and smoothness of human-machine cooperation.\par
Among DNNs, convolutional neural networks(CNNs) are good at extracting structural features, so CNN-based predictors become the mainstream of human motion prediction recently \cite{li2018convolutional,mao2019learning,mao2020history,liu2020trajectorycnn,sofianos2021space,ma2022progressively}. Besides the ability of CNNs themselves, CNNs can also be supported by graph and attention mechanism, which gives convolution-based models more options to achieve better performance. In short, CNNs play an indispensable role in human motion prediction due to their excellent performance and flexibility.\par
However, recent research has proved that DNNs, including CNNs, are vulnerable to adversarial attacks \cite{szegedy2014intriguing, goodfellow2014explaining, madry2018towards,liu2019universal}. Adversarial attacks can be achieved by generating adversarial examples, a combination of clean samples and quasi-perceptible noises designed deliberately, eventually misleading the prediction of deep learning models. Since discovering the existence of adversarial examples, many works have stimulated the research of adversarial attacks \cite{szegedy2014intriguing,moosavi2016deepfool,madry2018towards,moosavi2017universal,carlini2017towards,brendel2018decisionbased,liu2019universal,su2019one,luo2022frequency,he2022transferable,goodfellow2014explaining,chen2021robustness, yufeng2023light, schneider2023dual, wang2022ab, lu2019switched,bryniarski2022evading}, which alarms algorithm designers about the safety problem in real scenarios. Currently, the scope of adversarial attacks has extended to other tasks \cite{cisse2017houdini,zhang2022adversarial,jain2019robustness,liu2020adversarial,wang2021understanding,diao2021basar,zheng2020towards, li2023ats}.\par 
Nevertheless, to the best of our knowledge, there is no related work that considers evaluating the adversarial robustness of human motion prediction models. Adversarial robustness indicates how much a model can resist the disturbing effect caused by adversarial attacks on the prediction results, which can only be evaluated by adversarial attacks. Human motion prediction is essential for robot agents to interact with humans correctly. A wrong prediction of the future human pose can hinder the robot agent from completing tasks or even mislead the agent to directly hit a human, which may cause danger when facing a high-power robot. For example, in industrial applications, robot arms may avoid workers’ bodies and finish their jobs simultaneously, which realizes that humans and robots work in the same space. However, robot arms under attack may predict workers’ motion in the wrong place, swing their arms vigorously, and cause injuries to workers.\par 
To deal with the problem above in analyzing the potential threat from adversarial attacks, we propose the \textit{first} white-box adversarial attack method against convolution-based human motion predictors to evaluate their adversarial robustness. To satisfy the requirement of an adversarial attack for harmfulness and naturalness, the proposed attack method combines a clean human motion sequence and a set of minor adversarial vectors pointing to the susceptible direction of each joint to maximize the prediction error meanwhile considering the naturalness of human skeletons. \par
\begin{figure}
    \centering
    \includegraphics[width=0.8\textwidth]{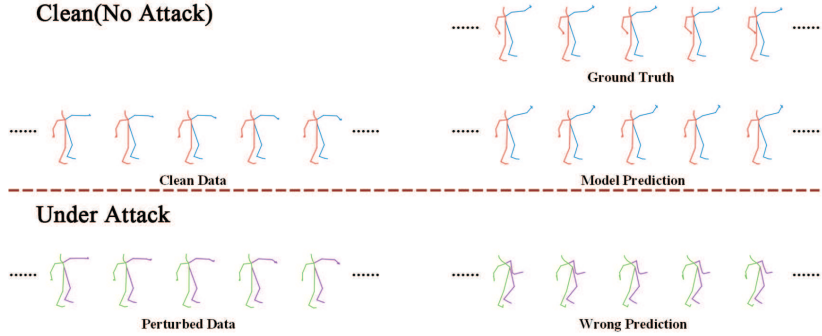}
    \caption{\label{fig:1}An example of the proposed attack against human motion prediction. The poses above the red line are the clean input and output of human motion prediction; the poses below the red line are the perturbed input and output of the prediction, which is obvious that the difference between the clean data and the perturbed data is imperceptible while the disturbance in prediction is significant.}
\end{figure}
Compared with adversarial attacks on image classification, attacking human motion prediction is highly challenging in two aspects. First, existing adversarial attack works can be used to disturb the motion prediction models, but the $L_2$ or $L_{\infty}$ norm constraint they deploy may cause physical unnaturalness in the human skeleton\cite{liu2020adversarial,wang2021understanding,diao2021basar,zheng2020towards}, which doesn’t meet the requirement of adversarial attacks for imperceptibility. To tackle this problem, we design two constraints for our attack method: temporal loss and bone length loss. Temporal loss aims to limit the change of joints’ speed and its derivatives between the clean poses and the adversarial poses, while bone length loss aims to avoid obvious deformation of each bone in the skeleton. Second, the coordinates of the human skeleton taken by different equipment are not uniform in scale. Thus, we design a practical scheme to help the proposed attack suit the scale of the target skeleton automatically, making it easy to attack different datasets.\par
In the proposed attack, the method first calculates the scale of the input sequence following the proposed scheme. Then, the method optimizes the adversarial perturbation by maximizing the prediction error to produce the attack effectiveness and minimizing the proposed loss to ensure the naturalness of the adversarial example. Finally, while the optimization process makes the adversarial perturbation more and more harmful, a hard constraint is needed to ensure the adversarial example is quasi-perceptible during the iterations. After each iteration, the method clips the perturbation by the $L_\infty$ norm constraint and gets the optimal adversarial perturbation when the maximum number of iterations is reached. \par
We evaluate the adversarial robustness of four different models\cite{mao2019learning,mao2020history,liu2020trajectorycnn,liu2022multi} on three datasets\cite{ionescu2013human3,von2018recovering,dang2021msr} by the proposed attack method. The attack effectiveness is shown in Figure \ref{fig:1}. The results show that the proposed attack is successful across datasets and models. Even the smallest perturbation can significantly enlarge the prediction error of all models by 22\% at least and 295\% at most, indicating the vulnerability of current human motion prediction algorithms. Comparing the results of each target model, we discover that prior knowledge and semantic information modeling are key to adversarial robustness. According to the qualitative results, the adversarial sample looks almost the same as the clean sample to human eyes, but it is relatively more noticeable when the motion information is animated. In general, the proposed attack successfully evaluates the adversarial robustness of the target models, which alarms that designers should pay attention to the adversarial robustness of human motion prediction.\par
Our main contributions are summarized as follows:\par
\begin{itemize}
    \item[1.] 
    To the best of our knowledge, we propose the \textit{first} adversarial attack method against human motion prediction models to evaluate their adversarial robustness. Inspired by \cite{wang2021understanding}, we constrain the perturbation optimization from both temporal feature and spatial feature, which enhances the imperceptibility of the adversarial example;
    \item[2.]
    We propose a novel adaptable scheme for the proposed attack method to suit the scale of the input motion sequence automatically, which solves the inconsistency of the data scales between different datasets and maximizes the attack effectiveness meanwhile ensuring the naturalness of the attack;
    \item[3.]
    We thoroughly evaluate the adversarial robustness of various prediction models under the proposed attack on different human motion datasets. The evaluation results show the vulnerability of the current models and reveal the keys to the adversarial robustness of human motion prediction
\end{itemize}

\section{Related Works} \label{sec:2}
\subsection{Convolution-based Human Motion Prediction} \label{sec:2.1}
With the brilliant ability to capture spatial and temporal features of human poses, CNNs are widely used in human motion prediction. The conventional CNN structure treats human motion sequences as a series of 2D data and extracts their feature effectively\cite{butepage2017deep,li2018convolutional,liu2020trajectorycnn,li2019efficient,tang2022temporal}. Butepage et al.\cite{butepage2017deep} utilized CNN to learn a generic representation that can be generalized to unseen future motions. Li et al.\cite{li2018convolutional} supported the RNN network by CNN to model spatial dependencies in human motion sequences. Li et al.\cite{li2019efficient} designed a convolutional hierarchical autoencoder framework to exploit the human body constraints efficiently. Liu et al.\cite{liu2020trajectorycnn} proposed a spatiotemporal feature learning CNN network, TrajectoryCNN, to capture the motion dynamics of the previous human motion sequence in the trajectory space and predict future human motion sequences in a non-recursive manner. However, conventional CNNs can capture the local features of human motion well but may neglect the global features.\par
As the skeleton-based human pose data are highly structured and can be presented by prior knowledge graphs, GCNs are popular in recent works, which can consider the local and global information of the human motion data at the same time. Mao et al.\cite{mao2019learning} first encoded the human skeleton as a graph structure and then captured spatial dependencies by GCN.\par
With the help of the graph, the GCN-based model can explore the inner relations of the human body\cite{li2020dynamic,cui2020learning,dang2021msr,aksan2019structured}. Li et al.\cite{li2020dynamic} designed a novel GCN containing a multi-scale graph to represent the human skeleton and used a graph-based GRU to predict future poses. Dang et al.\cite{dang2021msr} proposed a novel multi-scale residual graph convolution network whose ascending and descending blocks realized capturing fine-to-coarse and coarse-to-fine features.\par
Furthermore, human motion dynamics are also highly related to the temporal relationship, which developed into spatiotemporal relationships that need to be analyzed in GCN\cite{lebailly2020motion,cui2021towards,li2020multitask,zhong2022spatio,ma2022progressively,sofianos2021space}. Lebailly et al.\cite{lebailly2020motion} embedded human motion sequence into trajectories, then obtained residual motion prediction using GCN. Zhong et al.\cite{zhong2022spatio} adopted the gating network to enhance the generalization of GCN on human motion prediction. Sofianos et al.\cite{sofianos2021space} proposed a method that factorizes the graph adjacency matrix and extracts the interactions of joint-to-joint, time-to-time, and joint-to-time by GCNs.\par
What’s more, GCN can fuse with other networks to achieve improvement\cite{chao2020adversarial,li2021multiscale,mao2020history,mao2021multi}. Chao et al.\cite{chao2020adversarial} combined GCN structure and adversarial training to refine predicted human motion. Li et al.\cite{li2021multiscale} designed a multiscale spatiotemporal graph computing unit with GCN as the core part of the encoder and a graph-based GRU as the decoder. Wei et al.\cite{mao2020history} applied an attention mechanism to GCN structure, which unifies the short-term and long-term prediction and enhances the ability of the network to make use of motion repetitiveness. However, the relation described by the adjacent matrix may be rigid, which makes the network hard to understand the dynamic relationships between the joints in the human skeleton.\par
Although the above works achieve excellent performance, they don’t consider the adversarial robustness of motion prediction networks, which is not safe to deploy these models into real scenarios. In this paper, we focus on the adversarial robustness of current human motion prediction algorithms.\par
\subsection{Other Human Motion Prediction}\label{sec:2.2}
Besides convolution-based methods, human motion prediction can be completed by other approaches, including RNN-based, Transformer-based, MLP-based, and Generative model methods.\par
\textbf{RNN-based methods} Because human motion prediction has the serialized characteristic, most motion prediction works are based on RNN structure. Therefore, the potential of the RNN structures was exploited\cite{fragkiadaki2015recurrent, martinez2017on, wolter2018complex,ghosh2017learning,gui2018adversarial}. Martinez et al.\cite{martinez2017on} modeled the velocities of human motion sequences by a residual architecture and simplified the network by GRU. Since human motion prediction is highly related to the dynamic of the human body, the researchers focused on how to model the relation of each joint in the human skeleton\cite{yao2018multiple,liu2019towards,pavllo2020modeling}. Pavllo et al.\cite{pavllo2020modeling} investigated the quaternion-based representation to enhance the network in understanding the kinematics of the motion sequence. Because it was found that the ability of RNN itself to capture the temporal information is not enough for human motion prediction, temporal modeling was studied\cite{guo2019human, chiu2019action,shu2021spatiotemporal,gopalakrishnan2019neural,corona2020context}. Chiu et al.\cite{chiu2019action} proposed a hierarchical multi-scale RNN for human motion forecasting to capture the temporal dependencies with different temporal scales hierarchically.\par
\textbf{Transformer-based methods} With the ability to handle global data in parallel, Transformer\cite{dong2021dual,vaswani2017attention} can be adopted to solve the problem of human motion prediction\cite{aksan2021spatio,cai2020learning,martinez2021pose}. Aksan et al.\cite{aksan2021spatio} designed a Transformer-based architecture that explicitly learns intra- and inter-joint dependencies via its decoupled temporal and spatial attention blocks for human motion prediction. Martínez et al.\cite{martinez2021pose} proposed a non-autoregressive Transformer that can decode predictions in parallel from a query sequence generated in advance with elements from the input sequence.\par
\textbf{MLP-based methods} Recently, the MLP has been deployed for human motion prediction for its simpleness\cite{guo2023back, Bouazizi2022MotionMixer}. Bouazizi et al.\cite{Bouazizi2022MotionMixer} utilized MLPs to learn mixing features across the spatial and temporal domains, and designed a squeeze-and-excitation (SE) block to calibrate the influence of each time step in the pose sequence. Guo et al.\cite{guo2023back} only utilized fully connected layers, layer normalization, and transpose operations to predict future human motion, which achieved excellent performance with fewer parameters.\par
\textbf{Generative model methods} As human motion prediction can be regarded as generative tasks like inpainting, generative models were adopted in human motion prediction, such as VAE\cite{komura2017recurrent,cai2021unified,zhang2021we}, GAN\cite{barsoum2018hp,hernandez2019human}, normalizing flow\cite{yuan2020dlow}, and diffusion model\cite{barquero2023belfusion,wei2023human}. Cai et al.\cite{cai2021unified} designed a CVAE-based unified framework covering the tasks of motion prediction, completion, interpolation, and spatial-temporal recovery. Barsoum et al.\cite{barsoum2018hp} proposed a GAN-based network to predict multiple plausible future human motions from an observed human motion sequence. Yuan et al.\cite{yuan2020dlow} designed a novel sampling method for deep generative models to obtain a diverse set of future human motions. Barquero et al.\cite{barquero2023belfusion} proposed a latent diffusion model that exploits a behavioral latent space to render human motion predictions more realistic, accurate, and context-adaptive.\par
Although the methods above for human motion prediction achieve brilliant performance, the main target in this paper is the convolution-based method. Therefore, the attack effectiveness against these methods will only be discussed in Section \ref{sec:7.3}.\par

\subsection{Adversarial Attack} \label{sec:2.3}
Since Szegedy et al.\cite{szegedy2014intriguing} found that a series of quasi-perceptible perturbations can be designed to mislead deep learning network malfunctioning, plenty of works have proved the vulnerability of DNN in image classification\cite{szegedy2014intriguing,moosavi2016deepfool,madry2018towards,moosavi2017universal,carlini2017towards,brendel2018decisionbased,su2019one,liu2019universal,luo2022frequency,he2022transferable,goodfellow2014explaining, yufeng2023light, schneider2023dual, wang2022ab, lu2019switched,bryniarski2022evading}.\par
Among previous works of adversarial attack, optimization-based attack methods\cite{szegedy2014intriguing, goodfellow2014explaining,madry2018towards,carlini2017towards,luo2022frequency,he2022transferable,lu2019switched,bryniarski2022evading}, which generate noises according to the gradient of the loss function or the score function corresponding to the network, can be easily generalized to perturb most of the DNNs. For instance, Fast Gradient Sign Method (FGSM)\cite{goodfellow2014explaining} directs the adversarial perturbation by the gradient of input images and the sign function and constrains its magnitude by just one parameter, which realizes a one-step attack. Projected Gradient Descent (PGD)\cite{madry2018towards,lu2019switched,bryniarski2022evading} generates perturbations by accumulating gradients from each iteration perturbed images' backpropagation, which is proved to be the strongest attack utilizing the local first-order information about the network\cite{madry2018towards}. Meanwhile, optimization-based methods spread to other tasks like semantic segmentation\cite{cisse2017houdini}, reinforcement learning\cite{li2023ats}, autonomous driving\cite{zhang2022adversarial}, and pose estimation\cite{jain2019robustness,chen2021robustness} for evaluating the adversarial robustness of corresponding models.\par
The adversarial attack has involved human pose space in action recognition\cite{liu2020adversarial,wang2021understanding,diao2021basar,zheng2020towards} and shown that perturbations against skeleton-based data need to be restricted physically. Liu et al.\cite{liu2020adversarial} proposed the first adversarial attack on deep skeleton action recognition, which perturbs the input data from a spatiotemporal perspective. Wang et al.\cite{wang2021understanding} designed a new perceptual loss to constrain the perturbation, analyzed the adversarial robustness of action recognition models, and showed the difference between attacking images and skeletons: Unlike image data, human skeleton data are highly related to physical constraints and have much fewer Degrees of Freedom (DoFs) than image data, making human skeleton data low redundant and perceptual sensitive\cite{wang2021understanding,diao2021basar}. Diao et al.\cite{diao2021basar} proposed the first black-box attack on action recognition, which misleads the data to leave the correct classification distribution in the shortest distance and applies inverse kinematics to make the skeleton natural visually.\par
However, to the best of our knowledge, no existing work has studied the adversarial robustness of human motion prediction, which may cause fatal mistakes in human-machine interaction. For this reason, we propose an adversarial attack method against motion prediction to evaluate its potential problem.\par

\section{Problem Formulation} \label{sec:3}
\subsection{Human Motion Prediction} \label{sec:3.1}
In this work, we focus on human motion prediction, which is executed to forecast a future human motion sequence according to the corresponding past human motion sequence. First, we denote  $s_i$ as a true pose composed of $J$ joint points at time $i$, and each joint point has $D$ dimensions. Then we can denote $X=\left \{ s_1,s_2,...,s_{T_h} \right \}, X\in \mathbb R^{T_h\times J\times D}$ as an observed motion sequence of length $T_h$ and $Y=\left \{ s_{T_h+1},s_{T_h+2},...,s_{T_h+T_f} \right \}, Y\in \mathbb R^{T_f\times J\times D}$ as the future ground truth motion sequence of length $T_f$. This work focuses on Cartesian coordinates; therefore, $D$ equals 3. The prediction algorithm inputs the history human motion sequence $X$ and generates the predicted human motion sequence $P=\left \{\hat{s} _{T_h+1},\hat{s}_{T_h+2},...,\hat{s}_{T_h+T_f} \right \}, P\in \mathbb R^{T_f\times J\times D}$ where $\hat{s}_i$ is the predicted pose at time $i$. During training, the sequence $P$ is optimized to approximate the distribution of the future ground-truth motion sequence $Y$. Therefore, the prediction algorithms can output generally precise future poses.
\subsection{Attack Model} \label{sec:3.2}
In this paper, we focus on the scenario where human joints’ locations can be directly acquired by motion capture (MoCap) techniques. In this setting, human joints are recognized as rotation angles or 3D coordinates, which can be connected as a graph known as a skeleton. Based on human motion sequences, prediction algorithms model the spatiotemporal feature of the target and construct the motion dynamics of the target movement. On the contrary, the adversarial attack aims to disrupt the normal operation of the model without being perceived by human users. Thus, the adversary needs to perturb the history human motion sequence so that the prediction algorithm will mislead the motion in an utterly wrong direction, which may disturb the user’s prediction of the future human motion sequence. Such mistakes can cause users or robots to take wrong actions, which may lead to fatal damage.\par
Our goal is to analyze the adversarial robustness of current human motion prediction algorithms by an adversarial attack that can find worst-case perturbation against the algorithm while keeping the adversarial examples natural. Therefore, the problem can be described as follows: given a history human motion sequence $X$ with its future ground truth $Y$, the proposed attack aims to find $X'$ that is visually similar to $X$ and can fool the white-boxed predictor such that the error between the predicted sequence $P$ and $Y$ is maximized.\par
Based on optimization-based methods, the proposed attack formulates the problem as a constrained optimization problem, which can be solved by gradient descent. Because human pose data are low redundant and perceptual sensitive \cite{wang2021understanding,diao2021basar}, the adversarial perturbation for human motion sequences needs to be bounded by the $L_{\infty}$ norm and constrained by physical laws. The physical constraints in the proposed attack are formulated as two loss functions to constrain the spatial feature of the pose and the temporal feature of the pose, respectively. Therefore, the solution of the proposed perturbation can be defined as:\par
\begin{equation} \label{eq:1} 
    \begin{aligned}
    &\max_{X'} L_{pred}\left ( f\left ( X' \right ) ,Y \right ) -\lambda \cdot L_{phy}\left ( X,X' \right )\\
    &s.t.\left \| X'-X \right \|_\infty\in \left [ 0,\varepsilon  \right ]  
    \end{aligned}
\end{equation}
where $L_{pred}\left ( \cdot  \right ) $ denotes the prediction error function of human motion prediction, $L_{phy}\left ( \cdot  \right ) $ denotes the integrated physical constraint function,  $f\left ( \cdot  \right ) $ denotes the target human motion prediction model, X’ denotes the adversarial example, $\lambda$ is a trade-off parameter between attack effectiveness and naturalness, and $\epsilon$ is the boundary of the adversarial perturbation.\par

\section{Methodology} \label{sec:4}
\subsection{Optimization for Attack} \label{sec:4.1}
The basic objective of our attack is to deviate the prediction poses away from the true future poses; the further, the better. The proposed attack method generates adversarial samples by adding a minor perturbation to the original history trajectories. From the perspective of human posture, the perturbation, which can be regarded as a series of vectors, changes the 3D coordinates of each joint in human skeleton data and further sabotages the spatiotemporal features of the history motion sequence, which causes an utterly wrong prediction. In a realistic scenario, because the risk will accumulate as time passes, and the attack will work better when it has more parameters to perturb, the proposed attack will target the whole history sequence as the input data.\par
According to Equation (\ref{eq:1}), the attack effectiveness is based on maximizing the prediction error of the human motion prediction model, which indicates $L_{pred}$. Specifically, the basic optimization goal to generate attack effectiveness is the negation of the average prediction error of each joint in all considered time frames. For a single motion sequence, the $L_{pred}$ is formulated as:\par
\begin{equation} \label{eq:2} 
    \begin{aligned}
    &L_{pred}\left ( P,Y\right ) =\frac{1}{T_f\cdot J} \sum_{i=1}^{T_f} \sum_{j=1}^{J} \left \| \hat{s}_{T_h+i,j}-s_{T_h+i,j}  \right \|_2   
    \end{aligned}
\end{equation}
where $\hat{s}_{T_h+i,j}$ and $s_{T_h+i,j}$  denote the $j$ joint coordinate in frame $i$ respectively from the prediction motion sequence $P$ and the ground truth $Y$, which are defined in \ref{sec:3.1}. $T_h$ and $T_f$ are the length of the history motion sequence and the future motion sequence, respectively. $J$ is the number of joints of the human body. For a set of motion sequences, the loss is the average over the $L_{pred}$ of all motion sequences, which means the proposed attack perturbs all the motion sequences simultaneously.\par
\subsection{Constraints for Perceptual Naturalness} \label{sec:4.2}
As mentioned in \ref{sec:3.2}, human motion sequence data must be physically natural. As a result, we design a scheme and two constraints to enforce the adversarial sequence obeying the physical law. The physical law we consider for the proposed attack can be divided into two theories: Newton's Second Law for temporal features and the rigid body kinematics for spatial features.\par 
When analyzing human motion, the human body can be regarded as a particle system where the particles are called joints. Following Newton's Second Law, the magnitude of the acceleration obtained by a particle under force is directly proportional to the magnitude of the applied force. As the force applied to the particle won't make a sudden change, the velocity and acceleration of the joint should change smoothly. Therefore, the proposed attack shouldn't modify the velocity and acceleration of each joint by a large margin.\par
In rigid body kinematics, the human body can be connected by the joints orderly and composed of parts named bones. a bone can be regarded as a rigid body, as the length change of the bone is little, which is also applicable to human motion. In this case, the proposed attack shouldn't affect the length of each bone substantially. \par
Based on the physical laws mentioned above, we propose an adaptable scheme and two physical constraints for the temporal and spatial features.\par
\textbf{Adaptable Scheme} According to our observation, the scale of the skeleton data varies across different datasets, so it is not a good idea to use constant parameters to manipulate the vectors in adversarial perturbation. To solve this problem, we propose an adjustable scale function $S_c$ for each batch:\par
\begin{equation} \label{eq:3} 
    \begin{aligned}
    &f^x_{span}\left ( X \right ) =\min {\left ( \max\left ( X^x \right )-\min\left ( X^x \right )   \right ) }\\
    &S_c\left ( X \right ) =\min {\left ( f^x_{span}\left ( X \right ) ,f^y_{span}\left ( X \right ),f^z_{span}\left ( X \right ) \right ) }
    \end{aligned}
\end{equation}
where $X$ denotes the history human motion sequence; $x$, $y$, and $z$ denote the dimension that the function needs to measure, and $X^x$ denotes all the $x$-coordinate in the history motion sequence, similar to $X^y$ and $X^z$; $\min {\left ( \cdot  \right ) }$ and $\max {\left ( \cdot  \right ) }$ denote picking the minimum and maximum of the sequence, respectively.\par
In the proposed scheme, the scale function first finds a basic unit, and then the proposed attack utilizes this unit to perform operations like update and clip. In this scheme, we assume there is a bounding box that can just fit the target pose. Then, the scale function will calculate the minimum side lengths on each axis and choose the shortest one as the basic unit to perform the adversarial attack. As a result, the proposed scheme can help the attack adapt to each input sequence and facilitate users setting a set of unified parameters for the attack.\par
\textbf{Physical Constraints} To constrain the physical characteristic of the adversarial sample, the optimization needs to consider the temporal features of each joint and spatial features between specified joints. The spatiotemporal feature needs to be constrained by $L_{phy}$ in Equation(\ref{eq:1}) composed of a temporal constraint $L_{temp}$ and a spatial constraint $L_{BL}$.\par
For the temporal feature, inspired by \cite{wang2021understanding}, the speed of each joint and its derivatives shouldn't change significantly between each frame. According to this hypothesis, the attack minimizes the following loss function to constrain the perturbation:\par
\begin{equation} \label{eq:4} 
    \begin{aligned}
    &L_{temp}\left (  X,X'\right )=\frac{1}{N} \sum_{n=1}^{N}\left \| X^{(n)}-X'^{(n)} \right \|_2  
    \end{aligned}
\end{equation}
where $X$ denotes the history motion sequence while $X'$ denotes the adversarial motion sequence. $X^{(n)}$ denotes the $n$-th derivative of the sequence $X$. $N$ denotes the number of differential orders. In this work, we only consider an intuitive way to constrain each joint's velocity and acceleration, so we set $N$ as 2. During the optimization, the adversarial perturbation needs to minimize $L_{temp}$ so that the joints will not be moved too randomly and look like the target is shaking.\par
As for the spatial feature, bone length is a sensitive attribute. Without constraining the bone length, the limbs in the adversarial sequence will be elastic, which can be detected by human eyes easily. Therefore, the proposed attack calculates the length of each bone in every pose and minimizes the error between the adversarial and the clean. We apply the bone length loss function from \cite{wang2021understanding} to constrain the motion:\par
\begin{equation} \label{eq:5} 
    \begin{aligned}
    &L_{BL}\left ( X,X',C \right ) = BL\left ( X,C \right ) - BL\left ( X',C \right )\\
    &BL\left ( X,C \right ) =\sum_{t=1}^{T_h}\sum_{l=1}^{L_C}  \left \|{s}_{t,C_{l,0}}-s_{t,C_{l,1}}\right \| _2
    \end{aligned}
\end{equation}
where $C$ denotes the connectivity dictionary of the target pose data, and it contains how to connect the joints into a skeleton; $L_C$ denotes the length of $C$, which is the total number of the bones in the target pose; $C_{l,0}$ and $C_{l,1}$ denotes the joints' indexes at both ends of the $l$-th bone. By minimizing $L_{BL}$, the influence of adversarial perturbations on poses' bones can be largely decreased, enhancing the attack's imperceptibility. In conclusion, the goal of our attack method is to maximize the following loss:\par
\begin{equation} \label{eq:6} 
    \begin{aligned}
    &L=L_{pred}-\lambda \cdot L_{phy}\\
    &L_{phy}=L_{temp}+L_{BL}
    \end{aligned}
\end{equation}
where $\lambda$ denotes the trade-off parameter between the attack effectiveness and perceptual naturalness. After the scale $S_c(X)$ and the integrated loss $L$ are calculated, we can apply the proposed scheme to update the adversarial motion sequence by the following equation:\par
\begin{equation} \label{eq:7} 
    \begin{aligned}
    &X'=Clip_{X,\epsilon \cdot S_c(X)}\left \{ X'+ \epsilon _{step}\cdot S_c(X)\cdot sign(\nabla L_{X'}) \right \} \\
    &\left \| X'-X \right \| _\infty \le \epsilon \cdot S_c(X)
    \end{aligned}
\end{equation}
where $X'$ denotes the perturbed history sequence; $\epsilon$ denotes the parameter that bounds the perturbation. $Clip_{X,\epsilon \cdot S_c(X)}$ function clips deviations in each axis so that the absolute distances of the adversarial example in each axis will not exceed $\epsilon \cdot S_c(X)$ away from the clean example $X$. $\epsilon _{step}$ denotes the step-size parameter. $\nabla L_{X'}$ computes the gradient of the integrated loss function $L$ around the current adversarial example $X'$, and $sign(\cdot)$ is the sign function. With the help of the scale function, we enhance perturbation imperceptibility in a traditional $L_\infty$-norm way.\par
\subsection{Generation of Perturbations} \label{sec:4.3}
\begin{algorithm}
\caption{\label{alg:1} Overview of the proposed attack scheme} 
    \begin{algorithmic}[1]
    \Require history sequence $X$, future sequence $Y$, target model $f(\cdot)$, boundary $\epsilon$, iteration number $N$, step size $\epsilon_{step}$
    \Ensure perturbed sequence $X'$
    \State \textbf{Initialization}: Calculate the scale of X by Equation (\ref{eq:3}) Initialize perturbation $S_{pert}$ randomly under $\epsilon \cdot S_c(X)$
    \For  {$i = 1, 2, ..., N$}
    \State $X' = X + S_{pert}$
    \State Turn $X'$ into the form the model needs
    \State $P = f(X')$
    \State Calculate $L_{pred}$, $L_{temp}$, $L_{BL}$ by Equation(\ref{eq:2})(\ref{eq:4})(\ref{eq:5})
    \State $L=L_{pred}-\lambda \cdot \left ( L_{temp}+L_{BL} \right )$
    \State $X'=Clip_{X,\epsilon \cdot S_c(X)}\left \{ X'+ \epsilon _{step}\cdot S_c(X)\cdot sign(\nabla L_{X'}) \right \}$
    \EndFor
    \end{algorithmic}
\end{algorithm}
Our optimization method is based on Projected Gradient Descent (PGD)\cite{madry2018towards}. The process is summarized in Algorithm \ref{alg:1}.\par
Following the proposed scheme, The perturbation is initialized randomly, and then the input scale is calculated to ensure the size of the poses. We first add the current perturbation to each iteration's clean history human motion sequence. Next, the model predicts a human motion sequence based on the perturbed input sequence. Then, the algorithm calculates $L$ based on Equation (\ref{eq:7}) and updates the perturbation by gradient descent under the control of the scale function. Finally, the algorithm generates the adversarial perturbation, which can mislead motion prediction to a series of wrong poses.\par

\section{Experiment} \label{sec:5}
\subsection{Experimental Settings} \label{sec:5.1}
In this section, we will introduce the datasets(\ref{sec:5.1.1}), the target baselines(\ref{sec:5.1.2}), the evaluation metrics(\ref{sec:5.1.3}), and the implementing details(\ref{sec:5.1.4}). The code is open-sourced \href{https://github.com/ChengxuDuan/advHMP}{here}
\subsubsection{Datasets} \label{sec:5.1.1}
In the experiment, we choose Human3.6M\cite{ionescu2013human3}(H3.6M), CMU-Mocap(CMU)\footnote{Available at \href{http://mocap.cs.cmu.edu/}{http://mocap.cs.cmu.edu/}}, and 3D
Pose in the Wild dataset\cite{von2018recovering}(3DPW) as the evaluation dataset. Human3.6M is the most used large dataset for human motion prediction. CMU contains more sports activities than Human3.6M, which can help us evaluate the influence of the activities on the adversarial robustness. 3DPW includes various indoor and outdoor activities, which can be used to simulate the real scenario the model needs to face.\par
\textbf{Human3.6M(H3.6M)} has collected 15 types of actions from 7 actors(S1, S5, S6, S7, S8, S9 and S11). The original skeleton in Human3.6M has 32 joints presented by exponential maps. We convert the presentation to 3D coordinates and remove ten redundant joints. The global rotations and translations of poses are excluded. We downsample the frame rate from 50fps to 25fps. S5 and S11 are used for testing and validation, respectively, while other subjects are used for training. In order to compare the results conveniently, we take eight random samples per action for testing.\par
\textbf{CMU-Mocap(CMU)} holds eight categories of actions, with 38 joints presented by exponential maps. We convert the presentation to 3D coordinates as well. The global rotations and translations of poses are excluded, too. Following the set in \cite{mao2019learning}, 25 joints in each pose are kept, while the others are excluded. The division of training and testing datasets is also the same as \cite{mao2019learning}. Like H3.6M, CMU is also sampled into eight random samples per action for testing.\par
\textbf{The 3D Pose in the Wild(3DPW)} contains human motion captured from indoor and outdoor scenes, which are challenging for algorithms. The poses in the dataset are presented in 3D space at 30fps. Each pose contains 24 joints, and 23 joints are used.\par
\subsubsection{Baselines} \label{sec:5.1.2}
To evaluate the adversarial robustness of the convolution-based models thoroughly, we need to test the CNN-based model, GCN-based model, and their derivative methods by our attack method. Therefore, we choose LearningTrajectoryDependency \cite{mao2019learning} (LTD), TrajectoryCNN \cite{liu2020trajectorycnn} (TrajCNN), HistoryRepeatItself \cite{mao2020history} (HRI) and Multi-Head TrajectoryCNN \cite{liu2022multi}(MH TrajCNN).\par
\textbf{LearningTrajectoryDependency} is a representative GCN-based algorithm. LTD first transforms pose information by Discrete Cosine Transform (DCT) to gain a more compact representation and then encodes the spatial structure of human pose by GCNs. Therefore, we need to apply DCT to adversarial samples to make them feasible for LTD. \par
\textbf{TrajectoryCNN} represents a conventional CNN-based baseline that achieves excellent performance. TrajectoryCNN can simultaneously achieve trajectory space transformation, model motion dynamics, and predict future poses by combining the skeletal representation.\par
\textbf{HistoryRepeatItself} enhances the GCN-based structure by utilizing the motion attention mechanism we want to investigate, which can exploit motions instead of static frames to better leverage historical information for motion prediction. \par
\textbf{Multi-Head TrajectoryCNN} can complete the motion prediction task and action recognition task at the same time, which can show whether the proposed attack influences the semantic information of the predicted human motion sequence. Conversely, explicit semantic information modeled by the classification head of Multi-Head TrajectoryCNN may also influence the effectiveness of the proposed attack, which can help us study whether the semantic information is related to the adversarial robustness of human motion prediction models.\par
Because the target model needs to be a white box, we train our model to be the victim if the official model is not open-sourced, which may cause a difference in the result when we input a clean sequence.\par
\subsubsection{Evaluation Metrics} \label{sec:5.1.3}
To evaluate the adversarial robustness, we need to calculate how much the predicted motion sequence $P$ is away from the ground truth $Y$. As the result of predicting poses in Cartesian coordinate, we use Mean Per Joint Position Error (MPJPE), which is defined in Equation (\ref{eq:8}), to evaluate how much the prediction deviated from the ground truth.\par
\begin{equation} \label{eq:8} 
    \begin{aligned}
    &E_{MPJPE}\left ( P,Y,t\right )=\frac{1}{J}  \sum_{j=1}^{J} \left \| \hat{s}_{t,j}-s_{t,j}  \right \|_2   
    \end{aligned}
\end{equation}
where $t$ denotes the specific frame of the analyzed time interval. For example, in a 25fps human motion sequence, the fourth frame of its predicted human motion sequence corresponds to the time interval of 160ms. For a set of sequences, the error is the average of the MPJPEs of all sequences at the corresponding frame. For H3.6M and CMU, we calculate the average MPJPE of sequences for all activities as their corresponding results and the average result of MPJPEs of all the activities as the average prediction error.\par
In this work, we are curious how much the adversarial samples against motion prediction can also influence the semantic information of the predicted human motion sequence for action recognition. Therefore, we applied Multi-Head TrajectoryCNN\cite{liu2022multi}, which can complete motion prediction and action recognition at the same time, to verify this hypothesis. When the attack perturbs the motion prediction branch of Multi-Head TrajectoryCNN, we can observe how much the proposed attack influences the performance of the action recognition branch, which indicates whether the proposed attack can be easily detected by normal action recognition. In action recognition, accuracy is used to evaluate the algorithm's performance; hence, we use Attack Success Rate(ASR) to evaluate the attack effectiveness against action recognition. ASR is defined as:\par
\begin{equation} \label{eq:9} 
    \begin{aligned}
    &ASR=\frac{N_{adv}}{N_{right}}  
    \end{aligned}
\end{equation}
where $N_{right}$ is the number of clean samples classified correctly, and $N_{adv}$ is the number of adversarial samples misclassified but classified correctly when clean.\par
\subsubsection{Implementing Details} \label{sec:5.1.4}
In our PGD-based attack method, the bound $\epsilon$ is set from 0.01 to 0.05 with an interval of 0.01, while the step size $\epsilon_{step}$ is 0.1 times $\epsilon$. The iteration number is set to 50. The trade-off parameter $\lambda$ is set to 0.5. Finally, we attack the whole history human motion sequence in this experiment.\par
In this work, we train short-term and long-term models separately for LTD and TrajectoryCNN. All the input lengths are set to 10 frames, while their output lengths are set to 10 for short-term prediction and 25 for long-term prediction in H3.6M and CMU. In 3DPW, the input length is the same as others' settings, but the output length is set to 15 for short-term and 30 for long-term predictions. \par
As HRI can effectively make use of motion repetitiveness in long-term history, its input length is set to 50 frames, and its output length is set to 25 frames in H3.6M and CMU or 30 frames in 3DPW. The attention mechanism in HRI allows it to train a unified model for both short-term and long-term prediction. Thus, the HRI model we use will predict ten frames per iteration, three iterations in total, and cut off the prediction sequence. Because our experiment's datasets differ from Multi-Head TrajectoryCNN's original setting, we label the motions by the activities in the corresponding dataset and modify Multi-Head TrajectoryCNN's input into 50 frames and output length into 10 for short-term and 25 frames for long-term to facilitate the comparison.\par
However, the input length may influence the adversarial robustness of the target models. Therefore, to compare fairly and investigate the impact of the input length, we train LTD and TrajectoryCNN under the same set above, but the input length is changed to 50 frames. We denote them as LTD-50 and TrajCNN-50, respectively. Note that LTD-50 and TrajCNN-50 both contain a short-term and long-term model, which is for a fair comparison with the original models whose input lengths are 10.\par
\subsection{Experimental Results} \label{sec:5.2}
\begin{table}[htbp]
\centering
\caption{\label{tab:1}Average prediction errors of target models on H3.6M} before and after perturbation in variations of the boundary for the models whose input length is 10. The \textbf{bold} results are the worst-case of each time interval and the percentage is the least and the most growth rate from the clean error. 
\scalebox{0.5}{
\begin{tabular}{c|cccccc}
\hline
Intervals(ms) & 80  & 160 & 320 & 400 & 560 & 1000\\ \hline
LTD  & 12.2 & 25.0 & 51.0 & 61.3 & 78.5 & 114.3 \\
$\epsilon$=0.01        & \begin{tabular}[c]{@{}c@{}}35.5\\ (191.0\%↑)\end{tabular}          & \begin{tabular}[c]{@{}c@{}}73.9\\ (195.6\%↑)\end{tabular}           & \begin{tabular}[c]{@{}c@{}}137.4\\ (169.4\%↑)\end{tabular}          & \begin{tabular}[c]{@{}c@{}}154.6\\ (152.2\%↑)\end{tabular}          & \begin{tabular}[c]{@{}c@{}}184.7\\ (135.3\%↑)\end{tabular}          & \begin{tabular}[c]{@{}c@{}}210.5\\ (84.2\%↑)\end{tabular}           \\
$\epsilon$=0.02 & 45.9 & 93.0 & 168.2 & 187.2 & 222.1 & 241.5 \\
$\epsilon$=0.03 & 53.8 & 107.6 & 193.3 & 214.0 & 236.2 & 255.1 \\
$\epsilon$=0.04 & 56.9 & 111.2 & 194.7 & 213.9 & 248.3 & 264.4 \\
$\epsilon$=0.05        & \textbf{\begin{tabular}[c]{@{}c@{}}60.3\\ (394.3\%↑)\end{tabular}} & \textbf{\begin{tabular}[c]{@{}c@{}}115.7\\ (362.8\%↑)\end{tabular}} & \textbf{\begin{tabular}[c]{@{}c@{}}201.2\\ (294.5\%↑)\end{tabular}} & \textbf{\begin{tabular}[c]{@{}c@{}}220.6\\ (259.9\%↑)\end{tabular}} & \textbf{\begin{tabular}[c]{@{}c@{}}255.6\\ (225.6\%↑)\end{tabular}} & \textbf{\begin{tabular}[c]{@{}c@{}}266.5\\ (133.2\%↑)\end{tabular}} \\ \hline
TrajCNN & 10.3 & 23.6 & 50.3 & 60.6 & 78.0 & 110.6 \\
$\epsilon$=0.01        & \begin{tabular}[c]{@{}c@{}}40.3\\ (291.3\%↑)\end{tabular}          & \begin{tabular}[c]{@{}c@{}}93.3\\ (295.3\%↑)\end{tabular}           & \begin{tabular}[c]{@{}c@{}}187.2\\ (272.2\%↑)\end{tabular}          & \begin{tabular}[c]{@{}c@{}}208.0\\ (243.2\%↑)\end{tabular}          & \begin{tabular}[c]{@{}c@{}}199.8\\ (156.2\%↑)\end{tabular}          & \begin{tabular}[c]{@{}c@{}}233.8\\ (111.4\%↑)\end{tabular}          \\
$\epsilon$=0.02 & 58.8 & 131.8 & 255.7 & 280.0 & 255.8 & 283.0 \\
$\epsilon$=0.03 & 68.6 & 149.7 & 283.7 & 308.8 & 286.8 & 307.1 \\
$\epsilon$=0.04 & 72.9 & 156.3 & 292.6 & 317.7 & 290.8 & 306.0 \\
$\epsilon$=0.05        & \textbf{\begin{tabular}[c]{@{}c@{}}77.3\\ (650.5\%↑)\end{tabular}} & \textbf{\begin{tabular}[c]{@{}c@{}}163.7\\ (593.6\%↑)\end{tabular}} & \textbf{\begin{tabular}[c]{@{}c@{}}303.5\\ (503.4\%↑)\end{tabular}} & \textbf{\begin{tabular}[c]{@{}c@{}}328.1\\ (441.4\%↑)\end{tabular}} & \textbf{\begin{tabular}[c]{@{}c@{}}295.8\\ (279.2\%↑)\end{tabular}} & \textbf{\begin{tabular}[c]{@{}c@{}}313.9\\ (183.8\%↑)\end{tabular}} \\ \hline
\end{tabular}}
\end{table}
\textbf{Results on Human3.6M.} Table \ref{tab:1} and \ref{tab:2} shows the quantitative comparisons of short-term and long-term prediction on H3.6M between the above four networks under the attacks of five different intensities (bounds). We bold the highest error in each time interval of each algorithm and calculate the growth rate from the clean result to the weakest-case and strongest-case result for the convenience of comparison. Due to space limits, detailed tables are provided in \ref{app:1}.\par
In all cases, even the proposed attack with the minimal $\epsilon$ achieves significant disturbance (increasing errors by at least 31.2\%) to the prediction, proving the attack's effectiveness and the current models' vulnerability. These results mean the proposed attack can find the worst-case adversarial examples for each model, which can thoroughly analyze the adversarial robustness of the target models. However, the attack effectiveness will be weakened as the time interval grows. This weakening phenomenon may be attributed to the limited space where poses are set and the excellent comprehension of the target models to the human motion, which doesn't allow the predicted motion to perform impossible activities.\par
The model with a lower growth rate of prediction errors is more adversarial robust, which means the model can maintain its normal prediction ability and resist attack effectiveness. Therefore, LTD-50 is the most adversarial robust, then LTD, MH TrajCNN, HRI, TrajCNN, and finally TrajCNN-50 is the most vulnerable. These results show that the excellent performance of the human motion prediction algorithm in the clean circumstance is not related to its adversarial robustness.\par
In light of our observation, prior knowledge and semantic information are key to the adversarial robustness of human motion prediction. According to the ranking above, the GCN-based structure is more adversarial robust than the CNN structure. The adjacent matrices in GCN provide prior knowledge of human pose and constrain the model during the inference stage, which can suppress the disturbance from the proposed attack. Thus, the adversarial robustness of LTD comes out in front. Without constraints like graphs, the CNN-based structure has little resistance against the adversarial attack. Nevertheless, the semantic information modeled explicitly in MH TrajCNN enhances its adversarial robustness. MH TrajCNN can keep the error low when facing weak attacks but can't withstand intense attacks, which still enlarge the errors.\par
\begin{table}[htbp]
\centering
\caption{\label{tab:2}Average prediction errors of target models on H3.6M} before and after perturbation in boundary variations for the models whose input length is 50. The \textbf{bold} results are the worst-case of each time interval, and the percentage is the least and the most growth rate from the clean error.
\scalebox{0.5}{
\begin{tabular}{c|cccccc}
\hline
{\ul Intervals(ms)} & 80 & 160 & 320 & 400 & 560 & 1000 \\ \hline
LTD-50 & 13.8 & 27.8 & 54.3 & 66.7 & 87.4 & 116.7 \\
$\epsilon$=0.01              & \begin{tabular}[c]{@{}c@{}}26.7\\ (93.5\%↑)\end{tabular}           & \begin{tabular}[c]{@{}c@{}}50.5\\ (81.7\%↑)\end{tabular}            & \begin{tabular}[c]{@{}c@{}}87.8\\ (61.7\%↑)\end{tabular}            & \begin{tabular}[c]{@{}c@{}}100.8\\ (51.1\%↑)\end{tabular}           & \begin{tabular}[c]{@{}c@{}}140.3\\ (60.5\%↑)\end{tabular}           & \begin{tabular}[c]{@{}c@{}}169.1\\ (44.9\%↑)\end{tabular}           \\
$\epsilon$=0.02 & 34.0 & 65.8 & 115.0 & 128.7 & 177.0 & 204.1 \\
$\epsilon$=0.03 & 41.7 & 81.9 & 143.9 & 158.2 & 200.9 & 224.8 \\
$\epsilon$=0.04 & 48.8 & 96.0 & 169.5 & 185.5 & 216.9 & 238.2 \\
$\epsilon$=0.05              & \textbf{\begin{tabular}[c]{@{}c@{}}52.2\\ (278.3\%↑)\end{tabular}} & \textbf{\begin{tabular}[c]{@{}c@{}}101.3\\ (264.4\%↑)\end{tabular}} & \textbf{\begin{tabular}[c]{@{}c@{}}175.9\\ (223.9\%↑)\end{tabular}} & \textbf{\begin{tabular}[c]{@{}c@{}}192.0\\ (187.9\%↑)\end{tabular}} & \textbf{\begin{tabular}[c]{@{}c@{}}232.8\\ (166.4\%↑)\end{tabular}} & \textbf{\begin{tabular}[c]{@{}c@{}}250.8\\ (114.9\%↑)\end{tabular}} \\ \hline
TrajCNN-50 & 10.0 & 22.5 & 46.3 & 58.6 & 80.0 & 110.1 \\
$\epsilon$=0.01              & \begin{tabular}[c]{@{}c@{}}36.1\\ (261.0\%↑)\end{tabular}          & \begin{tabular}[c]{@{}c@{}}82.5\\ (266.7\%↑)\end{tabular}           & \begin{tabular}[c]{@{}c@{}}168.9\\ (264.8\%↑)\end{tabular}          & \begin{tabular}[c]{@{}c@{}}193.3\\ (229.9\%↑)\end{tabular}          & \begin{tabular}[c]{@{}c@{}}154.0\\ (92.5\%↑)\end{tabular}           & \begin{tabular}[c]{@{}c@{}}185.4\\ (68.4\%↑)\end{tabular}           \\
$\epsilon$=0.02 & 55.8 & 122.8 & 241.6 & 271.1 & 225.3 & 247.4 \\
$\epsilon$=0.03 & 69.7 & 149.2 & 286.2 & 317.5 & 275.1 & 292.2 \\
$\epsilon$=0.04 & 80.3 & 167.8 & 315.9 & 347.6 & 319.3 & 333.8 \\
$\epsilon$=0.05              & \textbf{\begin{tabular}[c]{@{}c@{}}92.0\\ (820.0\%↑)\end{tabular}} & \textbf{\begin{tabular}[c]{@{}c@{}}188.9\\ (739.6\%↑)\end{tabular}} & \textbf{\begin{tabular}[c]{@{}c@{}}350.1\\ (756.2\%↑)\end{tabular}} & \textbf{\begin{tabular}[c]{@{}c@{}}382.9\\ (653.4\%↑)\end{tabular}} & \textbf{\begin{tabular}[c]{@{}c@{}}356.5\\ (345.6\%↑)\end{tabular}} & \textbf{\begin{tabular}[c]{@{}c@{}}371.3\\ (237.2\%↑)\end{tabular}} \\ \hline
HRI & 10.4 & 22.6 & 47.1 & 58.3 & 77.3 & 112.1 \\
$\epsilon$=0.01              & \begin{tabular}[c]{@{}c@{}}33.6\\ (223.1\%↑)\end{tabular}          & \begin{tabular}[c]{@{}c@{}}75.9\\ (235.8\%↑)\end{tabular}           & \begin{tabular}[c]{@{}c@{}}153.9\\ (226.8\%↑)\end{tabular}          & \begin{tabular}[c]{@{}c@{}}178.2\\ (205.7\%↑)\end{tabular}          & \begin{tabular}[c]{@{}c@{}}201.0\\ (160.0\%↑)\end{tabular}          & \begin{tabular}[c]{@{}c@{}}209.4\\ (86.8\%↑)\end{tabular}           \\
$\epsilon$=0.02 & 46.6 & 100.5 & 194.4 & 221.1 & 243.3 & 244.8 \\
$\epsilon$=0.03 & 57.5 & 119.7 & 223.6 & 250.9 & 270.9 & 265.6 \\
$\epsilon$=0.04 & 65.7 & 133.0 & 242.3 & 270.4 & 290.0 & 281.5 \\
$\epsilon$=0.05              & \textbf{\begin{tabular}[c]{@{}c@{}}72.0\\ (592.3\%↑)\end{tabular}} & \textbf{\begin{tabular}[c]{@{}c@{}}142.4\\ (530.1\%↑)\end{tabular}} & \textbf{\begin{tabular}[c]{@{}c@{}}254.3\\ (439.9\%↑)\end{tabular}} & \textbf{\begin{tabular}[c]{@{}c@{}}281.9\\ (383.5\%↑)\end{tabular}} & \textbf{\begin{tabular}[c]{@{}c@{}}300.3\\ (288.5\%↑)\end{tabular}} & \textbf{\begin{tabular}[c]{@{}c@{}}288.4\\ (157.3\%↑)\end{tabular}} \\ \hline
MH TrajCNN & 12.1 & 26.4 & 52.7 & 65.5 & 84.7& 113.3  \\
$\epsilon$=0.01              & \begin{tabular}[c]{@{}c@{}}27.3\\ (125.6\%↑)\end{tabular}          & \begin{tabular}[c]{@{}c@{}}53.5\\ (102.7\%↑)\end{tabular}           & \begin{tabular}[c]{@{}c@{}}96.4\\ (82.9\%↑)\end{tabular}            & \begin{tabular}[c]{@{}c@{}}110.3\\ (68.4\%↑)\end{tabular}           & \begin{tabular}[c]{@{}c@{}}121.2\\ (43.1\%↑)\end{tabular}           & \begin{tabular}[c]{@{}c@{}}148.7\\ (31.2\%↑)\end{tabular}           \\
$\epsilon$=0.02 & 46.1 & 88.1 & 153.6 & 170.5 & 172.0 & 195.4 \\
$\epsilon$=0.03 & 61.6 & 114.6 & 193.1 & 211.6 & 217.6 & 237.3 \\
$\epsilon$=0.04 & 76.4 & 140.9 & 233.9 & 253.7 & 250.1 & 264.5\\
$\epsilon$=0.05              & \textbf{\begin{tabular}[c]{@{}c@{}}89.2\\ (637.2\%↑)\end{tabular}} & \textbf{\begin{tabular}[c]{@{}c@{}}161.0\\ (509.8\%↑)\end{tabular}} & \textbf{\begin{tabular}[c]{@{}c@{}}260.4\\ (394.1\%↑)\end{tabular}} & \textbf{\begin{tabular}[c]{@{}c@{}}279.4\\ (326.6\%↑)\end{tabular}} & \textbf{\begin{tabular}[c]{@{}c@{}}277.4\\ (227.5\%↑)\end{tabular}} & \textbf{\begin{tabular}[c]{@{}c@{}}288.7\\ (154.8\%↑)\end{tabular}} \\ \hline
\end{tabular}}
\end{table}
Comparing the errors of LTD and LTD-50 or TrajCNN and TrajCNN-50, we find that the length of input frames is also a significant factor for human motion prediction models. Except for TrajCNN with a large $\epsilon$, the errors and the growth rates when the input length is 50 are obviously lower than those when the input length is 10, and a large margin decreases the errors of repetitive activities like Walking. The reason input length influences the adversarial robustness may be that longer human motion sequences help the models learn better about the kinematics of the human body, especially the repetitive information. Nevertheless, increasing input length is a rapier for adversarial robustness because when the parameter quantity of the input data is enlarged, the chance for the adversary to attack is also increased. The results of TrajCNN when $\epsilon$ is 0.05 show that increasing input length is not always valid, and the disturbance of the attack can burst when the attack is intense enough.\par
However, on the basis of the gap between the results of HRI and LTD-50, the attention mechanism is a burden under attack for human motion prediction. The attention mechanism can help HRI learn the different sub-sequence contributions and the repetitive pattern of the input human motion sequence. The proposed attack can mislead the attention and enlarge the disturbance caused by the perturbed parts of the human motion sequence. To our surprise, in H3.6M, the augmentation used in HRI for the history motion sequence with the last prediction does not cause a further influence on the attack effectiveness, which may be because the original history human motion sequence takes the upper hand in the augmented history human motion sequence.\par
Most of the absolute average errors increased as the time interval enlarged, but the errors of TrajCNN and HRI don’t follow this trend in long-term prediction. The reason for this non-monotonic phenomenon is different for the two models. For TrajCNN, we attribute this phenomenon to the short- to long-term switching model. These two models may use different ways to deal with the perturbed input. For HRI, this phenomenon may be attributed to the attention mechanism, which gives HRI efficiency in capturing repetitiveness. As we only maximize the average prediction error, it is hard to ensure the specific errors in each time interval, especially for repetitive activities. This phenomenon only happens when $\epsilon$ is larger than 0.02, and the motion is more and more out of the control of the clean sequence when $\epsilon$ is larger and larger. According to our observation, the perturbed prediction motions hardly repeat in pattern, which makes it possible for the joints in repetitive motions to move close to the ground truth location when the time interval is large enough. As a result, the non-monotonic phenomenon occurs in HRI. This non-monotonic phenomenon is more frequent in specific activities and will be discussed in \ref{app:2} due to space limits.\par

\begin{table}[htbp]
\centering
\caption{\label{tab:3}Average prediction errors of target models on CMU before and after perturbation in boundary variations for the models whose input length is 10. The \textbf{bold} results are the worst-case of each time interval, and the percentage is the least and the most growth rate from the clean error.}
\scalebox{0.5}{
\begin{tabular}{c|cccccc}
\hline
{\ul Intervals(ms)} & 80 & 160 & 320 & 400 & 560 & 1000 \\ \hline
LTD & 10.1 & 18.0 & 36.3 & 46.8 & 62.7 & 95.0 \\
$\epsilon$=0.01              & \begin{tabular}[c]{@{}c@{}}20.9\\ (106.9\%↑)\end{tabular}          & \begin{tabular}[c]{@{}c@{}}41.1\\ (128.3\%↑)\end{tabular}          & \begin{tabular}[c]{@{}c@{}}86.1\\ (137.2\%↑)\end{tabular}           & \begin{tabular}[c]{@{}c@{}}107.5\\ (129.7\%↑)\end{tabular}          & \begin{tabular}[c]{@{}c@{}}140.9\\ (124.7\%↑)\end{tabular}          & \begin{tabular}[c]{@{}c@{}}181.0\\ (90.5\%↑)\end{tabular}           \\
$\epsilon$=0.02 & 29.6 & 56.5 & 114.7 & 141.5 & 173.0 & 211.1 \\
$\epsilon$=0.03 & 35.9 & 66.6 & 133.1 & 163.4 & 193.0 & 230.4 \\
$\epsilon$=0.04 & 39.7 & 71.9 & 140.8 & 171.7 & 203.7 & 242.0  \\
$\epsilon$=0.05              & \textbf{\begin{tabular}[c]{@{}c@{}}43.5\\ (330.7\%↑)\end{tabular}} & \textbf{\begin{tabular}[c]{@{}c@{}}76.3\\ (323.9\%↑)\end{tabular}} & \textbf{\begin{tabular}[c]{@{}c@{}}146.1\\ (302.5\%↑)\end{tabular}} & \textbf{\begin{tabular}[c]{@{}c@{}}177.5\\ (279.3\%↑)\end{tabular}} & \textbf{\begin{tabular}[c]{@{}c@{}}209.0\\ (233.3\%↑)\end{tabular}} & \textbf{\begin{tabular}[c]{@{}c@{}}245.8\\ (158.7\%↑)\end{tabular}} \\ \hline
TrajCNN & 8.4 & 15.2 & 32.4 & 42.5 & 60.7 & 94.0 \\
$\epsilon$=0.01              & \begin{tabular}[c]{@{}c@{}}22.0\\ (161.9\%↑)\end{tabular}          & \begin{tabular}[c]{@{}c@{}}43.4\\ (185.5\%↑)\end{tabular}          & \begin{tabular}[c]{@{}c@{}}94.3\\ (191.0\%↑)\end{tabular}           & \begin{tabular}[c]{@{}c@{}}118.7\\ (179.3\%↑)\end{tabular}          & \begin{tabular}[c]{@{}c@{}}128.7\\ (112.0\%↑)\end{tabular}          & \begin{tabular}[c]{@{}c@{}}185.3\\ (97.1\%↑)\end{tabular}           \\
$\epsilon$=0.02 & 33.5 & 65.0 & 136.4 & 169.3 & 179.2 & 243.8 \\
$\epsilon$=0.03 & 42.9 & 80.8 & 164.9 & 202.6 & 207.6 & 274.3 \\
$\epsilon$=0.04 & 48.6 & 90.4 & 181.5 & 221.7 & 224.4 & 291.1 \\
$\epsilon$=0.05              & \textbf{\begin{tabular}[c]{@{}c@{}}54.3\\ (546.4\%↑)\end{tabular}} & \textbf{\begin{tabular}[c]{@{}c@{}}98.1\\ (545.4\%↑)\end{tabular}} & \textbf{\begin{tabular}[c]{@{}c@{}}191.8\\ (492.0\%↑)\end{tabular}} & \textbf{\begin{tabular}[c]{@{}c@{}}232.2\\ (446.4\%↑)\end{tabular}} & \textbf{\begin{tabular}[c]{@{}c@{}}238.7\\ (293.2\%↑)\end{tabular}} & \textbf{\begin{tabular}[c]{@{}c@{}}307.1\\ (226.7\%↑)\end{tabular}} \\ \hline
\end{tabular}}
\end{table}
\textbf{Results on CMU-Mocap.} Table \ref{tab:3} and \ref{tab:4} shows the error comparisons of the target models on CMU-Mocap, and the detailed results are in \ref{app:1}. On this dataset, the target models are also severely interrupted. However, the most adversarial robust algorithm is LTD-50, LTD, then MH TrajCNN, TrajCNN, TrajCNN-50, and finally, HRI is the most vulnerable model.\par
CMU-Mocap contains more sports activities, which causes the difference in the attack effectiveness and the adversarial robustness ranking. By comparing the errors between H3.6M and CMU, we find that most of the growth rates in CMU are much less than those in H3.6M. This result shows that the activities with more movements may cause the attack effectiveness to be weaker, or the model trained with a specialized kind of activity may be more adversarial robust.\par
However, surpassed by TrajCNN, HRI drops to the last place in the ranking of adversarial robustness, which may be attributed to the attention mechanism and the augmentation with the last prediction. The misleading effect comes into force to HRI as it does in the H3.6M and is noticeable when predicting the “running” activity in CMU. For the activities in CMU, the augmentation in HRI stabilizes the attack effectiveness as the prediction time interval grows. Note that the kernel size of HRI is 10, which means 80ms, 160ms, 320ms, and 400ms are in the first window, 560ms is in the second window, and 1000ms is in the third window. The attack against other target models will weaken as the time interval is enlarged, while the attack effectiveness against HRI decreases much slower. By comparing the long-term results between LTD-50 and HRI, it is clear that the errors of HRI in 560ms and 1000ms increase much faster than those of LTD-50. Therefore, the last prediction used to augment the history motion sequence has more influence when the model focuses on sports activities.\par
\begin{table}[htbp]
\centering
\caption{\label{tab:4}Average prediction errors of target models on CMU before and after perturbation in boundary variations for the models whose input length is 50. The \textbf{bold} results are the worst-case of each time interval, and the percentage is the least and the most growth rate from the clean error.}
\scalebox{0.5}{
\begin{tabular}{c|cccccc}
\hline
{\ul Intervals(ms)} & 80 & 160 & 320 & 400 & 560 & 1000 \\ \hline
LTD-50 & 9.3 & 17.0 & 33.9 & 43.1 & 63.0 & 94.3 \\
$\epsilon$=0.01              & \begin{tabular}[c]{@{}c@{}}21.8\\ (134.4\%↑)\end{tabular}          & \begin{tabular}[c]{@{}c@{}}40.4\\ (137.6\%↑)\end{tabular}           & \begin{tabular}[c]{@{}c@{}}78.5\\ (131.6\%↑)\end{tabular}           & \begin{tabular}[c]{@{}c@{}}92.2\\ (113.9\%↑)\end{tabular}           & \begin{tabular}[c]{@{}c@{}}113.7\\ (80.5\%↑)\end{tabular}           & \begin{tabular}[c]{@{}c@{}}152.4\\ (61.6\%↑)\end{tabular}           \\
$\epsilon$=0.02 & 27.6 & 51.9 & 102.9 & 119.9 & 140.5 & 177.1 \\
$\epsilon$=0.03 & 32.2 & 59.2 & 116.1 & 134.5 & 159.9 & 195.7 \\
$\epsilon$=0.04 & 37.2 & 67.8 & 132.9 & 153.5 & 180.1 & 215.0 \\
$\epsilon$=0.05 & \textbf{\begin{tabular}[c]{@{}c@{}}41.2\\ (343.0\%↑)\end{tabular}} & \textbf{\begin{tabular}[c]{@{}c@{}}73.6\\ (332.9\%↑)\end{tabular}}  & \textbf{\begin{tabular}[c]{@{}c@{}}140.7\\ (315.0\%↑)\end{tabular}} & \textbf{\begin{tabular}[c]{@{}c@{}}161.3\\ (274.2\%↑)\end{tabular}} & \textbf{\begin{tabular}[c]{@{}c@{}}194.4\\ (208.6\%↑)\end{tabular}} & \textbf{\begin{tabular}[c]{@{}c@{}}229.5\\ (143.4\%↑)\end{tabular}} \\ \hline
TrajCNN-50 & 8.8 & 16.4 & 33.5 & 43.3 & 61.7 & 97.3 \\
$\epsilon$=0.01              & \begin{tabular}[c]{@{}c@{}}20.4\\ (131.8\%↑)\end{tabular}          & \begin{tabular}[c]{@{}c@{}}39.9\\ (143.3\%↑)\end{tabular}           & \begin{tabular}[c]{@{}c@{}}83.0\\ (147.8\%↑)\end{tabular}           & \begin{tabular}[c]{@{}c@{}}103.6\\ (139.3\%↑)\end{tabular}          & \begin{tabular}[c]{@{}c@{}}102.9\\ (66.8\%↑)\end{tabular}           & \begin{tabular}[c]{@{}c@{}}144.7\\ (48.7\%↑)\end{tabular}           \\
$\epsilon$=0.02 & 32.4 & 62.4 & 126.9 & 155.9 & 154.7 & 203.2 \\
$\epsilon$=0.03 & 42.3 & 79.3 & 156.2 & 189.5 & 198.7 & 248.2 \\
$\epsilon$=0.04 & 52.1 & 95.8 & 183.8 & 219.9 & 232.9 & 277.8  \\
$\epsilon$=0.05              & \textbf{\begin{tabular}[c]{@{}c@{}}59.8\\ (579.5\%↑)\end{tabular}} & \textbf{\begin{tabular}[c]{@{}c@{}}108.5\\ (561.6\%↑)\end{tabular}} & \textbf{\begin{tabular}[c]{@{}c@{}}207.7\\ (620.0\%↑)\end{tabular}} & \textbf{\begin{tabular}[c]{@{}c@{}}249.0\\ (475.1\%↑)\end{tabular}} & \textbf{\begin{tabular}[c]{@{}c@{}}249.1\\ (303.7\%↑)\end{tabular}} & \textbf{\begin{tabular}[c]{@{}c@{}}293.8\\ (302.0\%↑)\end{tabular}} \\ \hline
HRI & 9.5 & 17.0 & 33.9 & 43.2 & 59.8 & 101.6 \\
$\epsilon$=0.01              & \begin{tabular}[c]{@{}c@{}}22.6\\ (137.9\%↑)\end{tabular}          & \begin{tabular}[c]{@{}c@{}}48.1\\ (182.9\%↑)\end{tabular}           & \begin{tabular}[c]{@{}c@{}}114.5\\ (237.8\%↑)\end{tabular}          & \begin{tabular}[c]{@{}c@{}}148.8\\ (244.4\%↑)\end{tabular}          & \begin{tabular}[c]{@{}c@{}}210.8\\ (252.5\%↑)\end{tabular}          & \begin{tabular}[c]{@{}c@{}}285.8\\ (181.3\%↑)\end{tabular}          \\
$\epsilon$=0.02 & 34.1 & 70.6 & 159.0 & 202.9 & 273.8 & 335.5 \\
$\epsilon$=0.03 & 42.1 & 83.6 & 182.1 & 230.0 & 305.0 & 367.4 \\
$\epsilon$=0.04 & 48.6 & 93.3 & 196.5 & 245.8 & 319.0 & 377.4 \\
$\epsilon$=0.05              & \textbf{\begin{tabular}[c]{@{}c@{}}55.3\\ (482.1\%↑)\end{tabular}} & \textbf{\begin{tabular}[c]{@{}c@{}}104.1\\ (512.4\%↑)\end{tabular}} & \textbf{\begin{tabular}[c]{@{}c@{}}215.9\\ (536.9\%↑)\end{tabular}} & \textbf{\begin{tabular}[c]{@{}c@{}}268.6\\ (521.8\%↑)\end{tabular}} & \textbf{\begin{tabular}[c]{@{}c@{}}344.4\\ (475.9\%↑)\end{tabular}} & \textbf{\begin{tabular}[c]{@{}c@{}}397.4\\ (291.1\%↑)\end{tabular}} \\ \hline
MH TrajCNN & 9.7 & 17.7 & 36.9 & 47.5 & 68.4 & 103.3 \\
$\epsilon$=0.01              & \begin{tabular}[c]{@{}c@{}}16.8\\ (73.2\%↑)\end{tabular}           & \begin{tabular}[c]{@{}c@{}}29.0\\ (63.8\%↑)\end{tabular}            & \begin{tabular}[c]{@{}c@{}}56.2\\ (52.3\%↑)\end{tabular}            & \begin{tabular}[c]{@{}c@{}}69.3\\ (45.9\%↑)\end{tabular}            & \begin{tabular}[c]{@{}c@{}}93.8\\ (37.1\%↑)\end{tabular}            & \begin{tabular}[c]{@{}c@{}}130.3\\ (26.1\%↑)\end{tabular}           \\
$\epsilon$=0.02 & 26.1 & 43.5 & 80.7 & 97.7 & 127.0 & 169.1 \\
$\epsilon$=0.03 & 36.0 & 59.6 & 108.9 & 131.1 & 164.4 & 214.7 \\
$\epsilon$=0.04 & 45.8 & 74.4 & 133.5 & 159.4 & 194.9 & 244.9 \\
$\epsilon$=0.05              & \textbf{\begin{tabular}[c]{@{}c@{}}54.4\\ (460.8\%↑)\end{tabular}} & \textbf{\begin{tabular}[c]{@{}c@{}}86.7\\ (389.8\%↑)\end{tabular}}  & \textbf{\begin{tabular}[c]{@{}c@{}}151.6\\ (310.8\%↑)\end{tabular}} & \textbf{\begin{tabular}[c]{@{}c@{}}179.4\\ (277.7\%↑)\end{tabular}} & \textbf{\begin{tabular}[c]{@{}c@{}}220.7\\ (222.7\%↑)\end{tabular}} & \textbf{\begin{tabular}[c]{@{}c@{}}269.1\\ (160.5\%↑)\end{tabular}} \\ \hline
\end{tabular}}
\end{table}

\begin{table}[htbp]
\centering
\caption{\label{tab:5}Average prediction errors of target models on 3DPW before and after perturbation in boundary variations for the models whose input length is 10. The \textbf{bold} results are the worst-case of each time interval, and the percentage is the least and the most growth rate from the clean error.}
\scalebox{0.5}{
\begin{tabular}{c|ccccc}
\hline
{\ul Intervals(ms)} & 200 & 400 & 600 & 800 & 1000 \\ \hline
LTD & 34.4 & 65.7 & 93.6 & 105.6 & 113.5 \\
$\epsilon$=0.01              & \begin{tabular}[c]{@{}c@{}}56.3\\ (63.7\%↑)\end{tabular}            & \begin{tabular}[c]{@{}c@{}}103.9\\ (58.1\%↑)\end{tabular}           & \begin{tabular}[c]{@{}c@{}}181.6\\ (94.0\%↑)\end{tabular}           & \begin{tabular}[c]{@{}c@{}}205.7\\ (94.8\%↑)\end{tabular}           & \begin{tabular}[c]{@{}c@{}}211.0\\ (85.9\%↑)\end{tabular}           \\
$\epsilon$=0.02 & 68.8 & 125.0 & 209.8 & 236.0 & 240.1 \\
$\epsilon$=0.03 & 77.6 & 138.8 & 221.9 & 248.7 & 252.1 \\
$\epsilon$=0.04 & 83.6 & 147.4 & 227.6 & 254.4 & 257.3 \\
$\epsilon$=0.05              & \textbf{\begin{tabular}[c]{@{}c@{}}88.3\\ (156.7\%↑)\end{tabular}}  & \textbf{\begin{tabular}[c]{@{}c@{}}153.4\\ (133.5\%↑)\end{tabular}} & \textbf{\begin{tabular}[c]{@{}c@{}}231.0\\ (146.8\%↑)\end{tabular}} & \textbf{\begin{tabular}[c]{@{}c@{}}257.8\\ (144.1\%↑)\end{tabular}} & \textbf{\begin{tabular}[c]{@{}c@{}}260.3\\ (129.3\%↑)\end{tabular}} \\ \hline
TrajCNN & 29.9 & 60.0 & 85.3 & 99.3 & 107.5 \\
$\epsilon$=0.01              & \begin{tabular}[c]{@{}c@{}}66.3\\ (121.7\%↑)\end{tabular}           & \begin{tabular}[c]{@{}c@{}}129.2\\ (115.3\%↑)\end{tabular}          & \begin{tabular}[c]{@{}c@{}}132.7\\ (55.6\%↑)\end{tabular}           & \begin{tabular}[c]{@{}c@{}}147.3\\ (48.3\%↑)\end{tabular}           & \begin{tabular}[c]{@{}c@{}}150.3\\ (39.8\%↑)\end{tabular}           \\
$\epsilon$=0.02 & 96.7 & 179.2 & 168.8 & 182.1 & 181.9 \\
$\epsilon$=0.03 & 116.7 & 209.0 & 191.5 & 202.9 & 200.9 \\
$\epsilon$=0.04 & 128.7 & 225.9 & 204.8 & 214.8 & 212.1 \\
$\epsilon$=0.05              & \textbf{\begin{tabular}[c]{@{}c@{}}137.1\\ (358.5\%↑)\end{tabular}} & \textbf{\begin{tabular}[c]{@{}c@{}}237.0\\ (295.0\%↑)\end{tabular}} & \textbf{\begin{tabular}[c]{@{}c@{}}214.0\\ (150.9\%↑)\end{tabular}} & \textbf{\begin{tabular}[c]{@{}c@{}}223.0\\ (124.6\%↑)\end{tabular}} & \textbf{\begin{tabular}[c]{@{}c@{}}219.8\\ (104.5\%↑)\end{tabular}} \\ \hline
\end{tabular}}
\end{table}
\textbf{Results on 3DPW.} Table \ref{tab:5} and \ref{tab:6} shows the comparisons on 3DPW. The proposed attack still works well against all the target models. Based on the error growth rates, the models can be generally sorted from robust to vulnerable as LTD-50, LTD, TrajCNN-50, TrajCNN, and HRI. The activities from 3DPW are wilder than the other two datasets. Thus, the average results show more features of the complex activities.\par
One of the reasons why the growth rate of the errors under attack is relatively low is that the activities in 3DPW are wilder than those in CMU. As the results in CMU show that activities with more movements can weaken the attack, the attack effectiveness judged from the growth rates in 3DPW is not as drastic as in other datasets. Although another reason is that the errors under the clean circumstance are much higher than those in H3.6M and CMU, most of the absolute errors of the same time interval in 3DPW are lower, proving the influence of the complexity on the attack.\par
\begin{table}[htbp]
\centering
\caption{\label{tab:6}Average prediction errors of target models on 3DPW before and after perturbation in boundary variations for the models whose input length is 50. The \textbf{bold} results are the worst-case of each time interval, and the percentage is the least and the most growth rate from the clean error.}
\scalebox{0.5}{
\begin{tabular}{c|ccccc}
\hline
{\ul Intervals(ms)} & 200 & 400 & 600 & 800 & 1000 \\ \hline
LTD-50 & 34.8 & 65.1 & 93.1 & 104.9 & 113.5 \\
$\epsilon$=0.01              & \begin{tabular}[c]{@{}c@{}}54.3\\ (56.0\%↑)\end{tabular}            & \begin{tabular}[c]{@{}c@{}}91.8\\ (41.0\%↑)\end{tabular}            & \begin{tabular}[c]{@{}c@{}}131.2\\ (40.9\%↑)\end{tabular}           & \begin{tabular}[c]{@{}c@{}}144.4\\ (37.7\%↑)\end{tabular}           & \begin{tabular}[c]{@{}c@{}}153.0\\ (34.8\%↑)\end{tabular}           \\
$\epsilon$=0.02 & 63.2 & 108.5 & 158.1 & 172.0 & 180.9 \\
$\epsilon$=0.03 & 70.1 & 120.7 & 174.7 & 188.9 & 197.8 \\
$\epsilon$=0.04 & 75.8 & 130.0 & 186.0 & 200.2 & 209.2 \\
$\epsilon$=0.05              & \textbf{\begin{tabular}[c]{@{}c@{}}80.6\\ (131.6\%↑)\end{tabular}}  & \textbf{\begin{tabular}[c]{@{}c@{}}137.5\\ (111.2\%↑)\end{tabular}} & \textbf{\begin{tabular}[c]{@{}c@{}}194.2\\ (108.6\%↑)\end{tabular}} & \textbf{\begin{tabular}[c]{@{}c@{}}208.5\\ (98.8\%↑)\end{tabular}}  & \textbf{\begin{tabular}[c]{@{}c@{}}217.5\\ (91.6\%↑)\end{tabular}}  \\ \hline
TrajCNN-50 & 31.1 & 61.4 & 86.2 & 101.1& 111.2 \\
$\epsilon$=0.01              & \begin{tabular}[c]{@{}c@{}}58.6\\ (88.4\%↑)\end{tabular}            & \begin{tabular}[c]{@{}c@{}}109.0\\ (77.5\%↑)\end{tabular}           & \begin{tabular}[c]{@{}c@{}}114.7\\ (33.1\%↑)\end{tabular}           & \begin{tabular}[c]{@{}c@{}}129.8\\ (28.4\%↑)\end{tabular}           & \begin{tabular}[c]{@{}c@{}}136.1\\ (22.4\%↑)\end{tabular}           \\
$\epsilon$=0.02 & 90.8 & 161.3 & 148.1 & 162.3 & 165.2 \\
$\epsilon$=0.03 & 119.2 & 203.6 & 177.8 & 190.2 & 190.1 \\
$\epsilon$=0.04 & 142.6 & 235.8 & 202.1 & 212.2 & 209.7 \\
$\epsilon$=0.05              & \textbf{\begin{tabular}[c]{@{}c@{}}161.7\\ (419.9\%↑)\end{tabular}} & \textbf{\begin{tabular}[c]{@{}c@{}}260.3\\ (323.9\%↑)\end{tabular}} & \textbf{\begin{tabular}[c]{@{}c@{}}221.4\\ (156.8\%↑)\end{tabular}} & \textbf{\begin{tabular}[c]{@{}c@{}}229.3\\ (126.8\%↑)\end{tabular}} & \textbf{\begin{tabular}[c]{@{}c@{}}224.8\\ (102.2\%↑)\end{tabular}} \\ \hline
HRI & 43.1 & 85.9 & 114.6 & 133.6 & 146.8 \\
$\epsilon$=0.01              & \begin{tabular}[c]{@{}c@{}}98.6\\ (128.8\%↑)\end{tabular}           & \begin{tabular}[c]{@{}c@{}}185.9\\ (116.4\%↑)\end{tabular}          & \begin{tabular}[c]{@{}c@{}}228.0\\ (99.0\%↑)\end{tabular}           & \begin{tabular}[c]{@{}c@{}}244.4\\ (82.9\%↑)\end{tabular}           & \begin{tabular}[c]{@{}c@{}}247.5\\ (68.6\%↑)\end{tabular}           \\
$\epsilon$=0.02 & 128.1 & 232.3 & 275.9 & 288.3 & 287.1 \\
$\epsilon$=0.03 & 145.6 & 257.8 & 300.7 & 310.7 & 307.6 \\
$\epsilon$=0.04 & 158.3 & 275.1 & 317.4 & 325.5 & 321.0 \\
$\epsilon$=0.05              & \textbf{\begin{tabular}[c]{@{}c@{}}167.5\\ (288.6\%↑)\end{tabular}} & \textbf{\begin{tabular}[c]{@{}c@{}}287.2\\ (234.3\%↑)\end{tabular}} & \textbf{\begin{tabular}[c]{@{}c@{}}329.0\\ (187.1\%↑)\end{tabular}} & \textbf{\begin{tabular}[c]{@{}c@{}}336.0\\ (151.5\%↑)\end{tabular}} & \textbf{\begin{tabular}[c]{@{}c@{}}330.9\\ (125.4\%↑)\end{tabular}} \\ \hline
\end{tabular}}
\end{table}
\textbf{Semantic Information} With the help from the classification branch of MH TrajCNN, we can observe whether the proposed attack influences the semantic information of the predicted sequence based on the ASR results. As shown in Table \ref{tab:7}, most of the ASR is at a low level, which indicates that the proposed attack has little effect on the semantic information of the predicted sequence. \par
However, the long-term classifier in H3.6M and the short-term classifier in CMU are more sensitive than the others as their ASRs are relatively higher. The reason may be the different composition of the activities in these two datasets. The activities in H3.6M are more than in CMU in amount and variety. Moreover, there are more static activities in H3.6M, while most of the activities in CMU are from sports. Hence the semantic information learned from H3.6M may focus more on the detailed information, which makes the classification branch trained in the long-term model on H3.6M more sensitive. Meanwhile, distinguishing motions with high frequency is more common in CMU for the classification branch in MH TrajCNN, making the classifier focus more on the pattern information that needs to be expected for a longer time.\par
Considering the results in Table \ref{tab:7} and comparing the short-term and long-term MH TrajCNN errors, the MH TrajCNN with a more sensitive classifier suppresses the attack effectiveness less than the others when facing the proposed attack with $\epsilon=0.05$. This phenomenon indicates the sensitivity of the classification branch may negatively relate to the influence of the semantic information modeling on the adversarial robustness of the model when the model is facing an intense attack.\par

\begin{table}[htbp]
\centering
\caption{\label{tab:7}Accuracy and ASR of short-term and long-term Multi-Head TrajectoryCNN on H3.6M and CMU before and after perturbation in different boundaries.}
\scalebox{0.5}{
\begin{tabular}{ccccccc}
\hline
\multicolumn{1}{l}{{\ul }} & clean  & $\epsilon$=0.01 & $\epsilon$=0.02 & $\epsilon$=0.03 & $\epsilon$=0.04 & $\epsilon$=0.05 \\ \hline
\multicolumn{7}{c}{H3.6M}                                                            \\ \hline
\multicolumn{1}{c|}{Accuracy-short} & 63.3\% & 62.5\% & 62.5\% & 65.0\% & 60.8\% & 60.0\% \\
\multicolumn{1}{c|}{ASR-short}      & -      & 1.3\%  & 2.7\%  & 2.7\%  & 10.7\% & 10.7\% \\
\multicolumn{1}{c|}{Accuracy-long}  & 58.3\% & 56.7\% & 55.8\% & 55.0\% & 55.0\% & 50.0\% \\
\multicolumn{1}{c|}{ASR-long}       & -      & 2.9\%  & 4.3\%  & 8.6\%  & 10.0\% & 18.6\% \\ \hline
\multicolumn{7}{c}{CMU}                                                                   \\ \hline
\multicolumn{1}{c|}{Accuracy-short} & 82.8\% & 79.7\% & 78.1\% & 76.6\% & 68.8\% & 68.8\% \\
\multicolumn{1}{c|}{ASR-short}      & -      & 3.8\%  & 5.7\%  & 7.5\%  & 17.0\% & 17.0\% \\
\multicolumn{1}{c|}{Accuracy-long}  & 82.8\% & 81.3\% & 79.7\% & 78.1\% & 78.1\% & 75.0\% \\
\multicolumn{1}{c|}{ASR-long}       & -      & 1.9\%  & 3.8\%  & 5.7\%  & 5.7\%  & 9.4\% 
            \\ \hline
\end{tabular}}
\end{table}

\begin{figure}
    \centering
    \includegraphics[width=0.8\textwidth]{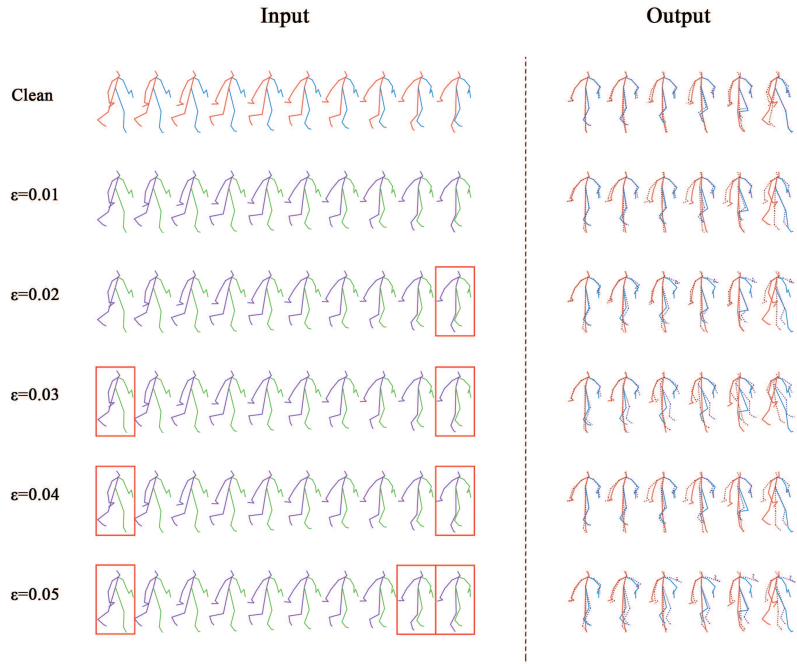}
    \caption{\label{fig:2}This figure shows LTD’s prediction result of “walking” before and after the attack. The poses with solid lines in red and blue are the clean input and ground truth; For the input sequences, the skeletons in purple and green are the perturbed sequence under the attack with the corresponding intensities written on the left. As for the output sequences, the dotted skeletons in the prediction are the predicted poses when their time intervals are 80ms, 160ms, 320ms, 400ms, 560ms, and 1000ms. Because the differences between the clean input and the perturbed input are tiny, we mark the poses with relatively noticeable changes with red boxes.}
\end{figure}
\subsection{Qualitative Results} \label{sec:5.3}
The qualitative results prove the stealthiness of the proposed attack. When every input pose is listed on the left side of Figure \ref{fig:2}, it is easy to find that the perturbations hardly change the input human motion sequences visually. On the contrary, the predicted poses shown on the right side of Figure \ref{fig:2} are significantly misled far away from the ground truth, indicating the proposed attack’s effectiveness.\par
The differences between the clean human motion sequence and the perturbed motion sequence are so tiny that it is hard to distinguish most parts in these two sequences when overlapping them. Therefore, we use red boxes to mark the poses with noticeable visual differences. According to the red boxes, the perturbation tends to modify the last frame of the input more than other frames. This tendency is not only happening to LTD but happening to all target models. The reason can be summarized as follows: 1) The last frame strongly correlates to predicting human motion, which means the last frame can define most of the information in the following motion, e.g., directions. Thus, the optimization for attack takes more weight on the last frame. 2) In LTD and TrajCNN, the last frame is regarded as the observed pose and repeats $T_f$ times, emphasizing the last frame in the optimization.\par
Although the perturbed sequence is difficult to notice when every frame is picked, the perturbation is easier to notice when the perturbed human motion sequence is made into an animation,e.g. GIF. Based on our observation, the perturbed motion looks like shaking, while its predicted motion will suddenly be bizarre, including making actions impossible for humans and falling like an offline puppet.\par
\begin{figure}
    \centering
    \includegraphics[width=0.8\textwidth]{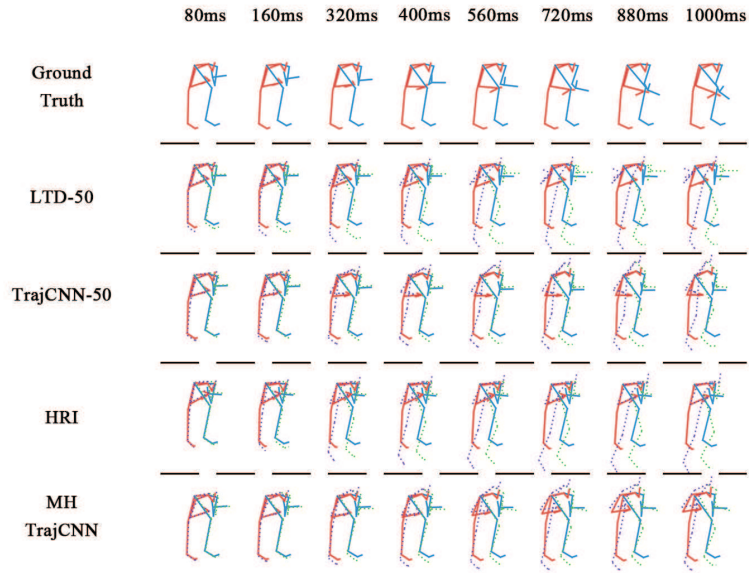}
    \caption{\label{fig:3}This figure shows the different models' prediction results of "sitting" before and after the input motion sequence is perturbed when $\epsilon$ is set as 0.01. The ground truth is shown at the top of the figure. Below the ground truth, the skeletons in red and blue are the clean predictions of the corresponding model named on the left, while the dotted skeletons in purple and green are the corresponding disturbed predictions. Because the output motion sequence is too long to fully display, the time intervals of the predicted poses from the left to the right shown in the figure are 80ms, 160ms, 320ms, 400ms, 560ms, 720ms, 880ms, and 1000ms.}
\end{figure}

Besides the stealthiness of the proposed attack, the adversarial robustness of the target models can be analyzed through the visualization of their prediction results before and after applying the attack. To fairly compare the adversarial robustness of each model, we unify the input length of the models and adopt the results of LTD-50, TrajCNN-50, HRI, and MH TrajCNN, which are shown in Figure \ref{fig:3}. Because the prediction results will be completely disordered when $\epsilon$ is large, we adopt the models' results of "sitting" when $\epsilon$ is 0.01.\par
The vulnerability of the target models is abundantly clear in the visualized results. After the attack, all of the predictions turn the activity of "sitting" into "standing" which is the opposite of the original activity. In addition, based on the observation of Figure \ref{fig:3}, the attack enlarges the prediction error by increasing the range of the original motion in the clean prediction motion sequence. The motion of two hands predicted by the same model before and after the attack is apparently in the same direction, and this phenomenon happens in all target models, which can also be the reason why all perturbed motion sequences are "standing" as the original predictions for two legs have a momentum to "stand". \par
Moreover, the visualized results show a clue of the HMP's adversarial robustness. For the relatively more adversarial robust models, LTD-50 and MH TrajCNN, although they are all misled to a wrong human motion, the motion of the human's torso in the prediction is influenced by a small degree. What's more, MH TrajCNN may learn the pattern of "sitting" and prevent the prediction results from acting like "raising the hands". On the contrary, for the relatively less adversarial robust models, TrajCNN-50 and HRI, the body trunk in their prediction may be led to a different direction, and the predicted motion sequence may act like "leaning" or "turning back". This phenomenon indicates that predicting the correct direction of human motion is more important than predicting the precise position for the adversarial robustness of human motion prediction.\par

To further investigate the general trend of optimized perturbations on the various joints, we analyze the intensities of the proposed attack on different joints. The attack's intensity here is defined as the magnitude of the perturbation on the target joint. To fairly evaluate the perturbation's tendency,  we calculate the expected value $\mu$ and the variance $\sigma$ of the modulus of all perturbations in each joint over all sequences in the test set of H3.6M. In this analysis, we set $\epsilon$ as 0.01 while LTD and LTD-50 are the target models to analyze whether the input length influences the intensity of the proposed attack. The results are shown in Table \ref{tab:8}.\par
\begin{table}[htbp]
\centering
\renewcommand{\arraystretch}{1.5}
\caption{\label{tab:8}The expected value and the variance of the perturbations' modulus against LTD and LTD-50 in each joint of H3.6M pose data. The \textbf{bold} results are the largest in their corresponding limb, while the {\ul underlined} results are the least.}
\scalebox{0.5}{
\begin{tabular}{cc|ccccccccc}
\hline
\multicolumn{2}{c|}{\multirow{3}{*}{joints}} & \multicolumn{9}{c}{right} \\ \cline{3-11} 
\multicolumn{2}{c|}{} & \multicolumn{4}{c|}{leg} & \multicolumn{5}{c}{arm} \\ \cline{3-11} 
\multicolumn{2}{c|}{} & thigh & tibia & sole & \multicolumn{1}{c|}{toe} & humerus & radius & wrist & palm & thumb \\ \hline
\multicolumn{1}{c|}{\multirow{2}{*}{LTD}} & $\mu$ & 3.248 & {\ul 3.206} & 3.251 & \multicolumn{1}{c|}{\textbf{3.264}} & 3.178 & {\ul 3.120} & 3.183 & 3.137 & \textbf{3.296} \\
\multicolumn{1}{c|}{} & $\sigma$ & 0.944 & 0.939 & {\ul 0.928} & \multicolumn{1}{c|}{\textbf{1.080}} & {\ul 0.826} & 0.878 & 1.012 & 0.881 & \textbf{1.375} \\ \hline
\multicolumn{1}{c|}{\multirow{2}{*}{LTD-50}} & $\mu$ & \textbf{2.684} & 2.668 & {\ul 2.665} & \multicolumn{1}{c|}{2.672} & 2.657 & {\ul 2.566} & 2.670 & 2.649 & \textbf{2.715} \\
\multicolumn{1}{c|}{} & $\sigma$ & 0.487 & {\ul 0.462} & 0.492 & \multicolumn{1}{c|}{\textbf{0.509}} & 0.481 & {\ul 0.454} & 0.491 & 0.486 & \textbf{0.574} \\ \hline
\multicolumn{2}{c|}{\multirow{3}{*}{joints}} & \multicolumn{9}{c}{left} \\ \cline{3-11} 
\multicolumn{2}{c|}{} & \multicolumn{4}{c|}{leg} & \multicolumn{5}{c}{arm} \\ \cline{3-11} 
\multicolumn{2}{c|}{} & thigh & tibia & sole & \multicolumn{1}{c|}{toe} & humerus & radius & wrist & palm & thumb \\ \hline
\multicolumn{1}{c|}{\multirow{2}{*}{LTD}} & $\mu$ & 3.188 & {\ul 3.140} & 3.201 & \multicolumn{1}{c|}{\textbf{3.356}} & {\ul 3.201} & 3.226 & \textbf{3.259} & 3.212 & 3.220 \\
\multicolumn{1}{c|}{} & $\sigma$ & 0.939 & {\ul 0.844} & 0.931 & \multicolumn{1}{c|}{\textbf{1.053}} & {\ul 0.823} & 0.830 & 1.026 & 0.950 & \textbf{1.104} \\ \hline
\multicolumn{1}{c|}{\multirow{2}{*}{LTD-50}} & $\mu$ & 2.673 & {\ul 2.657} & 2.668 & \multicolumn{1}{c|}{\textbf{2.684}} & 2.662 & 2.662 & {\ul 2.657} & 2.666 & \textbf{2.679} \\
\multicolumn{1}{c|}{} & $\sigma$ & \textbf{0.493} & 0.453 & {\ul 0.446} & \multicolumn{1}{c|}{0.449} & {\ul 0.416} & 0.452 & 0.441 & \textbf{0.485} & 0.475 \\ \hline
\multicolumn{2}{c|}{\multirow{3}{*}{joints}} & \multicolumn{4}{c}{\multirow{2}{*}{torso}} & \multicolumn{5}{c}{\multirow{2}{*}{}} \\
\multicolumn{2}{c|}{} & \multicolumn{4}{c}{} & \multicolumn{5}{c}{} \\ \cline{3-6}
\multicolumn{2}{c|}{} & backbone & neck & throat & head & \multicolumn{1}{l}{} & \multicolumn{1}{l}{} & \multicolumn{1}{l}{} & \multicolumn{1}{l}{} & \multicolumn{1}{l}{} \\ \cline{1-6}
\multicolumn{1}{c|}{\multirow{2}{*}{LTD}} & $\mu$ & 3.122 & {\ul 3.027} & 3.199 & \textbf{3.248} & \multicolumn{1}{l}{} & \multicolumn{1}{l}{} & \multicolumn{1}{l}{} & \multicolumn{1}{l}{} & \multicolumn{1}{l}{} \\
\multicolumn{1}{c|}{} & $\sigma$ & 0.745 & {\ul 0.642} & 0.822 & \textbf{1.075} & \multicolumn{1}{l}{} & \multicolumn{1}{l}{} & \multicolumn{1}{l}{} & \multicolumn{1}{l}{} & \multicolumn{1}{l}{} \\ \cline{1-6}
\multicolumn{1}{c|}{\multirow{2}{*}{LTD-50}} & $\mu$ & 2.636 & {\ul 2.628} & 2.668 & \textbf{2.692} & \multicolumn{1}{l}{} & \multicolumn{1}{l}{} & \multicolumn{1}{l}{} & \multicolumn{1}{l}{} & \multicolumn{1}{l}{} \\
\multicolumn{1}{c|}{} & $\sigma$ & 0.431 & {\ul 0.362} & 0.464 & \textbf{0.526} & \multicolumn{1}{l}{} & \multicolumn{1}{l}{} & \multicolumn{1}{l}{} & \multicolumn{1}{l}{} & \multicolumn{1}{l}{} \\ \cline{1-6}
\end{tabular}}
\end{table}
Comparing the expected values, the average attack intensities on different joints are similar. The reason can be that the attack effectiveness is better when the perturbation is larger, and the only boundary in the proposed attack is $\epsilon$. Meanwhile, human motion is systematic for the joints in pose data, so the importance of each joint for a human motion sequence is similar. Therefore, the optimization will maximize the perturbation on every vulnerable joint as much as possible.\par
However, comparing the variances may show the trend of the perturbations on different joints. The variances of distal ends of limbs, like the toe and finger, are significantly larger than the variances of the joints close to the body's center, the neck, which is close to the chest. The variance can be regarded as the space to adjust the perturbation. When the variance is large, the perturbation on the joint receives a more precise adjustment because the perturbation needs to be larger or smaller than it is on other joints to find the optimal position for the attack, which shows the importance of the joint. Hence, the optimized perturbation tends to attack the joints on the distal ends of limbs.\par
To our surprise, the thighs, which are close to the body's center, are other joints suffering from perturbation. As the motion of the leg determines the direction of the whole body, it is clear that the proper perturbation on the thighs can mislead the movement direction of the human motion, which can cause a significant disruption.\par
Comparing the results of LTD and LTD-50, the expected values and the variances of LTD are significantly larger than LTD-50. This observation indicates that redundant joints increase when the input length is enlarged, and their perturbations are relatively mild. Therefore, the perturbation is intense when the joint is vulnerable; otherwise, the perturbation will stay low.\par
Based on the observation in Figure \ref{fig:2}, the fact that the perturbation on the last frame is easier to notice mismatches the similarity of the perturbations' expected values. The reason can be that the momentum in the last frame is obviously deviated by the attack, while the momentum in other frames is close to the original one, which renders the perturbation to notice.\par

\section{Ablation Study} \label{sec:6}
\subsection{Adaptability to Different Scales} \label{sec:6.1}

\begin{table}[htbp]
\centering
\caption{\label{tab:9}The similar attack effectiveness against LTD using the fixed boundaries or the scale function in the proposed scheme. The fixed boundaries $\epsilon_{fix}$ are various and determined by the product of 0.01 and the average scales calculated by the proposed function, which is the same as the experiment setting when $\epsilon$=0.01. This result proves that the proposed function can handle the scale variance among different equipment and adapt to different activities, enhancing the attack's stealthiness.}
\scalebox{0.5}{
\begin{tabular}{ccllllcclllclcllcllclclllccllll}
\hline
\multicolumn{31}{c}{H3.6M} \\ \hline
\multicolumn{1}{c|}{Intervals(ms)}         & \multicolumn{5}{c}{80}   & \multicolumn{5}{c}{160}  & \multicolumn{5}{c}{320}   & \multicolumn{5}{c}{400}   & \multicolumn{5}{c}{560}   & \multicolumn{5}{c}{1000}  \\ \hline
\multicolumn{1}{c|}{Scale($\epsilon$=0.01)}         & \multicolumn{5}{c}{33.7} & \multicolumn{5}{c}{69.9} & \multicolumn{5}{c}{136.3} & \multicolumn{5}{c}{159.7} & \multicolumn{5}{c}{187.4} & \multicolumn{5}{c}{212.6} \\
\multicolumn{1}{c|}{\textit{$\epsilon_{fix}$= 3.0}}    & \multicolumn{5}{c}{32.7} & \multicolumn{5}{c}{68.1} & \multicolumn{5}{c}{133.6} & \multicolumn{5}{c}{157.0} & \multicolumn{5}{c}{185.1} & \multicolumn{5}{c}{211.3} \\ \hline
\multicolumn{31}{c}{CMU} \\ \hline
\multicolumn{1}{c|}{Intervals(ms)}         & \multicolumn{5}{c}{80}   & \multicolumn{5}{c}{160}  & \multicolumn{5}{c}{320}   & \multicolumn{5}{c}{400}   & \multicolumn{5}{c}{560}   & \multicolumn{5}{c}{1000}  \\ \hline
\multicolumn{1}{c|}{Scale($\epsilon$=0.01)}         & \multicolumn{5}{c}{20.9} & \multicolumn{5}{c}{39.5} & \multicolumn{5}{c}{82.6}  & \multicolumn{5}{c}{104.1} & \multicolumn{5}{c}{140.7} & \multicolumn{5}{c}{180.4} \\
\multicolumn{1}{c|}{\textit{$\epsilon_{fix}$= 4.0}}    & \multicolumn{5}{c}{21.1} & \multicolumn{5}{c}{39.8} & \multicolumn{5}{c}{83.3}  & \multicolumn{5}{c}{105.3} & \multicolumn{5}{c}{142.9} & \multicolumn{5}{c}{184.5} \\ \hline
\multicolumn{31}{c}{3DPW} \\ \hline
\multicolumn{1}{c|}{Intervals(ms)}         & \multicolumn{6}{c}{200}         & \multicolumn{6}{c}{400}        & \multicolumn{6}{c}{600}        & \multicolumn{6}{c}{800}        & \multicolumn{6}{c}{1000}       \\ \hline
\multicolumn{1}{c|}{Scale($\epsilon$=0.01)}         & \multicolumn{6}{c}{72.2}        & \multicolumn{6}{c}{138.0}      & \multicolumn{6}{c}{182.0}      & \multicolumn{6}{c}{206.1}      & \multicolumn{6}{c}{211.4}      \\
\multicolumn{1}{c|}{\textit{$\epsilon_{fix}$ = 0.003}} & \multicolumn{6}{c}{73.3}        & \multicolumn{6}{c}{140.4}      & \multicolumn{6}{c}{185.4}      & \multicolumn{6}{c}{210.0}      & \multicolumn{6}{c}{215.3}      \\ \hline
\end{tabular}}
\end{table}
The scale function in the proposed scheme can help the attack method find a proper unit of length automatically without considering the scale parameter of the collecting equipment, which the results in Table \ref{tab:9} can prove. In this study, we remove the scale function, use a fixed parameter $\epsilon_{fix}$ as the boundary and its 0.1 times as the step size, and achieve a set of similar errors as the results when we apply the scale function and set $\epsilon$ = 0.01. The study is all against the long-term LTD on each dataset. Besides the boundary and the step size, the proposed attack method has no more change.\par
According to the results in Table \ref{tab:9}, the scale function can adapt to the scale parameter of the equipment. To determine the value of the boundary $\epsilon_{fix}$, we calculate each dataset's average outcomes of the scale function. The approximate values are 329.5 for H3.6M, 403.0 for CMU, and 0.3 for 3DPW. As the boundary is the product of 0.01 and the calculated scale, $\epsilon_{fix}$ are 3, 4, 0.003 for H3.6M, CMU, and 3DPW, respectively. Note that the length unit of H3.6M, and CMU is a meter, while the unit of 3DPW is a millimeter. Thus, the scale of 3DPW is much different from the others. In the actual scenario, the adversary may not know the scale parameter. Suppose the adversary attacks the human motion data with an inappropriate scale. In that case, the perturbed motion sequence will have a weak attack effectiveness or the perturbed motion will be crushed into a mess. These are the situations we do not want the proposed attack to meet. For the proposed scale function, the scale of the motion sequence is calculated in advance, and the adversary can directly attack the model in a uniform parameter setting, which solves the problem of unknown scale parameters.\par
Comparing the $\epsilon_{fix}$ of H3.6M and CMU, we find that the scale function can also adapt to the range of the input motion. CMU contains more vigorous activities than H3.6M does. Thus, the average scale of CMU is larger than H3.6M. This operation can enhance the effectiveness of the proposed attack while ensuring its stealthiness at the same time. Human perception is based on relative relations, which means humans can tolerate a more significant perturbation in a more vigorous motion, while a small perturbation may be easily found in slow motion.\par
Although the $\epsilon_{fix}$ is determined by the average scale calculated by the proposed function, the errors when using fixed boundary on CMU and 3DPW are obviously higher than when using the scale function. The scale function aims to find basic length units suitable for the input motion sequences. Hence, the outcomes can be various. The mean cannot reflect the mode of most scales, and some activities with high scales in CMU and 3DPW lift the mean of the scales. In such circumstance, most of the activities with lower scales are attacked by the proposed attack but with a larger boundary, which causes the difference between using fixed boundaries and using the scale function.\par
\subsection{Vulnerability of Each Frame} \label{sec:6.2}
\begin{figure}
    \centering
    \includegraphics[width=0.8\textwidth]{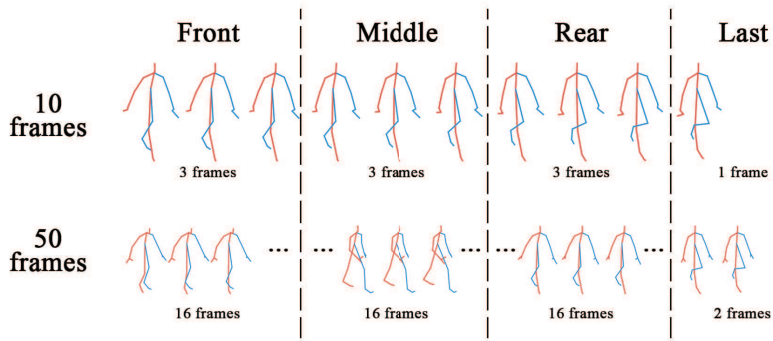}
    \caption{\label{fig:4}This figure shows how to divide the input human motion sequence into the \textit{front}, \textit{middle}, \textit{rear}, and \textit{last} parts when the input length is 10 or 50 frames. The number of frames in each part is written under the poses of the corresponding parts.}
\end{figure}

\begin{table}[htbp]
\centering
\caption{\label{tab:10}The comparison of average prediction errors when we perturb different positions of the input sequence. The bold results are the worst-case of each time interval.}
\scalebox{0.5}{
\begin{tabular}{ccccccc}
\hline
Intervals(ms) & 80 & 160 & 320 & 400 & 560 & 1000 \\ \hline
LTD & 13.5 & 26.8 & 51.9 & 62.0 & 78.5 & 114.3 \\
LTD(front) & 15.2 & 29.4 & 54.5 & 64.5 & 80.7 & 116.3 \\
LTD(middle) & 15.3 & 29.5 & 54.8 & 64.9 & 81.1 & 116.6 \\
LTD(rear) & 21.8 & \textbf{46.5} & \textbf{92.4}  & \textbf{109.4} & \textbf{132.5} & 162.9  \\
LTD(last) & \textbf{22.8} & 45.1 & 87.2 & 104.1 & 128.7 & \textbf{163.6} \\ \hline
TrajCNN & 11.7 & 26.0 & 52.5 & 62.4 & 78.0 & 110.6 \\
TrajCNN(front)  & 11.7 & 26.0 & 52.6 & 62.5 & 78.2 & 110.8 \\
TrajCNN(middle) & 11.7 & 26.0 & 52.6 & 62.5 & 78.2 & 110.8 \\
TrajCNN(rear)   & 18.1 & 41.0 & \textbf{87.0}  & \textbf{104.9} & \textbf{131.5} & 165.8 \\
TrajCNN(last) & \textbf{20.3} & \textbf{41.1} & 84.1           & 102.2 & 130.6 & \textbf{166.3} \\ \hline
HRI & 10.4 & 22.6 & 47.1 & 58.3 & 77.3 & 112.1 \\
HRI(front) & 11.3 & 24.1 & 49.9 & 60.8 & 78.4 & 112.9 \\
HRI(middle) & 11.3 & 24.1 & 49.9 & 60.7 & 78.3 & 112.9 \\
HRI(rear) & 12.8 & 28.6 & 60.8 & 73.9 & 93.1 & 124.8 \\
HRI(last) & \textbf{28.9} & \textbf{64.6} & \textbf{132.4} & \textbf{155.3} & \textbf{178.9} & \textbf{191.6} \\ \hline
\end{tabular}}
\end{table}
To discover the vulnerability of different frames in the input sequence, we divide the input sequence into four parts: \textit{front}, \textit{middle}, \textit{rear}, and \textit{last}, as shown in Figure \ref{fig:4}. For TrajCNN and LTD, which input 10 frames, the input sequence contains nine frames that can be separated into three equal parts without the \textit{last} frame: \textit{front}, \textit{middle}, and \textit{rear}. For HRI, which inputs 50 frames, we set the last two input frames as the \textit{last} part and then segment the rest of the frames into the rest three parts. In this experiment, we set $\epsilon$ to 0.01 and attack the long-term models we train. During the optimization, we mask all frames but the frames in the part we attack, and the results are shown in Table \ref{tab:10}.\par
The vulnerability of the frame increases as the frame gets closer to the observed frame. Compared to the apparent attack effectiveness of the latter two parts, attacking the \textit{front} and \textit{middle} parts can hardly disturb the prediction. Furthermore, attacking the \textit{last} frame sometimes is even more effective than the \textit{rear} frames. The result is clear that the \textit{rear} and \textit{last} frames are more vulnerable than the \textit{front} and \textit{middle} frames, which means the closer a frame is to the observed frame, the stronger attack effectiveness it may generate. Especially, the observed frame is the most important to the adversarial robustness of human motion prediction models, which corresponds to the observation that the last frame takes more weight during the optimization.\par
For LTD and TrajCNN, some activities show that attacking the \textit{rear} part of these activities gives more boost to the disturbance than attacking the \textit{last} part. The reason may be that the frames before the observed frame contribute more pattern information of the human motion. Only perturbing the \textit{last} frame gives momentum to lead the move away, but the remaining parts of the sequence are correct, which may partly remedy the prediction. The thought is also supported by the observation that the sum of increasing amounts when perturbing different parts is lower than the increasing amounts when perturbing the whole input generates.\par
However, HRI may pay more attention to the \textit{last} part because the errors when attacking the \textit{last} part are much higher than the errors when attacking the other parts. Although attacking the \textit{rear} part is relatively more effective than attacking the \textit{front} and \textit{middle} parts, attacking these three parts results in similar errors, which means there is almost no attack effectiveness. Note that the \textit{last} part of HRI’s input is the last two frames, which can contain more motion information than the last frame. Still, this phenomenon is exaggerated for a model that can make use of motion repetitiveness, which also means the contribution of the \textit{last} part is much more than the rest of the frames.\par
\subsection{Effect of Physical Constraints}\label{sec:6.3}
\begin{table}[htbp]
\centering
\caption{\label{tab:11}The comparison of average prediction errors on H3.6M} when the attack is under different constraints. $\overline{\bigtriangleup BL}$ is the average absolute change per bone in the whole test set. $\overline{\bigtriangleup V}$ and $\overline{\bigtriangleup a}$ are the average absolute change per joint of velocity and acceleration in the whole test. The bold results are the highest errors in each time interval and the smallest changes in $\overline{\bigtriangleup BL}$, $\overline{\bigtriangleup V}$, and $\overline{\bigtriangleup a}$, which indicates the best attack effectiveness and the best stealthiness.
\scalebox{0.5}{
\begin{tabular}{cccccccccc}
\hline
Intervals(ms) & 80 & 160 & 320 & 400 & 560 & 1000 & $\overline{\bigtriangleup BL}$ & $\overline{\bigtriangleup V}$ & $\overline{\bigtriangleup a}$ \\ \hline
LTD & 13.5 & 26.8 & 51.9 & 62.0 & 78.5 & 114.3 & - & - & - \\
LTD(no constraint) & \textbf{42.8} & \textbf{90.9} & \textbf{180.9} & \textbf{212.6} & \textbf{248.5} & \textbf{267.7} & 2.67 & 6.07 & 8.86 \\
LTD(temporal constraint) & 37.6 & 77.9 & 150.8 & 175.6 & 204.4 & 224.4 & 1.41 & 1.17 & 1.10 \\
LTD(bone-length constraint) & 36.6 & 76.8 & 150.9 & 177.1 & 207.8 & 232.5 & 1.08 & 3.45 & 4.41 \\
LTD(both constraints) & 33.1 & 68.5 & 133.6 & 156.8 & 184.7 & 210.5 & \textbf{1.00} & \textbf{1.08} & \textbf{1.08} \\ \hline
TrajCNN & 11.7 & 26.0 & 52.5 & 62.4 & 78.0 & 110.6 & - & - & - \\
TrajCNN(no constraint) & \textbf{34.7} & \textbf{76.1} & \textbf{164.1} & \textbf{200.2} & \textbf{249.4} & \textbf{284.6} & 2.67 & 7.52 & 10.82 \\
TrajCNN(temporal constraint) & 31.5 & 66.9 & 141.4 & 172.2 & 215.5 & 249.6 & 1.67 & 1.43 & 1.12 \\
TrajCNN(bone-length constraint) & 32.0 & 69.1 & 146.6 & 178.4 & 223.8 & 259.9 & 0.42 & 4.17 & 5.03 \\
TrajCNN(both constraints) & 29.7 & 63.2 & 132.2 & 160.1 & 199.8 & 233.8 & \textbf{0.40} & \textbf{1.30} & \textbf{1.10}  \\ \hline
HRI & 10.4 & 22.6 & 47.1 & 58.3 & 77.3 & 112.1 & - & - & - \\
HRI(no constraint) & \textbf{39.9} & \textbf{92.3} & \textbf{189.1} & \textbf{218.4} & \textbf{244.6} & \textbf{248.7} & 2.60 & 3.26 & 5.78 \\
HRI(temporal constraint) & 35.4 & 79.8 & 161.9 & 187.4 & 210.8  & 217.6 & 0.62 & 1.02 & \textbf{1.40} \\
HRI(bone-length constraint) & 35.7 & 81.0 & 164.1 & 190.0 & 213.6 & 220.6 & 0.35 & 2.74 & 4.43 \\
HRI(both constraints) & 33.6 & 75.9 & 153.9 & 178.2 & 201.0 & 209.4 & \textbf{0.32} & \textbf{0.96} & 1.41 \\ \hline
\end{tabular}}
\end{table}
To investigate the effect of the proposed physical constraints on the attack effectiveness and imperceptibility, we compare the attack under different constraints on H3.6M. In this section, we use $\overline{\bigtriangleup BL}$, $\overline{\bigtriangleup V}$, and $\overline{\bigtriangleup a}$to analyze the imperceptibility of the attack. $\overline{\bigtriangleup BL}$ is the average absolute change per bone in the whole test set. $\overline{\bigtriangleup V}$ and $\overline{\bigtriangleup a}$ is the average absolute change per joint in the whole test. The lower these three metrics are, the sneakier the adversarial perturbations are. The results are shown in Table \ref{tab:11}. To conveniently compare, we bold the worst errors in each time interval and the least change in each physical metric. In this experiment, we set $\epsilon$ to 0.01 and attack the long-term models we train.\par
According to the result, our constraints weaken the attack effectiveness within an acceptable range and simultaneously keep the adversarial example natural. Note that no constraints mean the original PGD method. Thus, the group with no physical constraints achieves the best attack effectiveness without considering any destructive influence on the human pose, which makes this group easy to be detected. On the contrary, our method with both constraints has the best imperceptibility and a little worse attack effectiveness, which meets the requirements of the attack to be adversarial and quasi-perceptible.\par
According to the comparison of the results from the two constraints, the temporal constraint and the bone-length constraint can complement each other in enhancing the naturalness of the example but not in weakening the attack's effectiveness. Moreover, the temporal constraint is better in these two abilities than the bone-length constraint. The reason may be that the temporal constraint is versatile and can simultaneously attend to the bone length. In contrast, the bong-length constraint is a particular constraint to keep the skeleton with minor deformation that human agents can't notice. Therefore, the two constraints limit the perturbation of a restricted space, which will not weaken the attack effectiveness too much.\par

\section{Discussion} \label{sec:7}
The results presented in the experiments above prove the vulnerability of current human motion predictors and cast light on several key components of the models that are more robust under attack. In the following, we discuss the findings of our research and share insights based on the proposed attack.\par

\subsection{The Factors Influencing Adversarial Robustness}
\label{sec:7.1}
According to the experimental results in Section \ref{sec:5.2}, all the target models are vulnerable when facing the proposed attack. The lowest growth rate of MPJPE is 22\%, while the highest growth rate can reach 295\%, which means all predictions are obviously away from the future human motion sequence. This outcome warns the model designers that convolutional human motion predictors should be more robust. Otherwise, they may cause dangers in their applications. \par
Luckily, the rankings of the target models can help us analyze the factors affecting the adversarial robustness of the target models. The overall ranking of the adversarial robustness from strong to weak is LTD-50, LTD, MH TrajectoryCNN, TrajectoryCNN, TrajectoryCNN-50, and HRI. \par
Comparing MH TrajectoryCNN and TrajectoryCNN, we can directly find that the classification branch boosts the CNN-based structure’s adversarial robustness. In MH TrajectoryCNN, the classification branch helps the model to learn the semantic information of human motion, suppressing the attack’s effectiveness. Furthermore, LTD achieves better results than MH TrajectoryCNN, so we infer that the adjacent matrix in GCN can provide structural prior knowledge of the human body, enhancing the adversarial robustness of the GCN-based predictors. Therefore, we suggest that building prior knowledge and semantic information during training is critical to the adversarial robustness of human motion predictors. \par
Comparing LTD and HRI, we can clearly notice that the attention mechanism hinders the GCN-based structure’s adversarial robustness. Although the attention mechanism can help models capture features dynamically, it may also benefit adversarial attacks. Based on this observation, we suggest that a robust human motion predictor should be aware of utilizing the attention mechanism. \par
Comparing the results between different datasets, we notice that the models trained on 3DPW gain the most adversarial robustness, followed by the CMU and H3.6M models. According to the results in \ref{app:1}, the attack effectiveness on the mild activities is much better than it is on the wild activities, and the attack effectiveness on the repetitive activities is better than it is on the one-time activities, which can be the reason why the H3.6M model achieves worse adversarial robustness. On the one hand, for mild activities, the proposed attack tends to lead the model to output a wild motion, which greatly enlarges the prediction error. On the other hand, for repetitive activities, the model can't recognize the pattern of the adversarial motion sequence, so the predicted motion may be led away from the correct direction. In addition, 3DPW holds the most complex activities that can also be regarded as wild activities, followed by CMU and H3.6M. This ranking is the same as the adversarial robustness ranking of different datasets. As the wild activities can improve the dataset to approach the real data distribution of natural human motions, the model can learn the intense motions from such a dataset and weaken the effect of the adversarial perturbation. Based on this observation, we suggest that the complexity of the activities can enhance the adversarial robustness. Moreover, H3.6M and 3DPW both contain more kinds of activities than CMU, while the CMU models are more robust than the H3.6M models but less robust than the 3DPW models. The reason can be that H3.6M holds more mild and repetitive activities that limit the model from learning real human motion, while 3DPW contains more wild activities that help the model learn more possibilities of real human motion. Therefore, we suggest that the training data should contain more various activities with wild movements, which can help the predictor better comprehend human motion. \par
Based on the result in Table \ref{tab:10}, it is evident that the closer a frame to the observation frame, the more decisive role it plays in human motion prediction, which means the models do not comprehend the motion law of the whole input sequence well. Similar to how humans predict the future motion of others, the prediction should be based on the entire observation of others’ motion rather than extremely decided by the late part of the observation. Therefore, we suggest that improving models’ comprehension of the motion law of the whole input sequence may be a way to improve the adversarial robustness. \par
Besides the insights based on the experimental results, we recommend adversarial training as an outstanding method to enhance the models’ adversarial robustness. Adversarial training utilizes adversarial samples for data augmentation during training, which can help models remedy their weak points in advance. Notably, the proposed attack with a large $\epsilon$ may perturb the last frame of the input sequence so much that the last frame is not continuous with the ground truth. Consequently, we suggest that utilizing the proposed attack with proper parameters for adversarial training is a valid way to enhance the predictors’ adversarial robustness. \par

\subsection{The Advantages and Disadvantages of the Proposed Attack} \label{sec:7.2}
As adversarial attack and defense can promote each other, it is necessary to analyze the proposed attack's advantages and disadvantages to improve the adversarial attack and the adversarial robustness of human motion prediction.\par
\textbf{The advantages.} The advantages of the proposed attack are naturalness and adaptability. Note that the ability to fool the predictor is a basic feature of an adversarial attack, so the destructiveness of the proposed attack is not mentioned as one of its advantages. \par 
The naturalness of the attack indicates that the perturbation has little change to the physical characteristics, like the velocity, of the human motion sequence, which increases the difficulty of noticing the attack for human eyes.\par
The attack's adaptability means that the parameters of the attack are adaptive to the scale of the input data. Moreover, the adaptable scheme helps find a unit to balance the attack's effectiveness and naturalness. \par
\textbf{The disadvantages.} The disadvantage of the proposed attack is that it is unsuitable for adversarial training when the attack is set too intense, which should be noticed for future defenses. Adversarial training requires the adversarial samples to match the ground truth and the human motion sequence to follow natural kinematic principles. The proposed attack with a small $\epsilon$ can meet the requirement of adversarial training for naturalness and matching between the input sequence and the ground truth. However, the proposed attack with a large $\epsilon$ is too intense to maintain the input sequence's natural kinematic completely. In this case, the proposed attack may not be suitable for adversarial training. Once such an adversarial sample is taken for training, the model may encounter difficulties learning the correct kinematic of human motion. \par
\subsection{The Applicability against the Other Methods} \label{sec:7.3}
Although the experiments are implemented on convolution-based human motion predictors, the proposed attack has the potential to fool other types of human motion predictors. Therefore, we will discuss the applicability of the proposed attack on other types of predictors.\par
The target model needs to utilize Cartesian coordination as its input and output to apply the proposed attack. As there are predictors that input and output quaternion\cite{pavllo2020modeling}, Lie algebra\cite{martinez2017on}, or other types of data to represent human poses, the proposed attack cannot directly be applied to these models because the kinematic laws of Cartesian coordination are different from other types of input, like gimbal lock for Euler angle. Except for this condition, the proposed attack can be applied to the target model directly or with slight modification.\par
As the proposed attack is optimizing perturbations to deviate the prediction away from the ground truth, we believe the proposed attack can fool most of the available human motion predictors to a similar extent as the convolution-based models.\par
For RNN-based predictors, the proposed attack may have a similar attack effectiveness for convolution-based models. The spatial and temporal features are critical to human motion prediction. Similarly, convolution-based networks focus more on spatial features, while RNN-based networks focus more on temporal features. Meanwhile, two types of models both try to build spatiotemporal features based on their foundation. What's more, HRI has a recurrent structure in its framework and suffers from the proposed attack. Therefore, RNN-based predictors can be victims of the proposed attack.\par
Since the attention mechanism in HRI weakens its adversarial robustness, the Transformer-based predictors probably suffer significantly from the proposed attack. The Transformer's framework helps the human motion predictor achieve outstanding performance, while it can also support the optimization of the attack to achieve better attack effectiveness by backpropagation.\par
Although MLP-based networks successfully reduce the parameters, the few parameters in the framework may cause their robustness to be weak. When the parameters decrease, the model's ability to describe the reflection from input human motion sequence to ground truth becomes coarse. In this case, the proposed attack can find the optimal perturbation more easily. Therefore, MLP-based predictors may suffer more from the proposed attack than the convolution-based predictors.\par
For the deterministic prediction, generative-based predictors can suffer less from the proposed attack because of their stochasticity. In addition, the GAN-based predictors can also learn the natural kinematics of human motion from the confrontation between the generator and discriminator.\par
Current generative-based predictors are adopted for the probabilistic prediction that predicts multiple possible future human motion sequences. The proposed attack may mislead the predicted human motion sequences away from the ground truth, but we are unsure whether the attack succeeds. Since the goal of probabilistic prediction is to exploit the possibilities of future human motion, the proposed attack may not cover all predicted human motion sequences, and some of the sequences can achieve the goal. Therefore, the attack effectiveness against generative-based probabilistic predictors is not optimistic.

\section{Conclusion}  \label{sec:8}
This paper proposes the \textit{first} adversarial attack against convolution-based human motion prediction and the \textit{first} adversarial robustness evaluation of current convolution-based human motion predictors. The experimental results show the models of human motion prediction are generally vulnerable to adversarial perturbation. As the analytical result, prior knowledge and semantic information can be critical to adversarial robustness. According to the quantitative and qualitative analysis, our adaptable scheme and two physical constraints effectively improve the attack generality across datasets and imperceptibility. In addition, the action prediction result from Multi-Head TrajectoryCNN indicates that the proposed attack barely influences the semantic information of the input sequence. In the end, we discuss the insights summarized from the experimental results, provide suggestions on how to enhance the adversarial robustness of human motion predictors, and analyze the advantages and disadvantages of the proposed attack. In the future, we aim to design a robust human motion predictor and enhance the attack's naturalness for adversarial training, which helps human motion predictors resist adversarial attacks. \par

\section{Acknowledge} \label{sec:9}
This work was supported partly by the Natural Science Foundation of Hainan Province (Grant No. 622RC675), the National Natural Science Foundation of China (Grant No. 62173045, 62273054), and the Fundamental Research Funds for the Central Universities (Grant No. 2020XD-A04-3).

\bibliographystyle{elsarticle-num}
\bibliography{ref}

\begin{thebibliography}{10}
\expandafter\ifx\csname url\endcsname\relax
  \def\url#1{\texttt{#1}}\fi
\expandafter\ifx\csname urlprefix\endcsname\relax\def\urlprefix{URL }\fi
\expandafter\ifx\csname href\endcsname\relax
  \def\href#1#2{#2} \def\path#1{#1}\fi

\bibitem{mao2019learning}
W.~Mao, M.~Liu, M.~Salzmann, H.~Li, Learning trajectory dependencies for human
  motion prediction, in: Proceedings of the IEEE/CVF International Conference
  on Computer Vision, 2019, pp. 9489--9497.

\bibitem{mao2020history}
W.~Mao, M.~Liu, M.~Salzmann, History repeats itself: Human motion prediction
  via motion attention, in: Computer Vision--ECCV 2020: 16th European
  Conference, Glasgow, UK, August 23--28, 2020, Proceedings, Part XIV 16,
  Springer, 2020, pp. 474--489.

\bibitem{liu2020trajectorycnn}
X.~Liu, J.~Yin, J.~Liu, P.~Ding, J.~Liu, H.~Liu, Trajectorycnn: a new
  spatio-temporal feature learning network for human motion prediction, IEEE
  Transactions on Circuits and Systems for Video Technology 31~(6) (2020)
  2133--2146.

\bibitem{ma2022progressively}
T.~Ma, Y.~Nie, C.~Long, Q.~Zhang, G.~Li, Progressively generating better
  initial guesses towards next stages for high-quality human motion prediction,
  in: Proceedings of the IEEE/CVF Conference on Computer Vision and Pattern
  Recognition, 2022, pp. 6437--6446.

\bibitem{li2018convolutional}
C.~Li, Z.~Zhang, W.~S. Lee, G.~H. Lee, Convolutional sequence to sequence model
  for human dynamics, in: Proceedings of the IEEE Conference on Computer Vision
  and Pattern Recognition, 2018, pp. 5226--5234.

\bibitem{sofianos2021space}
T.~Sofianos, A.~Sampieri, L.~Franco, F.~Galasso, Space-time-separable graph
  convolutional network for pose forecasting, in: Proceedings of the IEEE/CVF
  International Conference on Computer Vision, 2021, pp. 11209--11218.

\bibitem{szegedy2014intriguing}
C.~Szegedy, W.~Zaremba, I.~Sutskever, J.~Bruna, D.~Erhan, I.~Goodfellow,
  R.~Fergus, \href{https://openreview.net/forum?id=kklr_MTHMRQjG}{Intriguing
  properties of neural networks}, in: International Conference on Learning
  Representations, 2014.
\newline\urlprefix\url{https://openreview.net/forum?id=kklr_MTHMRQjG}

\bibitem{goodfellow2014explaining}
I.~Goodfellow, J.~Shlens, C.~Szegedy,
  \href{http://arxiv.org/abs/1412.6572}{Explaining and harnessing adversarial
  examples}, in: International Conference on Learning Representations, 2015.
\newline\urlprefix\url{http://arxiv.org/abs/1412.6572}

\bibitem{madry2018towards}
A.~Madry, A.~Makelov, L.~Schmidt, D.~Tsipras, A.~Vladu,
  \href{https://openreview.net/forum?id=rJzIBfZAb}{Towards deep learning models
  resistant to adversarial attacks}, in: International Conference on Learning
  Representations, 2018.
\newline\urlprefix\url{https://openreview.net/forum?id=rJzIBfZAb}

\bibitem{liu2019universal}
H.~Liu, R.~Ji, J.~Li, B.~Zhang, Y.~Gao, Y.~Wu, F.~Huang, Universal adversarial
  perturbation via prior driven uncertainty approximation, in: Proceedings of
  the IEEE/CVF International Conference on Computer Vision, 2019, pp.
  2941--2949.

\bibitem{moosavi2016deepfool}
S.-M. Moosavi-Dezfooli, A.~Fawzi, P.~Frossard, Deepfool: a simple and accurate
  method to fool deep neural networks, in: Proceedings of the IEEE Conference
  on Computer Vision and Pattern Recognition, 2016, pp. 2574--2582.

\bibitem{moosavi2017universal}
S.-M. Moosavi-Dezfooli, A.~Fawzi, O.~Fawzi, P.~Frossard, Universal adversarial
  perturbations, in: Proceedings of the IEEE Conference on Computer Vision and
  Pattern Recognition, 2017, pp. 1765--1773.

\bibitem{carlini2017towards}
N.~Carlini, D.~Wagner, Towards evaluating the robustness of neural networks,
  in: 2017 IEEE Symposium on Security and Privacy, IEEE, 2017, pp. 39--57.

\bibitem{brendel2018decisionbased}
W.~Brendel, J.~Rauber, M.~Bethge,
  \href{https://openreview.net/forum?id=SyZI0GWCZ}{Decision-based adversarial
  attacks: Reliable attacks against black-box machine learning models}, in:
  International Conference on Learning Representations, 2018.
\newline\urlprefix\url{https://openreview.net/forum?id=SyZI0GWCZ}

\bibitem{su2019one}
J.~Su, D.~V. Vargas, K.~Sakurai, One pixel attack for fooling deep neural
  networks, IEEE Transactions on Evolutionary Computation 23~(5) (2019)
  828--841.

\bibitem{luo2022frequency}
C.~Luo, Q.~Lin, W.~Xie, B.~Wu, J.~Xie, L.~Shen, Frequency-driven imperceptible
  adversarial attack on semantic similarity, in: Proceedings of the IEEE/CVF
  Conference on Computer Vision and Pattern Recognition, 2022, pp.
  15315--15324.

\bibitem{he2022transferable}
Z.~He, W.~Wang, J.~Dong, T.~Tan, Transferable sparse adversarial attack, in:
  Proceedings of the IEEE/CVF Conference on Computer Vision and Pattern
  Recognition, 2022, pp. 14963--14972.

\bibitem{chen2021robustness}
Z.~Chen, Y.~Huang, L.~Wang, On the robustness of 3d human pose estimation, in:
  2020 25th International Conference on Pattern Recognition (ICPR), IEEE, 2021,
  pp. 5326--5332.

\bibitem{yufeng2023light}
L.~Yufeng, Y.~Fengyu, L.~Qi, L.~Jiangtao, C.~Chenhong, Light can be dangerous:
  Stealthy and effective physical-world adversarial attack by spot light,
  Computers \& Security (2023) 103345.

\bibitem{schneider2023dual}
J.~Schneider, G.~Apruzzese, Dual adversarial attacks: Fooling humans and
  classifiers, Journal of Information Security and Applications 75 (2023)
  103502.

\bibitem{wang2022ab}
Y.~Wang, J.~Liu, X.~Chang, J.~Wang, R.~J. Rodr{\'\i}guez, Ab-fgsm: Adabelief
  optimizer and fgsm-based approach to generate adversarial examples, Journal
  of Information Security and Applications 68 (2022) 103227.

\bibitem{lu2019switched}
A.-Y. Lu, G.-H. Yang, Switched projected gradient descent algorithms for secure
  state estimation under sparse sensor attacks, Automatica 103 (2019) 503--514.

\bibitem{bryniarski2022evading}
O.~Bryniarski, N.~Hingun, P.~Pachuca, V.~Wang, N.~Carlini,
  \href{https://openreview.net/forum?id=af1eUDdUVz}{Evading adversarial example
  detection defenses with orthogonal projected gradient descent}, in:
  International Conference on Learning Representations, 2022.
\newline\urlprefix\url{https://openreview.net/forum?id=af1eUDdUVz}

\bibitem{cisse2017houdini}
M.~Cisse, Y.~Adi, N.~Neverova, J.~Keshet, Houdini: Fooling deep structured
  visual and speech recognition models with adversarial examples, in:
  Proceedings of the 31st International Conference on Neural Information
  Processing Systems, NIPS'17, Curran Associates Inc., Red Hook, NY, USA, 2017,
  p. 6980–6990.

\bibitem{zhang2022adversarial}
Q.~Zhang, S.~Hu, J.~Sun, Q.~A. Chen, Z.~M. Mao, On adversarial robustness of
  trajectory prediction for autonomous vehicles, in: Proceedings of the
  IEEE/CVF Conference on Computer Vision and Pattern Recognition, 2022, pp.
  15159--15168.

\bibitem{jain2019robustness}
N.~Jain, S.~Shah, A.~Kumar, A.~Jain, On the robustness of human pose
  estimation, in: Proceedings of the IEEE/CVF Conference on Computer Vision and
  Pattern Recognition Workshops, 2019, pp. 29--38.

\bibitem{liu2020adversarial}
J.~Liu, N.~Akhtar, A.~Mian, Adversarial attack on skeleton-based human action
  recognition, IEEE Transactions on Neural Networks and Learning Systems 33~(4)
  (2020) 1609--1622.

\bibitem{wang2021understanding}
H.~Wang, F.~He, Z.~Peng, T.~Shao, Y.-L. Yang, K.~Zhou, D.~Hogg, Understanding
  the robustness of skeleton-based action recognition under adversarial attack,
  in: Proceedings of the IEEE/CVF Conference on Computer Vision and Pattern
  Recognition, 2021, pp. 14656--14665.

\bibitem{diao2021basar}
Y.~Diao, T.~Shao, Y.-L. Yang, K.~Zhou, H.~Wang, Basar: black-box attack on
  skeletal action recognition, in: Proceedings of the IEEE/CVF Conference on
  Computer Vision and Pattern Recognition, 2021, pp. 7597--7607.

\bibitem{zheng2020towards}
T.~Zheng, S.~Liu, C.~Chen, J.~Yuan, B.~Li, K.~Ren, Towards understanding the
  adversarial vulnerability of skeleton-based action recognition, arXiv
  preprint arXiv:2005.07151 (2020).

\bibitem{li2023ats}
X.~Li, Y.~Li, Z.~Feng, Z.~Wang, Q.~Pan, Ats-o2a: A state-based adversarial
  attack strategy on deep reinforcement learning, Computers \& Security 129
  (2023) 103259.

\bibitem{liu2022multi}
X.~Liu, J.~Yin, Multi-head trajectorycnn: A new multi-task framework for action
  prediction, Applied Sciences 12~(11) (2022) 5381.

\bibitem{ionescu2013human3}
C.~Ionescu, D.~Papava, V.~Olaru, C.~Sminchisescu, Human3.6m: Large scale
  datasets and predictive methods for 3d human sensing in natural environments,
  IEEE Transactions on Pattern Analysis and Machine Intelligence 36~(7) (2013)
  1325--1339.

\bibitem{von2018recovering}
T.~Von~Marcard, R.~Henschel, M.~J. Black, B.~Rosenhahn, G.~Pons-Moll,
  Recovering accurate 3d human pose in the wild using imus and a moving camera,
  in: Proceedings of the European Conference on Computer Vision (ECCV), 2018,
  pp. 601--617.

\bibitem{dang2021msr}
L.~Dang, Y.~Nie, C.~Long, Q.~Zhang, G.~Li, Msr-gcn: Multi-scale residual graph
  convolution networks for human motion prediction, in: Proceedings of the
  IEEE/CVF International Conference on Computer Vision, 2021, pp. 11467--11476.

\bibitem{butepage2017deep}
J.~Butepage, M.~J. Black, D.~Kragic, H.~Kjellstrom, Deep representation
  learning for human motion prediction and classification, in: Proceedings of
  the IEEE Conference on Computer Vision and Pattern Recognition, 2017, pp.
  6158--6166.

\bibitem{li2019efficient}
Y.~Li, Z.~Wang, X.~Yang, M.~Wang, S.~I. Poiana, E.~Chaudhry, J.~Zhang,
  Efficient convolutional hierarchical autoencoder for human motion prediction,
  The Visual Computer 35 (2019) 1143--1156.

\bibitem{tang2022temporal}
J.~Tang, J.~Zhang, J.~Yin, Temporal consistency two-stream cnn for human motion
  prediction, Neurocomputing 468 (2022) 245--256.

\bibitem{li2020dynamic}
M.~Li, S.~Chen, Y.~Zhao, Y.~Zhang, Y.~Wang, Q.~Tian, Dynamic multiscale graph
  neural networks for 3d skeleton based human motion prediction, in:
  Proceedings of the IEEE/CVF Conference on Computer Vision and Pattern
  Recognition, 2020, pp. 214--223.

\bibitem{cui2020learning}
Q.~Cui, H.~Sun, F.~Yang, Learning dynamic relationships for 3d human motion
  prediction, in: Proceedings of the IEEE/CVF Conference on Computer Vision and
  Pattern Recognition, 2020, pp. 6519--6527.

\bibitem{aksan2019structured}
E.~Aksan, M.~Kaufmann, O.~Hilliges, Structured prediction helps 3d human motion
  modelling, in: Proceedings of the IEEE/CVF International Conference on
  Computer Vision, 2019, pp. 7144--7153.

\bibitem{lebailly2020motion}
T.~Lebailly, S.~Kiciroglu, M.~Salzmann, P.~Fua, W.~Wang, Motion prediction
  using temporal inception module, in: Proceedings of the Asian Conference on
  Computer Vision, 2020.

\bibitem{cui2021towards}
Q.~Cui, H.~Sun, Towards accurate 3d human motion prediction from incomplete
  observations, in: Proceedings of the IEEE/CVF Conference on Computer Vision
  and Pattern Recognition, 2021, pp. 4801--4810.

\bibitem{li2020multitask}
B.~Li, J.~Tian, Z.~Zhang, H.~Feng, X.~Li, Multitask non-autoregressive model
  for human motion prediction, IEEE Transactions on Image Processing 30 (2020)
  2562--2574.

\bibitem{zhong2022spatio}
C.~Zhong, L.~Hu, Z.~Zhang, Y.~Ye, S.~Xia, Spatio-temporal gating-adjacency gcn
  for human motion prediction, in: Proceedings of the IEEE/CVF Conference on
  Computer Vision and Pattern Recognition, 2022, pp. 6447--6456.

\bibitem{chao2020adversarial}
X.~Chao, Y.~Bin, W.~Chu, X.~Cao, Y.~Ge, C.~Wang, J.~Li, F.~Huang, H.~Leung,
  Adversarial refinement network for human motion prediction, in: Proceedings
  of the Asian Conference on Computer Vision, 2020.

\bibitem{li2021multiscale}
M.~Li, S.~Chen, Y.~Zhao, Y.~Zhang, Y.~Wang, Q.~Tian, Multiscale spatio-temporal
  graph neural networks for 3d skeleton-based motion prediction, IEEE
  Transactions on Image Processing 30 (2021) 7760--7775.

\bibitem{mao2021multi}
W.~Mao, M.~Liu, M.~Salzmann, H.~Li, Multi-level motion attention for human
  motion prediction, International Journal of Computer Vision 129~(9) (2021)
  2513--2535.

\bibitem{fragkiadaki2015recurrent}
K.~Fragkiadaki, S.~Levine, P.~Felsen, J.~Malik, Recurrent network models for
  human dynamics, in: Proceedings of the IEEE international conference on
  computer vision, 2015, pp. 4346--4354.

\bibitem{martinez2017on}
J.~Martinez, M.~J. Black, J.~Romero, On human motion prediction using recurrent
  neural networks, in: Proceedings of the IEEE Conference on Computer Vision
  and Pattern Recognition, 2017.

\bibitem{wolter2018complex}
M.~Wolter, A.~Yao, Complex gated recurrent neural networks, Advances in neural
  information processing systems 31 (2018).

\bibitem{ghosh2017learning}
P.~Ghosh, J.~Song, E.~Aksan, O.~Hilliges, Learning human motion models for
  long-term predictions, in: 2017 International Conference on 3D Vision (3DV),
  IEEE, 2017, pp. 458--466.

\bibitem{gui2018adversarial}
L.-Y. Gui, Y.-X. Wang, X.~Liang, J.~M. Moura, Adversarial geometry-aware human
  motion prediction, in: Proceedings of the European conference on computer
  vision (ECCV), 2018, pp. 786--803.

\bibitem{yao2018multiple}
T.~Yao, M.~Wang, B.~Ni, H.~Wei, X.~Yang, Multiple granularity group interaction
  prediction, in: Proceedings of the IEEE Conference on Computer Vision and
  Pattern Recognition, 2018, pp. 2246--2254.

\bibitem{liu2019towards}
Z.~Liu, S.~Wu, S.~Jin, Q.~Liu, S.~Lu, R.~Zimmermann, L.~Cheng, Towards natural
  and accurate future motion prediction of humans and animals, in: Proceedings
  of the IEEE/CVF Conference on Computer Vision and Pattern Recognition, 2019,
  pp. 10004--10012.

\bibitem{pavllo2020modeling}
D.~Pavllo, C.~Feichtenhofer, M.~Auli, D.~Grangier, Modeling human motion with
  quaternion-based neural networks, International Journal of Computer Vision
  128 (2020) 855--872.

\bibitem{guo2019human}
X.~Guo, J.~Choi, Human motion prediction via learning local structure
  representations and temporal dependencies, in: Proceedings of the AAAI
  Conference on Artificial Intelligence, Vol.~33, 2019, pp. 2580--2587.

\bibitem{chiu2019action}
H.-k. Chiu, E.~Adeli, B.~Wang, D.-A. Huang, J.~C. Niebles, Action-agnostic
  human pose forecasting, in: 2019 IEEE winter conference on applications of
  computer vision (WACV), IEEE, 2019, pp. 1423--1432.

\bibitem{shu2021spatiotemporal}
X.~Shu, L.~Zhang, G.-J. Qi, W.~Liu, J.~Tang, Spatiotemporal co-attention
  recurrent neural networks for human-skeleton motion prediction, IEEE
  Transactions on Pattern Analysis and Machine Intelligence 44~(6) (2021)
  3300--3315.

\bibitem{gopalakrishnan2019neural}
A.~Gopalakrishnan, A.~Mali, D.~Kifer, L.~Giles, A.~G. Ororbia, A neural
  temporal model for human motion prediction, in: Proceedings of the IEEE/CVF
  Conference on Computer Vision and Pattern Recognition, 2019, pp.
  12116--12125.

\bibitem{corona2020context}
E.~Corona, A.~Pumarola, G.~Alenya, F.~Moreno-Noguer, Context-aware human motion
  prediction, in: Proceedings of the IEEE/CVF Conference on Computer Vision and
  Pattern Recognition, 2020, pp. 6992--7001.

\bibitem{dong2021dual}
X.~Dong, C.~Long, W.~Xu, C.~Xiao, Dual graph convolutional networks with
  transformer and curriculum learning for image captioning, in: Proceedings of
  the 29th ACM International Conference on Multimedia, 2021, pp. 2615--2624.

\bibitem{vaswani2017attention}
A.~Vaswani, N.~Shazeer, N.~Parmar, J.~Uszkoreit, L.~Jones, A.~N. Gomez,
  {\L}.~Kaiser, I.~Polosukhin, Attention is all you need, Advances in neural
  information processing systems 30 (2017).

\bibitem{aksan2021spatio}
E.~Aksan, M.~Kaufmann, P.~Cao, O.~Hilliges, A spatio-temporal transformer for
  3d human motion prediction, in: 2021 International Conference on 3D Vision
  (3DV), IEEE, 2021, pp. 565--574.

\bibitem{cai2020learning}
Y.~Cai, L.~Huang, Y.~Wang, T.-J. Cham, J.~Cai, J.~Yuan, J.~Liu, X.~Yang,
  Y.~Zhu, X.~Shen, et~al., Learning progressive joint propagation for human
  motion prediction, in: Computer Vision--ECCV 2020: 16th European Conference,
  Glasgow, UK, August 23--28, 2020, Proceedings, Part VII 16, Springer, 2020,
  pp. 226--242.

\bibitem{martinez2021pose}
A.~Mart{\'\i}nez-Gonz{\'a}lez, M.~Villamizar, J.-M. Odobez, Pose transformers
  (potr): Human motion prediction with non-autoregressive transformers, in:
  Proceedings of the IEEE/CVF International Conference on Computer Vision,
  2021, pp. 2276--2284.

\bibitem{guo2023back}
W.~Guo, Y.~Du, X.~Shen, V.~Lepetit, X.~Alameda-Pineda, F.~Moreno-Noguer, Back
  to mlp: A simple baseline for human motion prediction, in: Proceedings of the
  IEEE/CVF Winter Conference on Applications of Computer Vision, 2023, pp.
  4809--4819.

\bibitem{Bouazizi2022MotionMixer}
A.~Bouazizi, A.~Holzbock, U.~Kressel, K.~Dietmayer, V.~Belagiannis,
  Motionmixer: Mlp-based 3d human body pose forecasting, in: Proceedings of the
  Thirty-First International Joint Conference on Artificial Intelligence,
  {IJCAI-22}, International Joint Conferences on Artificial Intelligence
  Organization, 2022, pp. 791--798.

\bibitem{komura2017recurrent}
T.~Komura, I.~Habibie, D.~Holden, J.~Schwarz, J.~Yearsley,
  \href{https://bmvc2017.london/}{A recurrent variational autoencoder for human
  motion synthesis}, in: The 28th British Machine Vision Conference (BMVC
  2017), 2017, the 28th British Machine Vision Conference , BMVC 2017 ;
  Conference date: 04-09-2017 Through 07-09-2017.
\newblock \href {https://doi.org/10.5244/C.31.119}
  {\path{doi:10.5244/C.31.119}}.
\newline\urlprefix\url{https://bmvc2017.london/}

\bibitem{cai2021unified}
Y.~Cai, Y.~Wang, Y.~Zhu, T.-J. Cham, J.~Cai, J.~Yuan, J.~Liu, C.~Zheng, S.~Yan,
  H.~Ding, et~al., A unified 3d human motion synthesis model via conditional
  variational auto-encoder, in: Proceedings of the IEEE/CVF International
  Conference on Computer Vision, 2021, pp. 11645--11655.

\bibitem{zhang2021we}
Y.~Zhang, M.~J. Black, S.~Tang, We are more than our joints: Predicting how 3d
  bodies move, in: Proceedings of the IEEE/CVF Conference on Computer Vision
  and Pattern Recognition, 2021, pp. 3372--3382.

\bibitem{barsoum2018hp}
E.~Barsoum, J.~Kender, Z.~Liu, Hp-gan: Probabilistic 3d human motion prediction
  via gan, in: Proceedings of the IEEE conference on computer vision and
  pattern recognition workshops, 2018, pp. 1418--1427.

\bibitem{hernandez2019human}
A.~Hernandez, J.~Gall, F.~Moreno-Noguer, Human motion prediction via
  spatio-temporal inpainting, in: Proceedings of the IEEE/CVF International
  Conference on Computer Vision, 2019, pp. 7134--7143.

\bibitem{yuan2020dlow}
Y.~Yuan, K.~Kitani, Dlow: Diversifying latent flows for diverse human motion
  prediction, in: Computer Vision--ECCV 2020: 16th European Conference,
  Glasgow, UK, August 23--28, 2020, Proceedings, Part IX 16, Springer, 2020,
  pp. 346--364.

\bibitem{barquero2023belfusion}
G.~Barquero, S.~Escalera, C.~Palmero, Belfusion: Latent diffusion for
  behavior-driven human motion prediction, in: Proceedings of the IEEE/CVF
  International Conference on Computer Vision, 2023, pp. 2317--2327.

\bibitem{wei2023human}
D.~Wei, H.~Sun, B.~Li, J.~Lu, W.~Li, X.~Sun, S.~Hu, Human joint kinematics
  diffusion-refinement for stochastic motion prediction, in: Proceedings of the
  AAAI Conference on Artificial Intelligence, Vol.~37, 2023, pp. 6110--6118.

\end{thebibliography}

\appendix
\setcounter{table}{0}   
\renewcommand{\thetable}{A\arabic{table}}

\section{Detailed prediction errors of the target models}
\label{app:1}
\begin{table}[H]
\centering
\caption{\label{Atab:1}Short-term prediction errors of LTD on H3.6M before and after perturbation in variations of the boundary.}
\scalebox{0.5}{
\begin{tabular}{c|cccc|cccc|cccc|cccc}
\hline
Intervals(ms) & 80 & 160 & 320 & 400 & 80 & 160 & 320 & 400 & 80 & 160 & 320 & 400 & 80 & 160 & 320 & 400 \\ \hline
scenarios & \multicolumn{4}{c|}{Walking} & \multicolumn{4}{c|}{Eating} & \multicolumn{4}{c|}{Smoking} & \multicolumn{4}{c}{Discussion} \\ \hline
LTD & 8.9 & 15.7 & 29.2 & 33.4 & 8.8 & 18.9 & 39.4 & 47.2 & 7.8 & 14.9 & 25.3 & 28.8 & 9.8 & 22.1 & 39.6 & 44.1 \\
$\epsilon$=0.01 & 35.5 & 71.0 & 123.2 & 137.0 & 29.6 & 61.4 & 114.9 & 130.6 & 27.0 & 56.5 & 104.8 & 119.0 & 32.9 & 70.6 & 123.2 & 130.8 \\
$\epsilon$=0.02 & 49.1 & 97.5 & 171.1 & 189.4 & 41.4 & 83.0 & 152.4 & 173.6 & 36.1 & 74.5 & 140.0 & 159.2 & 42.8 & 87.6 & 144.9 & 150.5 \\
$\epsilon$=0.03 & 54.8 & 109.2 & 202.5 & 225.2 & 49.6 & 98.1 & \textbf{174.5} & \textbf{195.6} & 45.4 & 94.2 & \textbf{179.1} & \textbf{203.0} & 51.7 & 105.5 & 186.1 & 201.0 \\
$\epsilon$=0.04 & 62.5 & 122.5 & 212.9 & 230.7 & 52.3 & 98.2 & 168.7 & 188.2 & 48.5 & 97.0 & 173.3 & 192.7 & 57.8 & 117.1 & 201.9 & 215.6 \\
$\epsilon$=0.05 & \textbf{63.8} & \textbf{124.1} & \textbf{223.0} & \textbf{246.3} & \textbf{55.8} & \textbf{101.5} & 164.2 & 178.1 & \textbf{51.7} & \textbf{101.5} & 178.6 & 196.8 & \textbf{61.7} & \textbf{123.3} & \textbf{212.8} & \textbf{227.7} \\
\hline
\end{tabular}}
\end{table}

\begin{table}[H]
\centering
\caption{\label{Atab:2}Long-term prediction errors of LTD on H3.6M before and after perturbation in variations of the boundary.}
\scalebox{0.5}{
\begin{tabular}{c|cc|cc|cc|cc|cc|cc|cc|cc}
\hline
Intervals(ms) & 560 & 1000 & 560 & 1000 & 560 & 1000 & 560 & 1000 & 560 & 1000 & 560 & 1000 & 560 & 1000 & 560 & 1000 \\ \hline
scenarios & \multicolumn{2}{c|}{Walking} & \multicolumn{2}{c|}{Eating} & \multicolumn{2}{c|}{Smoking} & \multicolumn{2}{c|}{Discussion} & \multicolumn{2}{c|}{Direction} & \multicolumn{2}{c|}{Greeting} & \multicolumn{2}{c|}{Phoning} & \multicolumn{2}{c}{Posing} \\ \hline
LTD & 42.3 & 51.4 & 56.5 & 68.6 & 32.3 & 60.5 & 70.5 & 103.5 & 85.7 & 109.3 & 91.8 & 87.4 & 65.1 & 113.7 & 113.3 & 220.4 \\
$\epsilon$=0.01 & 109.2 & 121.3 & 142.6 & 137.2 & 138.0 & 159.2 & 175.9 & 208.0 & 151.7 & 162.6 & 198.4 & 174.5 & 131.3 & 165.0 & 259.7 & 331.1 \\
$\epsilon$=0.02 & 166.6 & 182.1 & 204.8 & 188.1 & 204.0 & 225.5 & 208.5 & 240.7 & 171.2 & 177.0 & 234.3 & 206.0 & 164.1 & 194.3 & 289.0 & 345.0 \\
$\epsilon$=0.03 & 172.7 & 183.3 & 193.8 & 201.7 & 215.5 & 224.2 & 221.1 & \textbf{255.5} & 171.1 & 183.2 & 248.0 & 216.0 & 210.8 & 236.5 & 313.3 & \textbf{362.5} \\
$\epsilon$=0.04 & 193.0 & \textbf{207.8} & 227.7 & 229.1 & 244.5 & \textbf{264.8} & 225.2 & 251.9 & 169.6 & 178.5 & 256.8 & 221.0 & 224.2 & \textbf{255.5} & \textbf{322.4} & 349.4 \\
$\epsilon$=0.05 & \textbf{226.3} & 202.1 & \textbf{250.2} & \textbf{240.2} & \textbf{245.3} & 259.9 & \textbf{227.2} & 252.6 & \textbf{182.5} & \textbf{191.7} & \textbf{262.6} & \textbf{236.5} & \textbf{234.9} & 254.6 & 294.1 & 324.8 \\
\hline
\end{tabular}}
\end{table}

\begin{table}[H]
\centering
\caption{\label{Atab:3}Short-term prediction errors of LTD-50 on H3.6M before and after perturbation in variations of the boundary.}
\scalebox{0.5}{
\begin{tabular}{c|cccc|cccc|cccc|cccc}
\hline
Intervals(ms) & 80 & 160 & 320 & 400 & 80 & 160 & 320 & 400 & 80 & 160 & 320 & 400 & 80 & 160 & 320 & 400 \\ \hline
scenarios & \multicolumn{4}{c|}{Walking} & \multicolumn{4}{c|}{Eating} & \multicolumn{4}{c|}{Smoking} & \multicolumn{4}{c}{Discussion} \\ \hline
LTD-50 & 11.3 & 18.7 & 28.3 & 35.2 & 10.0 & 15.4 & 28.3 & 35.4 & 8.5 & 13.7 & 25.0 & 29.1 & 15.9 & 37.8 & 81.5 & 104.4 \\
$\epsilon$=0.01 & 23.8 & 38.5 & 51.3 & 59.0 & 24.7 & 40.7 & 57.6 & 64.7 & 14.4 & 24.9 & 47.2 & 53.0 & 32.6 & 65.5 & 121.8 & 144.7 \\
$\epsilon$=0.02 & 29.9 & 50.3 & 68.5 & 75.3 & 30.3 & 53.5 & 81.7 & 90.5 & 24.4 & 47.5 & 90.2 & 99.3 & 38.3 & 76.3 & 140.3 & 161.2 \\
$\epsilon$=0.03 & 37.9 & 67.8 & 103.3 & 108.0 & 37.5 & 69.8 & 114.5 & 125.4 & 33.6 & 67.0 & 125.0 & 135.7 & 46.2 & 93.6 & 168.2 & 190.2 \\
$\epsilon$=0.04 & 46.1 & 83.3 & 127.7 & 134.7 & 44.5 & 83.0 & 136.2 & 148.4 & 39.3 & 78.8 & 143.8 & 155.5 & 49.6 & 100.3 & 181.9 & 205.5 \\
$\epsilon$=0.05 & \textbf{52.1} & \textbf{92.9} & \textbf{140.1} & \textbf{147.5} & \textbf{48.8} & \textbf{89.8} & \textbf{145.2} & \textbf{157.0} & \textbf{44.2} & \textbf{86.3} & \textbf{156.2} & \textbf{168.1} & \textbf{60.0} & \textbf{120.5} & \textbf{217.9} & \textbf{241.7} \\
\hline
\end{tabular}}
\end{table}

\begin{table}[H]
\centering
\caption{\label{Atab:4}Long-term prediction errors of LTD-50 on H3.6M before and after perturbation in variations of the boundary.}
\scalebox{0.5}{
\begin{tabular}{c|cc|cc|cc|cc|cc|cc|cc|cc}
\hline
Intervals(ms) & 560 & 1000 & 560 & 1000 & 560 & 1000 & 560 & 1000 & 560 & 1000 & 560 & 1000 & 560 & 1000 & 560 & 1000 \\ \hline
scenarios & \multicolumn{2}{c|}{Walking} & \multicolumn{2}{c|}{Eating} & \multicolumn{2}{c|}{Smoking} & \multicolumn{2}{c|}{Discussion} & \multicolumn{2}{c|}{Direction} & \multicolumn{2}{c|}{Greeting} & \multicolumn{2}{c|}{Phoning} & \multicolumn{2}{c}{Posing} \\ \hline
LTD-50 & 39.0 & 47.0 & 53.0 & 74.3 & 36.0 & 54.5 & 119.7 & 128.4 & 95.4 & 123.1 & 104.5 & 132.2 & 90.1 & 120.7 & 157.6 & 228.2 \\
$\epsilon$=0.01 & 52.4 & 57.9 & 96.7 & 124.6 & 121.9 & 144.8 & 174.6 & 167.6 & 142.7 & 168.3 & 135.5 & 148.6 & 127.2 & 160.3 & 203.0 & 263.2 \\
$\epsilon$=0.02 & 68.2 & 69.4 & 137.6 & 163.8 & 192.3 & 221.7 & 218.9 & 203.8 & 178.8 & 195.7 & 158.4 & 167.8 & 160.7 & 194.5 & 216.4 & 275.9 \\
$\epsilon$=0.03 & 89.7 & 90.7 & 153.6 & 173.7 & 214.9 & 241.3 & 226.3 & 202.8 & 199.8 & 220.8 & 183.4 & 187.9 & 188.7 & 215.5 & 230.7 & 290.2 \\
$\epsilon$=0.04 & 121.9 & \textbf{118.2} & 165.9 & \textbf{188.9} & 233.8 & 245.1 & \textbf{259.1} & 225.5 & 223.0 & 237.3 & 192.4 & 200.1 & 202.4 & \textbf{229.7} & 232.9 & 292.0 \\
$\epsilon$=0.05 & \textbf{122.3} & 112.6 & \textbf{173.5} & 188.7 & \textbf{258.8} & \textbf{284.4} & 255.6 & \textbf{228.0} & \textbf{232.0} & \textbf{248.2} & \textbf{201.7} & \textbf{203.3} & \textbf{204.2} & 223.8 & \textbf{261.6} & \textbf{315.7} \\
\hline
\end{tabular}}
\end{table}

\begin{table}[H]
\centering
\caption{\label{Atab:5}Short-term prediction errors of TrajectoryCNN on H3.6M before and after perturbation in variations of boundary.}
\scalebox{0.5}{
\begin{tabular}{c|cccc|cccc|cccc|cccc}
\hline
Intervals(ms) & 80 & 160 & 320 & 400 & 80 & 160 & 320 & 400 & 80 & 160 & 320 & 400 & 80 & 160 & 320 & 400 \\ \hline
scenarios & \multicolumn{4}{c|}{Walking} & \multicolumn{4}{c|}{Eating} & \multicolumn{4}{c|}{Smoking} & \multicolumn{4}{c}{Discussion} \\ \hline
TrajCNN & 8.3 & 15.2 & 31.3 & 36.3 & 8.6 & 18.9 & 38.0 & 45.6 & 6.3 & 12.8 & 23.7 & 28.0 & 7.8 & 20.3 & 41.7 & 48.0 \\
$\epsilon$=0.01 & 37.7 & 85.4 & 164.3 & 177.1 & 38.8 & 87.9 & 175.7 & 194.5 & 29.2 & 64.8 & 130.7 & 148.2 & 42.3 & 97.7 & 189.7 & 205.4 \\
$\epsilon$=0.02 & 60.5 & 134.1 & 259.1 & 279.9 & 55.1 & 120.9 & 231.2 & 249.4 & 43.8 & 95.4 & 186.9 & 211.2 & 62.2 & 139.0 & 259.2 & 275.5 \\
$\epsilon$=0.03 & 71.8 & 155.2 & 298.3 & 319.5 & 64.3 & 136.7 & 252.9 & 270.8 & 51.2 & 110.1 & 214.6 & 240.1 & 70.3 & 150.4 & 273.9 & 293.4 \\
$\epsilon$=0.04 & 75.6 & 160.6 & 299.6 & 318.0 & 68.4 & 142.3 & 259.5 & 277.1 & 53.5 & 112.8 & 218.1 & \textbf{243.8} & 76.9 & 165.0 & \textbf{302.0} & \textbf{325.7} \\
$\epsilon$=0.05 & \textbf{83.3} & \textbf{180.4} & \textbf{345.6} & \textbf{371.9} & \textbf{72.9} & \textbf{151.6} & \textbf{277.7} & \textbf{296.3} & \textbf{57.3} & \textbf{117.7} & \textbf{220.5} & 242.9 & \textbf{80.4} & \textbf{167.8} & 299.6 & 319.1 \\
\hline
\end{tabular}}
\end{table}

\begin{table}[H]
\centering
\caption{\label{Atab:6}Long-term prediction errors of TrajectoryCNN on H3.6M before and after perturbation in variations of the boundary.}
\scalebox{0.5}{
\begin{tabular}{c|cc|cc|cc|cc|cc|cc|cc|cc}
\hline
Intervals(ms) & 560 & 1000 & 560 & 1000 & 560 & 1000 & 560 & 1000 & 560 & 1000 & 560 & 1000 & 560 & 1000 & 560 & 1000 \\ \hline
scenarios & \multicolumn{2}{c|}{Walking} & \multicolumn{2}{c|}{Eating} & \multicolumn{2}{c|}{Smoking} & \multicolumn{2}{c|}{Discussion} & \multicolumn{2}{c|}{Direction} & \multicolumn{2}{c|}{Greeting} & \multicolumn{2}{c|}{Phoning} & \multicolumn{2}{c}{Posing} \\ \hline
TrajCNN & 38.5 & 46.0 & 59.9 & 71.3 & 33.1 & 60.2 & 79.5 & 109.4 & 78.4 & 102.9 & 89.3 & 85.2 & 61.4 & 111.3 & 118.7 & 219.6 \\
$\epsilon$=0.01 & 88.2 & 82.0 & 187.4 & 205.5 & 183.5 & 227.7 & 200.7 & 236.3 & 224.0 & 270.4 & 233.4 & 240.9 & 145.9 & 183.9 & 274.8 & 365.1 \\
$\epsilon$=0.02 & 134.4 & 122.2 & 251.9 & 279.5 & 227.8 & 268.6 & 274.5 & 317.6 & 264.9 & 301.6 & 294.9 & 296.1 & 197.0 & 222.8 & 305.2 & 357.4 \\
$\epsilon$=0.03 & 226.9 & 200.9 & \textbf{277.0} & 291.8 & 246.2 & \textbf{284.1} & 290.3 & 324.4 & \textbf{313.7} & \textbf{350.8} & 317.5 & 310.2 & 241.6 & 259.3 & 347.3 & 371.0 \\
$\epsilon$=0.04 & \textbf{264.7} & \textbf{253.7} & 272.4 & \textbf{299.1} & 245.0 & 277.5 & 286.1 & 321.8 & 255.4 & 252.3 & \textbf{335.6} & \textbf{333.4} & \textbf{246.2} & \textbf{262.6} & 368.1 & 347.1 \\
$\epsilon$=0.05 & 193.2 & 177.6 & 262.0 & 286.1 & \textbf{249.3} & 283.0 & \textbf{320.1} & \textbf{360.3} & 270.5 & 274.0 & 333.2 & 324.3 & 235.9 & 257.1 & \textbf{389.4} & \textbf{394.7} \\
\hline
\end{tabular}}
\end{table}

\begin{table}[H]
\centering
\caption{\label{Atab:7}Short-term prediction errors of TrajectoryCNN-50 on H3.6M before and after perturbation in variations of the boundary.}
\scalebox{0.5}{
\begin{tabular}{c|cccc|cccc|cccc|cccc}
\hline
Intervals(ms) & 80 & 160 & 320 & 400 & 80 & 160 & 320 & 400 & 80 & 160 & 320 & 400 & 80 & 160 & 320 & 400 \\ \hline
scenarios & \multicolumn{4}{c|}{Walking} & \multicolumn{4}{c|}{Eating} & \multicolumn{4}{c|}{Smoking} & \multicolumn{4}{c}{Discussion} \\ \hline
TrajCNN-50 & 8.8 & 16.6 & 23.1 & 29.9 & 7.0 & 12.4 & 25.3 & 32.1 & 6.9 & 11.5 & 21.9 & 25.7 & 10.4 & 29.6 & 72.1 & 94.7 \\
$\epsilon$=0.01 & 32.4 & 67.7 & 124.4 & 136.3 & 31.2 & 67.6 & 134.9 & 152.6 & 28.1 & 61.3 & 128.3 & 146.5 & 39.9 & 96.5 & 200.5 & 232.7 \\
$\epsilon$=0.02 & 54.6 & 113.7 & 212.4 & 230.6 & 51.1 & 107.5 & 203.0 & 223.6 & 43.9 & 95.1 & 196.6 & 225.0 & 61.2 & 135.8 & 268.2 & 303.8 \\
$\epsilon$=0.03 & 70.2 & 146.6 & 263.3 & 279.9 & 63.5 & 129.4 & 238.9 & 261.3 & 55.1 & 115.7 & 235.9 & 269.1 & 74.6 & 161.5 & 308.1 & 341.9 \\
$\epsilon$=0.04 & 81.3 & 166.6 & 312.3 & 339.0 & 74.0 & 148.2 & 269.8 & 293.6 & 61.9 & 125.1 & 244.6 & 275.1 & 87.7 & 183.3 & 350.3 & 388.9 \\
$\epsilon$=0.05 & \textbf{96.2} & \textbf{195.9} & \textbf{351.4} & \textbf{372.2} & \textbf{82.5} & \textbf{160.7} & \textbf{278.7} & \textbf{296.4} & \textbf{74.0} & \textbf{150.8} & \textbf{298.2} & \textbf{336.6} & \textbf{98.8} & \textbf{201.3} & \textbf{373.5} & \textbf{410.9} \\
\hline
\end{tabular}}
\end{table}

\begin{table}[H]
\centering
\caption{\label{Atab:8}Long-term prediction errors of TrajectoryCNN-50 on H3.6M before and after perturbation in variations of the boundary.}
\scalebox{0.5}{
\begin{tabular}{c|cc|cc|cc|cc|cc|cc|cc|cc}
\hline
Intervals(ms) & 560 & 1000 & 560 & 1000 & 560 & 1000 & 560 & 1000 & 560 & 1000 & 560 & 1000 & 560 & 1000 & 560 & 1000 \\ \hline
scenarios & \multicolumn{2}{c|}{Walking} & \multicolumn{2}{c|}{Eating} & \multicolumn{2}{c|}{Smoking} & \multicolumn{2}{c|}{Discussion} & \multicolumn{2}{c|}{Direction} & \multicolumn{2}{c|}{Greeting} & \multicolumn{2}{c|}{Phoning} & \multicolumn{2}{c}{Posing} \\ \hline
TrajCNN-50 & 30.8 & 35.8 & 44.8 & 66.2 & 40.7 & 58.3 & 121.3 & 124.4 & 97.3 & 122.4 & 114.2 & 144.3 & 76.2 & 110.0 & 125.3 & 191.0 \\
$\epsilon$=0.01 & 45.2 & 48.2 & 120.6 & 149.1 & 143.1 & 170.3 & 187.8 & 179.2 & 207.3 & 228.3 & 167.2 & 198.4 & 136.4 & 169.6 & 184.6 & 223.9 \\
$\epsilon$=0.02 & 78.4 & 84.7 & 182.3 & 206.6 & 233.0 & 265.5 & 253.1 & 211.2 & 279.2 & 290.3 & 225.6 & 246.7 & 205.6 & 240.3 & 257.3 & 278.9 \\
$\epsilon$=0.03 & 151.2 & 144.8 & 223.6 & 234.9 & 270.1 & 295.5 & 310.7 & 267.1 & 307.2 & 306.0 & 262.2 & 276.9 & 248.0 & 277.2 & 311.9 & 320.7 \\
$\epsilon$=0.04 & 193.8 & 186.4 & 244.4 & 245.2 & 294.0 & 308.7 & 339.6 & 305.4 & \textbf{383.0} & \textbf{397.8} & 310.9 & 323.8 & 281.9 & 313.1 & 362.1 & 353.1 \\
$\epsilon$=0.05 & \textbf{272.9} & \textbf{262.2} & \textbf{310.7} & \textbf{330.2} & \textbf{318.3} & \textbf{339.5} & \textbf{386.4} & \textbf{362.4} & 368.9 & 363.8 & \textbf{345.9} & \textbf{365.0} & \textbf{317.0} & \textbf{340.2} & \textbf{399.6} & \textbf{380.1} \\
\hline
\end{tabular}}
\end{table}

\begin{table}[H]
\centering
\caption{\label{Atab:9}Short-term prediction errors of HRI on H3.6M before and after perturbation in variations of the boundary.}
\scalebox{0.5}{
\begin{tabular}{c|cccc|cccc|cccc|cccc}
\hline
Intervals(ms) & 80 & 160 & 320 & 400 & 80 & 160 & 320 & 400 & 80 & 160 & 320 & 400 & 80 & 160 & 320 & 400 \\ \hline
scenarios & \multicolumn{4}{c|}{Walking} & \multicolumn{4}{c|}{Eating} & \multicolumn{4}{c|}{Smoking} & \multicolumn{4}{c}{Discussion} \\ \hline
HRI & 10.0 & 19.5 & 34.2 & 39.8 & 6.4 & 14.0 & 28.7 & 36.2 & 7.0 & 14.9 & 29.9 & 36.4 & 10.2 & 23.4 & 52.1 & 65.4 \\
$\epsilon$=0.01 & 33.5 & 69.0 & 116.7 & 129.1 & 29.9 & 68.3 & 136.1 & 155.6 & 32.1 & 74.5 & 156.6 & 185.4 & 30.2 & 68.6 & 131.0 & 148.6 \\
$\epsilon$=0.02 & 48.3 & 97.8 & 174.1 & 193.3 & 43.8 & 96.1 & 181.7 & 203.4 & 42.3 & 92.6 & 185.4 & 215.0 & 40.2 & 85.2 & 152.2 & 168.1 \\
$\epsilon$=0.03 & 58.5 & 116.3 & 202.8 & 219.9 & 53.9 & 112.3 & 207.4 & 232.2 & 56.6 & 121.0 & 228.3 & 259.1 & 51.9 & 103.3 & 176.7 & 192.7 \\
$\epsilon$=0.04 & 67.7 & 129.8 & 220.2 & 241.2 & 59.7 & 120.6 & 214.5 & 236.0 & 64.2 & 131.7 & 241.2 & 269.6 & 61.2 & 120.1 & 199.8 & 214.2 \\
$\epsilon$=0.05 & \textbf{71.5} & \textbf{135.7} & \textbf{232.2} & \textbf{256.7} & \textbf{67.9} & \textbf{131.5} & \textbf{230.9} & \textbf{254.7} & \textbf{76.0} & \textbf{152.1} & \textbf{265.4} & \textbf{291.4} & \textbf{67.4} & \textbf{128.3} & \textbf{215.9} & \textbf{235.5} \\
\hline
\end{tabular}}
\end{table}

\begin{table}[H]
\centering
\caption{\label{Atab:10}Long-term prediction errors of HRI on H3.6M before and after perturbation in variations of the boundary.}
\scalebox{0.5}{
\begin{tabular}{c|cc|cc|cc|cc|cc|cc|cc|cc}
\hline
Intervals(ms) & 560 & 1000 & 560 & 1000 & 560 & 1000 & 560 & 1000 & 560 & 1000 & 560 & 1000 & 560 & 1000 & 560 & 1000 \\ \hline
scenarios & \multicolumn{2}{c|}{Walking} & \multicolumn{2}{c|}{Eating} & \multicolumn{2}{c|}{Smoking} & \multicolumn{2}{c|}{Discussion} & \multicolumn{2}{c|}{Direction} & \multicolumn{2}{c|}{Greeting} & \multicolumn{2}{c|}{Phoning} & \multicolumn{2}{c}{Posing} \\ \hline
HRI & 47.4 & 58.1 & 50.0 & 75.7 & 47.6 & 69.5 & 86.6 & 119.8 & 73.9 & 106.5 & 101.9 & 138.8 & 67.4 & 105.0 & 107.6 & 178.2 \\
$\epsilon$=0.01 & 131.3 & 116.5 & 170.3 & 154.3 & 215.0 & 217.1 & 167.1 & 155.3 & 183.5 & 182.5 & 235.0 & 209.8 & 135.6 & 168.7 & 238.9 & 269.5 \\
$\epsilon$=0.02 & 196.8 & 171.7 & 217.2 & 198.2 & 240.8 & 240.4 & 193.4 & 207.9 & 229.1 & 217.7 & 286.7 & 253.1 & 195.5 & 200.4 & 306.2 & 314.1 \\
$\epsilon$=0.03 & 218.6 & 201.6 & 247.8 & 222.1 & 280.9 & 274.5 & 210.4 & 218.5 & 225.3 & 213.6 & 317.0 & 276.9 & 238.6 & 228.3 & 346.9 & 332.7 \\
$\epsilon$=0.04 & 234.6 & 199.8 & 247.4 & 231.9 & 286.6 & 274.0 & 235.6 & \textbf{248.7} & 266.6 & 245.1 & 316.9 & 266.3 & 224.7 & 229.8 & \textbf{376.2} & \textbf{359.2} \\
$\epsilon$=0.05 & \textbf{262.9} & \textbf{224.4} & \textbf{261.7} & \textbf{238.0} & \textbf{302.9} & \textbf{284.6} & \textbf{250.2} & 213.6 & \textbf{271.2} & \textbf{251.8} & \textbf{342.7} & \textbf{298.8} & \textbf{244.5} & \textbf{244.4} & 354.5 & 350.3 \\
\hline
\end{tabular}}
\end{table}

\begin{table}[H]
\centering
\caption{\label{Atab:11}Short-term prediction errors of Multi-Head TrajectoryCNN on H3.6M before and after perturbation in variations of the boundary.}
\scalebox{0.5}{
\begin{tabular}{c|cccc|cccc|cccc|cccc}
\hline
Intervals(ms) & 80 & 160 & 320 & 400 & 80 & 160 & 320 & 400 & 80 & 160 & 320 & 400 & 80 & 160 & 320 & 400 \\ \hline
scenarios & \multicolumn{4}{c|}{Walking} & \multicolumn{4}{c|}{Eating} & \multicolumn{4}{c|}{Smoking} & \multicolumn{4}{c}{Discussion} \\ \hline
MH-TrajCNN & 10.3 & 18.7 & 26.7 & 33.7 & 8.5 & 15.7 & 29.0 & 36.1 & 8.5 & 14.1 & 26.3 & 30.0 & 13.7 & 36.9 & 80.9 & 100.5 \\
$\epsilon$=0.01 & 21.7 & 35.7 & 48.9 & 52.7 & 22.7 & 41.1 & 69.6 & 77.6 & 22.9 & 40.9 & 72.3 & 80.0 & 31.2 & 69.1 & 132.4 & 153.1 \\
$\epsilon$=0.02 & 40.0 & 67.7 & 102.6 & 107.3 & 40.5 & 74.4 & 127.6 & 140.1 & 42.4 & 80.4 & 143.0 & 157.2 & 51.1 & 105.6 & 190.9 & 213.9 \\
$\epsilon$=0.03 & 55.8 & 94.1 & 140.1 & 142.9 & 55.0 & 99.3 & 160.7 & 172.3 & 56.3 & 106.2 & 186.2 & 203.9 & 66.2 & 130.3 & 223.0 & 247.8 \\
$\epsilon$=0.04 & 72.5 & 125.3 & 184.9 & 187.0 & 66.8 & 118.7 & 192.1 & 204.8 & 68.7 & 127.8 & 215.4 & 233.0 & 81.3 & 156.6 & 265.4 & 291.1 \\
$\epsilon$=0.05 & \textbf{92.0} & \textbf{160.3} & \textbf{240.3} & \textbf{245.2} & \textbf{80.4} & \textbf{140.0} & \textbf{214.9} & \textbf{225.9} & \textbf{79.4} & \textbf{146.1} & \textbf{243.8} & \textbf{264.3} & \textbf{92.5} & \textbf{168.8} & \textbf{275.8} & \textbf{300.1} \\
\hline
\end{tabular}}
\end{table}

\begin{table}[H]
\centering
\caption{\label{Atab:12}Long-term prediction errors of Multi-head TrajectoryCNN on H3.6M before and after perturbation in variations of the boundary.}
\scalebox{0.5}{
\begin{tabular}{c|cc|cc|cc|cc|cc|cc|cc|cc}
\hline
Intervals(ms) & 560 & 1000 & 560 & 1000 & 560 & 1000 & 560 & 1000 & 560 & 1000 & 560 & 1000 & 560 & 1000 & 560 & 1000 \\ \hline
scenarios & \multicolumn{2}{c|}{Walking} & \multicolumn{2}{c|}{Eating} & \multicolumn{2}{c|}{Smoking} & \multicolumn{2}{c|}{Discussion} & \multicolumn{2}{c|}{Direction} & \multicolumn{2}{c|}{Greeting} & \multicolumn{2}{c|}{Phoning} & \multicolumn{2}{c}{Posing} \\ \hline
MH-TrajCNN & 32.7 & 36.9 & 49.5 & 69.6 & 37.4 & 57.2 & 121.3 & 122.0 & 97.5 & 125.3 & 106.7 & 143.2 & 76.2 & 103.3 & 149.3 & 207.3 \\
$\epsilon$=0.01 & 37.3 & 39.8 & 77.2 & 99.4 & 74.3 & 93.7 & 164.7 & 160.5 & 155.2 & 177.5 & 129.4 & 165.1 & 99.7 & 126.8 & 190.3 & 233.8 \\
$\epsilon$=0.02 & 51.1 & 51.9 & 139.9 & 164.5 & 151.6 & 173.5 & 199.6 & 183.1 & 222.0 & 237.0 & 161.5 & 188.6 & 135.7 & 156.4 & 230.9 & 267.3 \\
$\epsilon$=0.03 & 95.4 & 93.8 & 178.1 & 196.3 & 219.5 & 240.6 & 224.5 & 204.0 & 262.3 & 269.4 & 184.3 & 214.3 & 175.4 & 194.1 & 267.5 & 291.9 \\
$\epsilon$=0.04 & 113.0 & 107.4 & 204.0 & 218.5 & 256.8 & 271.7 & 276.2 & 241.2 & 276.5 & 268.9 & 213.3 & 232.6 & 212.7 & 231.9 & 311.4 & 329.5 \\
$\epsilon$=0.05 & \textbf{154.2} & \textbf{146.2} & \textbf{224.1} & \textbf{234.9} & \textbf{275.8} & \textbf{292.2} & \textbf{326.6} & \textbf{280.3} & \textbf{314.8} & \textbf{314.0} & \textbf{228.9} & \textbf{254.8} & \textbf{243.3} & \textbf{251.6} & \textbf{338.8} & \textbf{346.9} \\
\hline
\end{tabular}}
\end{table}

\begin{table}[H]
\centering
\caption{\label{Atab:13}Short-term prediction errors of LTD on CMU before and after perturbation in variations of the boundary.}
\scalebox{0.5}{
\begin{tabular}{c|cccc|cccc|cccc}
\hline
Intervals(ms) & 80 & 160 & 320 & 400 & 80 & 160 & 320 & 400 & 80 & 160 & 320 & 400 \\ \hline
scenarios & \multicolumn{4}{c|}{Basketball} & \multicolumn{4}{c|}{Basketball signal} & \multicolumn{4}{c}{Directing traffic} \\ \hline
LTD & 12.8 & 23.1 & 46.9 & 60.1 & 3.0 & 5.2 & 10.8 & 14.1 & 5.8 & 11.3 & 27.3 & 36.6 \\
$\epsilon$=0.01 & 20.9 & 40.0 & 83.2 & 104.3 & 15.6 & 31.6 & 68.0 & 85.6 & 19.4 & 40.3 & 89.1 & 112.3 \\
$\epsilon$=0.02 & 28.4 & 53.3 & 107.4 & 133.0 & 24.8 & 48.2 & 101.6 & 127.1 & 28.9 & 56.9 & 118.8 & 146.8 \\
$\epsilon$=0.03 & 33.0 & 59.3 & 118.7 & 146.4 & 33.8 & 63.9 & 128.9 & \textbf{158.4} & 38.0 & 71.2 & 143.8 & 177.1 \\
$\epsilon$=0.04 & 36.4 & 64.0 & 124.1 & 152.5 & 34.8 & 63.1 & 121.1 & 146.5 & 41.9 & 77.5 & 155.8 & 190.6 \\
$\epsilon$=0.05 & \textbf{41.6} & \textbf{70.8} & \textbf{132.9} & \textbf{161.5} & \textbf{39.5} & \textbf{69.1} & \textbf{130.1} & 157.0 & \textbf{44.7} & \textbf{81.2} & \textbf{161.2} & \textbf{197.2} \\
\hline
\end{tabular}}
\end{table}

\begin{table}[H]
\centering
\caption{\label{Atab:14}Long-term prediction errors of LTD on CMU before and after perturbation in variations of the boundary.}
\scalebox{0.5}{
\begin{tabular}{c|cc|cc|cc|cccc}
\hline
Intervals(ms) & 560 & 1000 & 560 & 1000 & 560 & 1000 & 560 & \multicolumn{1}{c|}{1000} & 560 & 1000 \\ \hline
scenarios & \multicolumn{2}{c|}{Basketball} & \multicolumn{2}{c|}{Basketball signal} & \multicolumn{2}{c|}{Directing traffic} & \multicolumn{2}{c|}{Jumping} & \multicolumn{2}{c}{Running} \\ \hline
LTD & 78.3 & 97.6 & 24.8 & 49.7 & 64.4 & 139.3 & 122.2 & \multicolumn{1}{c|}{168.7} & 41.1 & 68.5 \\
$\epsilon$=0.01 & 138.2 & 156.2 & 130.1 & 195.7 & 161.3 & 251.5 & 173.9 & \multicolumn{1}{c|}{208.9} & 125.8 & 107.3 \\
$\epsilon$=0.02 & 163.0 & 191.1 & 172.6 & 231.2 & 189.0 & 270.8 & 199.1 & \multicolumn{1}{c|}{244.0} & 178.1 & 143.6 \\
$\epsilon$=0.03 & 176.6 & 201.0 & 179.6 & 239.0 & 203.9 & 288.9 & 228.9 & \multicolumn{1}{c|}{254.1} & 208.8 & 172.3 \\
$\epsilon$=0.04 & 194.9 & \textbf{228.5} & 195.2 & 255.0 & \textbf{218.2} & \textbf{299.4} & \textbf{231.9} & \multicolumn{1}{c|}{\textbf{267.2}} & 207.4 & 169.4 \\
$\epsilon$=0.05 & \textbf{200.1} & 227.3 & \textbf{206.7} & \textbf{263.8} & 217.7 & 295.3 & 227.4 & \multicolumn{1}{c|}{249.7} & \textbf{223.1} & \textbf{185.8} \\
\end{tabular}}
\end{table}

\begin{table}[H]
\centering
\caption{\label{Atab:15}Short-term prediction errors of LTD-50 on CMU before and after perturbation in variations of the boundary.}
\scalebox{0.5}{
\begin{tabular}{c|cccc|cccc|cccc}
\hline
Intervals(ms) & 80 & 160 & 320 & 400 & 80 & 160 & 320 & 400 & 80 & 160 & 320 & 400 \\ \hline
scenarios & \multicolumn{4}{c|}{Basketball} & \multicolumn{4}{c|}{Basketball signal} & \multicolumn{4}{c}{Directing traffic} \\ \hline
LTD-50 & 11.0 & 20.4 & 43.7 & 57.0 & 2.6 & 4.8 & 10.3 & 13.4 & 6.4 & 12.2 & 27.8 & 37.4 \\
$\epsilon$=0.01 & 23.2 & 42.3 & 84.5 & 101.5 & 14.5 & 27.8 & 57.1 & 64.7 & 18.7 & 37.4 & 81.1 & 97.0 \\
$\epsilon$=0.02 & 26.8 & 49.7 & 100.6 & 118.9 & 21.6 & 42.4 & 91.2 & 104.4 & 24.4 & 49.1 & 105.7 & 124.2 \\
$\epsilon$=0.03 & 29.9 & 54.7 & 107.2 & 126.4 & 26.8 & 51.7 & 111.8 & 128.6 & 27.7 & 53.8 & 112.6 & 131.6 \\
$\epsilon$=0.04 & 33.9 & 61.2 & 118.7 & 138.6 & 32.1 & 61.2 & \textbf{131.3} & \textbf{151.1} & 31.1 & 60.2 & 126.8 & 146.6 \\
$\epsilon$=0.05 & \textbf{34.5} & \textbf{62.0} & \textbf{121.3} & \textbf{141.3} & \textbf{37.1} & \textbf{66.1} & 128.5 & 144.8 & \textbf{36.8} & \textbf{70.0} & \textbf{142.7} & \textbf{163.9} \\
\hline
\end{tabular}}
\end{table}

\begin{table}[H]
\centering
\caption{\label{Atab:16}Long-term prediction errors of LTD-50 on CMU before and after perturbation in variations of the boundary.}
\scalebox{0.5}{
\begin{tabular}{c|cc|cc|cc|cccc}
\hline
Intervals(ms) & 560 & 1000 & 560 & 1000 & 560 & 1000 & 560 & \multicolumn{1}{c|}{1000} & 560 & 1000 \\ \hline
scenarios & \multicolumn{2}{c|}{Basketball} & \multicolumn{2}{c|}{Basketball signal} & \multicolumn{2}{c|}{Directing traffic} & \multicolumn{2}{c|}{Jumping} & \multicolumn{2}{c}{Running} \\ \hline
LTD-50 & 77.2 & 102.4 & 24.7 & 53.0 & 69.1 & 134.9 & 126.6 & \multicolumn{1}{c|}{159.8} & 42.6 & 77.2 \\
$\epsilon$=0.01 & 116.8 & 141.5 & 112.2 & 170.0 & 138.7 & 227.8 & 152.1 & \multicolumn{1}{c|}{180.0} & 85.9 & 78.6 \\
$\epsilon$=0.02 & 139.1 & 168.6 & 153.3 & 198.5 & 165.4 & 248.6 & 165.3 & \multicolumn{1}{c|}{199.9} & 110.5 & 84.3 \\
$\epsilon$=0.03 & 154.5 & 183.6 & 170.2 & 207.7 & 179.0 & 262.9 & 175.5 & \multicolumn{1}{c|}{209.1} & 131.0 & 100.5 \\
$\epsilon$=0.04 & 171.5 & 198.7 & 205.6 & \textbf{248.5} & 196.1 & 281.2 & 192.4 & \multicolumn{1}{c|}{\textbf{223.9}} & 145.5 & 109.8 \\
$\epsilon$=0.05 & \textbf{186.9} & \textbf{205.6} & \textbf{206.4} & 240.6 & \textbf{197.9} & \textbf{281.3} & \textbf{192.9} & \multicolumn{1}{c|}{217.3} & \textbf{169.7} & \textbf{136.9} \\
\end{tabular}}
\end{table}

\begin{table}[H]
\centering
\caption{\label{Atab:17}Short-term prediction errors of TrajectoryCNN 
on CMU before and after perturbation in variations of the boundary.}
\scalebox{0.5}{
\begin{tabular}{c|cccc|cccc|cccc}
\hline
Intervals(ms) & 80 & 160 & 320 & 400 & 80 & 160 & 320 & 400 & 80 & 160 & 320 & 400 \\ \hline
scenarios & \multicolumn{4}{c|}{Basketball} & \multicolumn{4}{c|}{Basketball signal} & \multicolumn{4}{c}{Directing traffic} \\ \hline
TrajCNN & 11.1 & 19.2 & 42.6 & 56.4 & 2.1 & 3.3 & 8.1 & 11.6 & 5.8 & 10.8 & 22.2 & 29.2 \\
$\epsilon$=0.01 & 23.4 & 43.5 & 94.2 & 119.3 & 17.2 & 35.2 & 81.7 & 105.3 & 24.5 & 48.7 & 107.1 & 135.4 \\
$\epsilon$=0.02 & 34.0 & 63.3 & 132.9 & 166.4 & 29.5 & 60.2 & 132.2 & 166.5 & 38.3 & 75.7 & 162.3 & 203.1 \\
$\epsilon$=0.03 & 42.7 & 76.7 & 152.2 & 187.6 & 38.8 & 77.4 & 168.2 & 211.4 & 49.5 & 94.2 & 191.1 & 233.9 \\
$\epsilon$=0.04 & 48.7 & 86.7 & 170.2 & 208.4 & 45.5 & 88.7 & 189.9 & 237.3 & 54.1 & 102.2 & 204.6 & 249.3 \\
$\epsilon$=0.05 & \textbf{54.9} & \textbf{95.9} & \textbf{183.4} & \textbf{221.7} & \textbf{52.2} & \textbf{99.2} & \textbf{209.1} & \textbf{260.8} & \textbf{62.9} & \textbf{115.3} & \textbf{222.3} & \textbf{265.8} \\
\hline
\end{tabular}}
\end{table}

\begin{table}[H]
\centering
\caption{\label{Atab:18}Long-term prediction errors of TrajectoryCNN on CMU before and after perturbation in variations of the boundary.}
\scalebox{0.5}{
\begin{tabular}{c|cc|cc|cc|cccc}
\hline
Intervals(ms) & 560 & 1000 & 560 & 1000 & 560 & 1000 & 560 & \multicolumn{1}{c|}{1000} & 560 & 1000 \\ \hline
scenarios & \multicolumn{2}{c|}{Basketball} & \multicolumn{2}{c|}{Basketball signal} & \multicolumn{2}{c|}{Directing traffic} & \multicolumn{2}{c|}{Jumping} & \multicolumn{2}{c}{Running} \\ \hline
TrajCNN & 72.9 & 112.0 & 24.8 & 59.8 & 53.7 & 116.3 & 134.5 & \multicolumn{1}{c|}{175.3} & 42.2 & 45.3 \\
$\epsilon$=0.01 & 124.0 & 168.1 & 99.8 & 173.4 & 186.1 & 312.9 & 189.6 & \multicolumn{1}{c|}{241.0} & 59.7 & 56.4 \\
$\epsilon$=0.02 & 155.9 & 206.9 & 156.3 & 236.0 & 253.4 & 374.8 & 226.9 & \multicolumn{1}{c|}{279.9} & 82.2 & 72.1 \\
$\epsilon$=0.03 & 185.2 & \textbf{241.1} & 203.6 & 285.3 & 292.6 & \textbf{412.0} & 252.6 & \multicolumn{1}{c|}{315.6} & 95.7 & 82.2 \\
$\epsilon$=0.04 & 187.6 & 237.7 & 226.1 & 302.0 & 285.1 & 406.4 & 265.6 & \multicolumn{1}{c|}{327.3} & 128.2 & 115.6 \\
$\epsilon$=0.05 & \textbf{191.5} & 240.1 & \textbf{226.7} & \textbf{316.2} & \textbf{302.9} & 408.6 & \textbf{289.9} & \multicolumn{1}{c|}{\textbf{350.6}} & \textbf{158.3} & \textbf{147.6} \\
\end{tabular}}
\end{table}

\begin{table}[H]
\centering
\caption{\label{Atab:19}Short-term prediction errors of TrajectoryCNN-50 on CMU before and after perturbation in variations of the boundary.}
\scalebox{0.5}{
\begin{tabular}{c|cccc|cccc|cccc}
\hline
Intervals(ms) & 80 & 160 & 320 & 400 & 80 & 160 & 320 & 400 & 80 & 160 & 320 & 400 \\ \hline
scenarios & \multicolumn{4}{c|}{Basketball} & \multicolumn{4}{c|}{Basketball signal} & \multicolumn{4}{c}{Directing traffic} \\ \hline
TrajCNN-50 & 11.8 & 19.7 & 41.4 & 53.6 & 1.8 & 3.1 & 7.4 & 10.6 & 5.8 & 11.7 & 25.6 & 34.1 \\
$\epsilon$=0.01 & 19.4 & 34.6 & 70.7 & 88.4 & 15.8 & 31.5 & 70.1 & 90.0 & 17.1 & 36.4 & 82.4 & 105.4 \\
$\epsilon$=0.02 & 27.8 & 50.4 & 101.9 & 125.8 & 30.0 & 60.1 & 130.6 & 164.2 & 28.5 & 59.1 & 131.9 & 167.2 \\
$\epsilon$=0.03 & 36.3 & 65.5 & 127.2 & 154.9 & 39.5 & 75.8 & 160.1 & 200.6 & 36.6 & 73.8 & 155.9 & 192.7 \\
$\epsilon$=0.04 & 45.3 & 80.4 & 153.9 & 185.7 & 49.1 & 93.3 & 185.7 & 224.5 & 47.3 & 91.9 & 188.1 & 230.8 \\
$\epsilon$=0.05 & \textbf{53.9} & \textbf{95.0} & \textbf{178.3} & \textbf{213.4} & \textbf{57.5} & \textbf{108.1} & \textbf{216.8} & \textbf{264.0} & \textbf{55.1} & \textbf{107.2} & \textbf{219.5} & \textbf{268.3} \\
\hline
\end{tabular}}
\end{table}

\begin{table}[H]
\centering
\caption{\label{Atab:20}Long-term prediction errors of TrajectoryCNN-50 on CMU before and after perturbation in variations of the boundary.}
\scalebox{0.5}{
\begin{tabular}{c|cc|cc|cc|cccc}
\hline
Intervals(ms) & 560 & 1000 & 560 & 1000 & 560 & 1000 & 560 & \multicolumn{1}{c|}{1000} & 560 & 1000 \\ \hline
scenarios & \multicolumn{2}{c|}{Basketball} & \multicolumn{2}{c|}{Basketball signal} & \multicolumn{2}{c|}{Directing traffic} & \multicolumn{2}{c|}{Jumping} & \multicolumn{2}{c}{Running} \\ \hline
TrajCNN-50 & 80.0 & 116.8 & 19.5 & 52.8 & 63.8 & 131.6 & 115.8 & \multicolumn{1}{c|}{156.7} & 56.3 & 77.7 \\
$\epsilon$=0.01 & 107.5 & 141.1 & 67.2 & 117.6 & 120.4 & 212.7 & 153.6 & \multicolumn{1}{c|}{193.0} & 79.7 & 89.9 \\
$\epsilon$=0.02 & 135.0 & 165.3 & 135.2 & 187.7 & 171.0 & 271.9 & 185.9 & \multicolumn{1}{c|}{223.9} & 109.4 & 112.1 \\
$\epsilon$=0.03 & 169.5 & 194.6 & 205.2 & 266.0 & 214.8 & 317.2 & 212.1 & \multicolumn{1}{c|}{246.4} & 132.4 & 125.7 \\
$\epsilon$=0.04 & 189.2 & 204.7 & 227.1 & 295.6 & 239.4 & 325.2 & 272.0 & \multicolumn{1}{c|}{296.5} & 173.6 & 163.0 \\
$\epsilon$=0.05 & \textbf{211.4} & \textbf{230.1} & \textbf{271.1} & \textbf{342.4} & \textbf{261.7} & \textbf{359.1} & \textbf{293.0} & \multicolumn{1}{c|}{\textbf{314.4}} & \textbf{173.7} & \textbf{159.1} \\
\end{tabular}}
\end{table}

\begin{table}[H]
\centering
\caption{\label{Atab:21}Short-term prediction errors of HRI on CMU before and after perturbation in variations of the boundary.}
\scalebox{0.5}{
\begin{tabular}{c|cccc|cccc|cccc}
\hline
Intervals(ms) & 80 & 160 & 320 & 400 & 80 & 160 & 320 & 400 & 80 & 160 & 320 & 400 \\ \hline
scenarios & \multicolumn{4}{c|}{Basketball} & \multicolumn{4}{c|}{Basketball signal} & \multicolumn{4}{c}{Directing traffic} \\ \hline
HRI & 12.4 & 22.8 & 50.5 & 65.6 & 2.7 & 4.6 & 9.8 & 13.1 & 5.8 & 11.9 & 26.5 & 35.6 \\
$\epsilon$=0.01 & 20.6 & 43.9 & 109.3 & 144.2 & 21.6 & 50.1 & 127.1 & 167.6 & 20.4 & 47.7 & 114.5 & 148.6 \\
$\epsilon$=0.02 & 28.6 & 59.8 & 144.9 & 189.6 & 34.2 & 73.2 & 170.2 & 217.6 & 31.1 & 68.7 & 160.5 & 206.3 \\
$\epsilon$=0.03 & 36.1 & 73.1 & 171.1 & 221.1 & 39.5 & 80.5 & 180.4 & 228.9 & 39.4 & 81.6 & 182.2 & 232.6 \\
$\epsilon$=0.04 & 44.2 & 86.3 & 192.6 & 245.7 & 45.6 & 89.1 & 192.1 & 240.6 & 42.7 & 87.2 & 193.6 & 246.0 \\
$\epsilon$=0.05 & \textbf{50.1} & \textbf{96.5} & \textbf{209.9} & \textbf{265.6} & \textbf{51.4} & \textbf{98.8} & \textbf{209.9} & \textbf{261.7} & \textbf{50.6} & \textbf{101.3} & \textbf{217.9} & \textbf{273.1} \\
\hline
\end{tabular}}
\end{table}

\begin{table}[H]
\centering
\caption{\label{Atab:22}Long-term prediction errors of HRI on CMU before and after perturbation in variations of the boundary.}
\scalebox{0.5}{
\begin{tabular}{c|cc|cc|cc|cccc}
\hline
Intervals(ms) & 560 & 1000 & 560 & 1000 & 560 & 1000 & 560 & \multicolumn{1}{c|}{1000} & 560 & 1000 \\ \hline
scenarios & \multicolumn{2}{c|}{Basketball} & \multicolumn{2}{c|}{Basketball signal} & \multicolumn{2}{c|}{Directing traffic} & \multicolumn{2}{c|}{Jumping} & \multicolumn{2}{c}{Running} \\ \hline
HRI & 90.1 & 126.3 & 21.2 & 48.2 & 59.9 & 136.8 & 116.4 & \multicolumn{1}{c|}{175.4} & 35.0 & 70.7 \\
$\epsilon$=0.01 & 206.0 & 268.1 & 239.6 & 320.7 & 220.1 & 323.0 & 231.2 & \multicolumn{1}{c|}{334.3} & 186.5 & 230.0 \\
$\epsilon$=0.02 & 266.2 & 337.4 & 290.1 & 348.9 & 289.0 & 377.7 & 288.8 & \multicolumn{1}{c|}{369.0} & 250.3 & 269.9 \\
$\epsilon$=0.03 & 303.3 & 376.5 & 303.2 & 362.6 & 314.8 & 403.8 & 329.3 & \multicolumn{1}{c|}{416.8} & 292.3 & 305.3 \\
$\epsilon$=0.04 & 328.4 & 393.2 & 309.8 & 368.4 & 331.2 & 412.2 & 329.3 & \multicolumn{1}{c|}{401.3} & 325.3 & 349.4 \\
$\epsilon$=0.05 & \textbf{349.4} & \textbf{421.9} & \textbf{334.7} & \textbf{387.5} & \textbf{357.3} & \textbf{431.4} & \textbf{375.9} & \multicolumn{1}{c|}{\textbf{430.3}} & \textbf{350.8} & \textbf{372.7} \\
\end{tabular}}
\end{table}

\begin{table}[H]
\centering
\caption{\label{Atab:23}Short-term prediction errors of Multi-head TrajectoryCNN on CMU before and after perturbation in variations of the boundary.}
\scalebox{0.5}{
\begin{tabular}{c|cccc|cccc|cccc}
\hline
Intervals(ms) & 80 & 160 & 320 & 400 & 80 & 160 & 320 & 400 & 80 & 160 & 320 & 400 \\ \hline
scenarios & \multicolumn{4}{c|}{Basketball} & \multicolumn{4}{c|}{Basketball signal} & \multicolumn{4}{c}{Directing traffic} \\ \hline
MH-TrajCNN & 11.6 & 20.6 & 43.0 & 56.5 & 2.4 & 5.0 & 11.9 & 15.9 & 6.2 & 12.8 & 29.1 & 39.8 \\
$\epsilon$=0.01 & 16.6 & 28.7 & 57.5 & 73.2 & 9.8 & 15.6 & 28.0 & 34.0 & 12.3 & 22.4 & 46.3 & 59.9 \\
$\epsilon$=0.02 & 23.6 & 38.7 & 74.3 & 92.6 & 18.8 & 29.5 & 52.2 & 62.3 & 20.2 & 35.1 & 68.2 & 85.5 \\
$\epsilon$=0.03 & 30.4 & 49.4 & 91.4 & 111.8 & 28.7 & 45.2 & 82.1 & 99.4 & 29.5 & 50.7 & 96.8 & 119.3 \\
$\epsilon$=0.04 & 38.3 & 61.9 & 110.1 & 132.6 & 39.4 & 61.3 & 108.0 & 129.7 & 37.1 & 62.6 & 118.6 & 144.2 \\
$\epsilon$=0.05 & \textbf{45.4} & \textbf{70.7} & \textbf{122.9} & \textbf{146.4} & \textbf{45.9} & \textbf{69.9} & \textbf{118.9} & \textbf{140.0} & \textbf{45.0} & \textbf{76.6} & \textbf{146.5} & \textbf{178.9} \\
\hline
\end{tabular}}
\end{table}

\begin{table}[H]
\centering
\caption{\label{Atab:24}Long-term prediction errors of Multi-head TrajectoryCNN on CMU before and after perturbation in variations of the boundary.}
\scalebox{0.5}{
\begin{tabular}{c|cc|cc|cc|cccc}
\hline
Intervals(ms) & 560 & 1000 & 560 & 1000 & 560 & 1000 & 560 & \multicolumn{1}{c|}{1000} & 560 & 1000 \\ \hline
scenarios & \multicolumn{2}{c|}{Basketball} & \multicolumn{2}{c|}{Basketball signal} & \multicolumn{2}{c|}{Directing traffic} & \multicolumn{2}{c|}{Jumping} & \multicolumn{2}{c}{Running} \\ \hline
MH-TrajCNN & 84.1 & 126.5 & 32.7 & 73.3 & 71.7 & 135.4 & 118.7 & \multicolumn{1}{c|}{158.8} & 65.0 & 76.1 \\
$\epsilon$=0.01 & 97.2 & 137.2 & 56.0 & 102.1 & 102.1 & 177.9 & 142.0 & \multicolumn{1}{c|}{180.1} & 86.6 & 88.9 \\
$\epsilon$=0.02 & 116.5 & 154.7 & 97.4 & 156.6 & 133.7 & 219.5 & 166.6 & \multicolumn{1}{c|}{201.1} & 106.7 & 100.8 \\
$\epsilon$=0.03 & 133.8 & 171.5 & 116.7 & 177.1 & 167.0 & 264.5 & 202.8 & \multicolumn{1}{c|}{239.4} & 131.0 & 120.2 \\
$\epsilon$=0.04 & 147.2 & 182.1 & 148.6 & 215.2 & 192.3 & 290.4 & 233.4 & \multicolumn{1}{c|}{270.4} & 160.1 & 147.1 \\
$\epsilon$=0.05 & \textbf{166.4} & \textbf{199.2} & \textbf{169.5} & \textbf{237.6} & \textbf{204.8} & \textbf{297.6} & \textbf{269.0} & \multicolumn{1}{c|}{\textbf{301.1}} & \textbf{204.0} & \textbf{194.7} \\
\end{tabular}}
\end{table}
\section{Discussion about the Non-Monotonic Growth}
\label{app:2}
According to the results in \ref{app:1}, the non-monotonic growth of specific activities' MPJPE as $\epsilon$ or time interval grows happens differently among models. Based on our observation, LTD encounters this non-monotonic growth phenomenon as $\epsilon$ is enlarged more than TrajectoryCNN, with HRI following behind. In contrast, Multi-Head TrajectoryCNN hardly meets such a phenomenon. Note that our adversarial attack only affects the output by perturbing input data while the algorithm still models the predicted motion. Although the disturbed output is far from the ground truth, its output still follows the motion that the algorithm comprehends, which cannot determine the concrete location of the joints in our attack method. The increment of $\epsilon$ indicates the increment of the attack intensity. If the $\epsilon$ alteration can change the errors directly, the model thoroughly understands the perturbed human motion sequence. Therefore, the ranking above can be considered as the ranking of how thoroughly the models comprehend the relation between the history and future motion sequence. The classification head of the Multi-Head TrajectoryCNN directly helps its comprehension of human motion, which is why it hardly meets the phenomenon. The attention mechanism helps HRI focus on the critical feature of human motion, which boosts the understanding of human motion in HRI. TrajectoryCNN and LTD do not have direct enhancements, so they fall back in the ranking. Comparing the results between LTD and LTD-50, or TrajCNN and TrajCNN-50, we can find that the non-monotonic phenomenon happens much less, which indicates that increasing the input length is a simple way to support the model.\par
Contrary to the above ranking, the ranking of meeting the non-monotonic growth of errors as the time interval is enlarged from frequent to occasional is HRI, Multi-Head TrajectoryCNN, LTD, TrajectoryCNN. Except for the sudden decrease caused by changing model, the non-monotonic growth along time intervals only happens in long-term prediction. Except for the results of HRI, Most activities meet this phenomenon after $\epsilon$ reaches 0.03. Only “greeting” predicted by LTD, the results of TrajectoryCNN on 3DPW, the results of HRI on H3.6M and 3DPW, “discussion,” “waiting,” “walking together,” and “running” predicted by Multi-Head TrajectoryCNN are earlier than others. Clean “greeting” motion prediction of LTD also has non-monotonic growth. Therefore, finding the same trend under attack is possible. The predicted sequence begins from the last pose the model observed, so short-term predictions are within a limited range and will not face non-monotonic growth. As the time interval enlarged, the possibility of predicted joints’ location increased. Therefore, non-monotonic growth is observed in a wide range. To figure out what kind of activity will cause the non-monotonic growth, we find “walking,” “waiting,” “eating,” “greeting,” “walking together,” and “running" encounter the phenomenon. Empirically, these activities are regular or rapid. The activities from 3DPW are varied from the other two datasets, which contain more activities under wild conditions. As a result, distances between the perturbed prediction and ground truth of these activities are hard to be determined even though we maximize the average error. However, half of the activities predicted by HRI meet the non-monotonic growth along time intervals. This may be because HRI utilizes the predicted sequence as the input to reach the long-term prediction, and finally, the prediction will be more and more out of control.\par

\end{document}